\documentclass[11pt]{article}
\usepackage[final]{acl}

\usepackage{times}
\usepackage{latexsym}

\usepackage[T1]{fontenc}
\usepackage[utf8]{inputenc}

\usepackage{microtype}

\usepackage{inconsolata}

\usepackage{graphicx}

\usepackage{booktabs}
\usepackage{pifont}
\usepackage{array}
\usepackage{balance}

\usepackage[ruled,vlined,linesnumbered]{algorithm2e}
\SetAlgoNlRelativeSize{-1}
\SetAlCapSkip{0.5em}

\usepackage{listings}
\usepackage{xcolor}
\usepackage{colortbl}
\usepackage{amsmath}
\usepackage{amssymb}
\usepackage{multirow}
\usepackage{longtable}
\usepackage{xltabular}
\usepackage{placeins}
\usepackage{pdflscape}
\usepackage{siunitx} 
\usepackage{url}

\newcommand{\squishlist}{
   \begin{list}{$\bullet$}
    { \setlength{\itemsep}{1pt}
      \setlength{\parsep}{0pt}
      \setlength{\topsep}{2pt}
      \setlength{\partopsep}{0pt}
      \setlength{\listparindent}{-2pt}
      \setlength{\itemindent}{-5pt}
      \setlength{\leftmargin}{1.5em}
      \setlength{\labelwidth}{0em}
      \setlength{\labelsep}{0.5em} 
    } 
}

\newcommand{\DBN}{\scalebox{0.92}[1.0]{\mbox{{\textsc{DNBench}}}}}
\newcommand{\DNS}{\scalebox{0.92}[1.0]{\mbox{{\textsc{DNB-Score}}}}}

\newcommand{\squishend}{
    \end{list}  }

\long\def\shorten#1{}

\newcommand{\cmark}{\textcolor[HTML]{228B22}{\ding{51}}}
\newcommand{\xmark}{\ding{55}}
\renewcommand{\lstlistingname}{Prompt}

\title{Can LLMs Normalize Databases?\\A Benchmark and Multi-Agent Framework for Schema Normalization}

\author{
\textbf{Dong-Jae Koh\textsuperscript{*}}
\textbf{Huisu Kim\textsuperscript{*}}
\textbf{SeongHwan Yoon\textsuperscript{*}}\\
\textbf{Lasse M. Jantsch}
\textbf{Chun-Hee Lee}
\textbf{Seonghyeon Lee}
\textbf{Young-Kyoon Suh$^{\dagger}$}
\\
School of Computer Science and Engineering, Kyungpook National University, South Korea
\\
\texttt{\{djkoh, kimisu0712, shyoon0214,lassejantsch, chunhee, sh0416, yksuh\}@knu.ac.kr}
}

\let\svthefootnote\thefootnote
\newcommand\freefootnote[1]{%
    \let\thefootnote\relax%
    \footnotetext{#1}%
    \let\thefootnote\svthefootnote%
}
\begin{document}
\maketitle
\freefootnote{\textsuperscript{*} Equal Contributions}
\freefootnote{$^{\dagger}$ Corresponding author}
\begin{abstract}
  Large Language Models (LLMs) are increasingly used to generate structured outputs, but their reliability remains unclear when those outputs must satisfy \hbox{database-level} constraints. We study this issue through database normalization, which requires formal schema constraints and relational dependencies. We introduce a \emph{Database Normalization \hbox{Benchmark}} (\DBN), comprising \hbox{3,275} samples for evaluating \shorten{\hbox{end-to-end}} \hbox{LLM-based} database normalization from 1NF to BCNF. {\DBN} uses a three-axis protocol to measure \emph{semantic equivalence}, \emph{structural accuracy}, and \emph{logical validity}. Across \emph{Single}, \emph{Complex}, and \emph{FD-withheld} settings, {\DBN} uncovers recurring failures in dependency inference, schema decomposition, and inter-table constraint reconstruction. We further propose \emph{\hbox{Multi-Agent} Reasoning for Schemas} (MARS), which separates dependency extraction, violation diagnosis, and decomposition planning from schema generation and verification. MARS improves the {\DNS} by $0.227$ in absolute terms, a relative gain of $84.0\%$, over the single-prompt baseline. 
  All artifacts will be released upon acceptance.
\end{abstract}

\section{Introduction}
Large language models (LLMs)~\cite{NEURIPS2020_1457c0d6} have been explored for structured-data reasoning and database-related tasks. These tasks require outputs that satisfy formal constraints. In this context, LLMs’ ability to follow \hbox{natural-language} instructions, reason over tabular data, and generate structured outputs makes them a promising approach for automating database \hbox{workflows}.

Normalizing databases using LLMs highlights the inherent difficulty of these tasks because they require both reasoning about dependencies and generating schemas that satisfy formal relational constraints. Database normalization is essential for eliminating redundancy, anomalies, and data inconsistency in relational schemas~\cite{Codd1970ARM, kent1983simple}. However, normalization is hard to automate because it requires reasoning over functional dependencies, candidate keys, lossless-join decompositions, and dependency preservation~\cite{codd1972further}.

Despite its foundational role in data management, existing studies \hbox{provide} limited evidence for whether LLMs can handle database normalization. NormTab~\cite{nahid-rafiei-2024-normtab} and TABARD~\cite{choudhury-etal-2025-tabard} address normalization-related issues primarily in \hbox{instance-level} table-cleaning or anomaly-detection settings, rather than formal schema decomposition. Miffie~\cite{jo2026databasenormalizationdualllmselfrefinement} targets 1NF-to-3NF normalization using an LLM generator-verifier loop. Since its verifier is also LLM-based, it lacks explicit schema-level guarantees such as lossless join or key validity. Consequently, the reliability of LLM-based database normalization remains \hbox{insufficiently} understood.

\begin{table*}[!t]
  \centering
  \footnotesize
  \renewcommand{\arraystretch}{1.35}
  \setlength{\tabcolsep}{3pt}
  \begin{tabular*}{\textwidth}{@{\extracolsep{\fill}}|l|c|*{7}{>{\centering\arraybackslash}m{1.05cm}|}c|}
    \hline
    \multirow{2}{*}{\textbf{Method}}
    & \multirow{2}{*}{\shortstack{\textbf{LLM}\\\textbf{Based}}}
    & \multicolumn{5}{c|}{\textbf{Normalization}}
    & \multicolumn{2}{c|}{\textbf{Evaluation}}
    & \multirow{2}{*}{\shortstack{\textbf{Benchmark}\\\textbf{Provided}}} \\
    \cline{3-9}
    & & {\scriptsize\textbf{1NF}} & {\scriptsize\textbf{2NF}} & {\scriptsize\textbf{3NF}} & {\scriptsize\textbf{BCNF}} & {\scriptsize\textbf{Complex}}
    & {\scriptsize\textbf{Schema}} & {\scriptsize\textbf{Expl.}} & \\
    \hline
    RDBNorma~\citeyearpar{dongare2011rdbnormasemiautomatedtool}          & \xmark & \xmark & \cmark & \cmark & \xmark & \xmark & \xmark & \xmark & \xmark \\
    EDNA~\citeyearpar{floyd2014edna}                     & \xmark & \xmark & \cmark & \cmark & \cmark & \xmark & \xmark & \xmark & \xmark \\
    NormTab~\citeyearpar{nahid-rafiei-2024-normtab}             & \cmark & \cmark & \xmark & \xmark & \xmark & \xmark & \xmark & \xmark & \xmark \\
    TABARD~\citeyearpar{choudhury-etal-2025-tabard}      & \cmark & \xmark & \xmark & \xmark & \xmark & \xmark & \cmark & \xmark & \cmark \\
    Miffie (Dual-LLM SR)~\citeyearpar{jo2026databasenormalizationdualllmselfrefinement}  & \cmark & \cmark & \cmark & \cmark & \xmark & \xmark & \cmark & \xmark & \xmark \\
    \hline
    \rowcolor{gray!10}
    \textbf{{\DBN} (ours)}                                  & \cmark & \cmark & \cmark & \cmark & \cmark & \cmark & \cmark & \cmark & \cmark \\
    \hline
  \end{tabular*}
  \caption{Comparison of {\DBN} with prior tools and methods. \textbf{Normalization} groups 1NF to BCNF support and Complex (multi-violation) handling. \textbf{Evaluation} groups schema- and explanation-level scoring.}
  \label{tab:comparison}
  \vspace{-2ex}
\end{table*}

To address these concerns, we introduce a comprehensive \emph{Database Normalization Benchmark} (\DBN) for evaluating \hbox{end-to-end} \hbox{LLM-driven} database normalization. \DBN \space provides 3,275 samples from realistic relational schemas, covering 1NF-to-BCNF violations with gold labels and decomposition targets. Notably, it also includes compliant schemas with no normal-form violations, enabling the evaluation of whether models can correctly recognize already-normalized schemas. To quantify the quality of generated data definition language (DDL) statements on these real-world-derived source schemas with controlled synthetic violations, we propose a three-axis evaluation protocol that measures \emph{semantic equivalence}, \emph{structural accuracy}, and \emph{logical validity}, unified into the \DNS. We also design \emph{Multi-Agent Reasoning for Schemas} (MARS). This framework decomposes database normalization into subtasks and distributes them among collaborating LLM agents that extract dependency evidence, diagnose violations, generate schemas, and verify the results.

Our experiments comprehensively assess \hbox{LLM-driven} database normalization through {\DBN}, which quantifies the quality of generated DDLs. We also show that MARS achieves, on average, a $0.227$ higher {\DNS} ($84.0\%$ relative) than the single-prompt baseline.  

Our contributions are as follows:
\squishlist
\item We present {\DBN}, a 3,275-sample benchmark to evaluate LLM-based database normalization across 1NF-to-BCNF violations and already-normalized schemas.
\item We propose a three-axis evaluation protocol that assesses semantic equivalence, structural accuracy, and logical validity.
\item We show that current LLMs struggle with dependency inference, schema decomposition, and inter-table constraint reconstruction.
\item We propose MARS, a multi-agent normalization framework that substantially outperforms single-prompt and self-refinement baselines.
\squishend

\begin{figure*}[!tp]
  \centering
  \includegraphics[
    width=\textwidth,
  ]{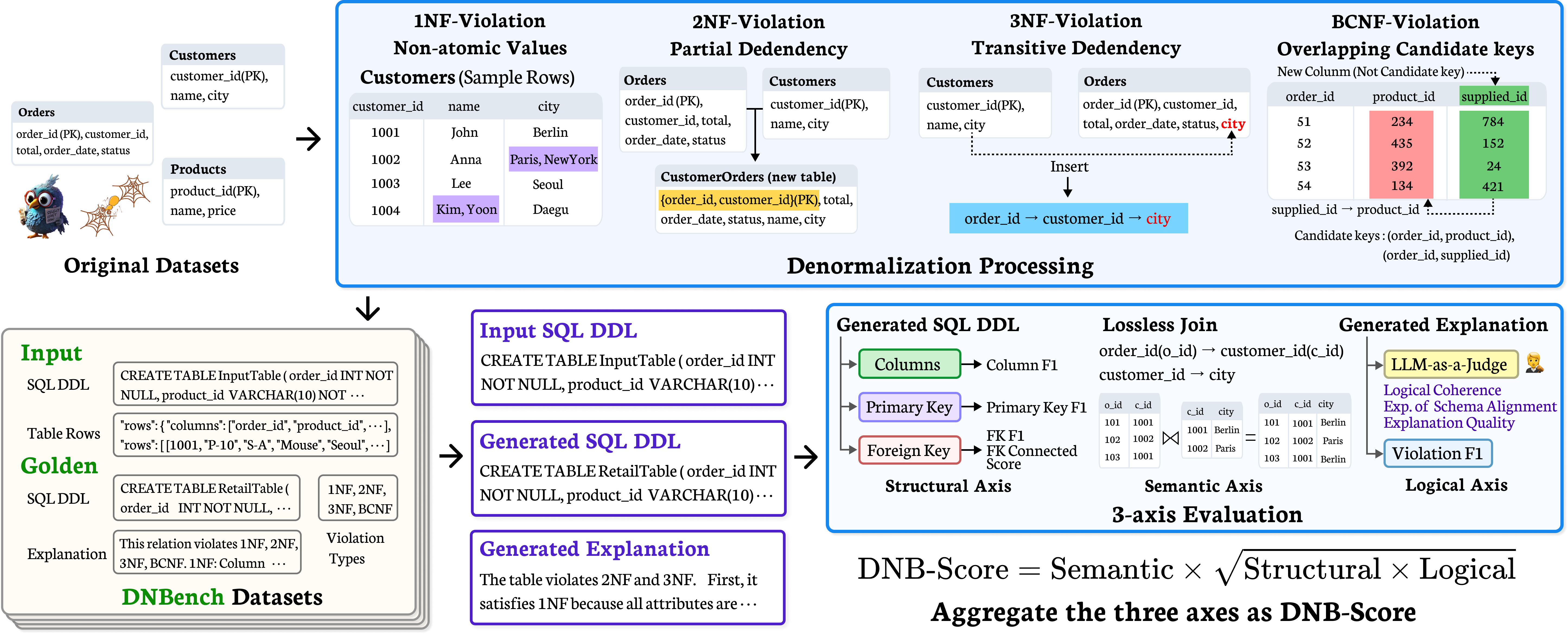}
  \vspace{-3.5ex}
  \caption{Overview of the database normalization benchmark (\DBN) pipeline. {\DBN} generates controlled denormalized schemas with gold violation labels and decompositions, then evaluates LLM-generated DDL, violation types, and explanations along semantic, structural, and logical axes.}
  \label{fig:overview}
  \vspace{-1.5ex}
\end{figure*}

\section{Related Work\label{sec:related}}
Table~\ref{tab:comparison} compares {\DBN} with prior normalization tools and LLM-based methods. Existing work falls into two categories: \emph{traditional tools} that assist with normalization rules and \emph{LLM-based systems} that address \hbox{table cleaning} or limited normalization. However, neither provides a benchmark to evaluate whether LLMs can \hbox{reliably} perform database \hbox{normalization}. {\DBN} bridges this gap with a comprehensive benchmark and a multi-faceted evaluation protocol.

\subsection{Traditional Normalization Tools}
Traditional normalization tools are predominantly rule-based or semi-automatic assistant systems rather than evaluation benchmarks. Tools such as RDBNorma \citep{dongare2011rdbnormasemiautomatedtool} and EDNA \citep{floyd2014edna} support normalization up to 3NF or BCNF using predefined rules. However, their evaluations focus primarily on runtime efficiency or rule execution, lacking quantitative protocols for semantic preservation, key validity, and complex violation diagnosis.

\subsection{LLM-based Normalization}
Recent work applies LLMs to tabular reasoning but targets narrower tasks or lacks structural guarantees. Methods like NormTab \citep{nahid-rafiei-2024-normtab} and TABARD \citep{choudhury-etal-2025-tabard} focus on table preprocessing for QA or cell-level anomaly detection, without addressing functional dependency (FD) reasoning or schema decomposition. Most relevant to our work, Miffie \citep{jo2026databasenormalizationdualllmselfrefinement} uses a Dual-LLM framework for \hbox{1NF-to-3NF} normalization. However, its LLM-based verifier lacks explicit checks for lossless join, key validity, or foreign-key connectivity. {\DBN} instead combines explicit semantic and structural checks with LLM-based logical judgment to test whether decompositions are semantically equivalent, structurally accurate, and logically valid.
\section{Database Normalization Benchmark\label{sec:methodology}}
Figure~\ref{fig:overview} presents an overview of {\DBN}, which consists of two components: (i)~a denormalization pipeline that injects controlled normal form (NF) violations into relational schemas, yielding samples annotated with gold violation labels and gold decomposition targets, and (ii)~a \hbox{three-axis} evaluation protocol that measures the semantic equivalence, structural accuracy, and logical validity of model outputs. Given a sample, an LLM generates normalized DDL, violation labels, and an explanation; the three-axis scores are then combined into a unified {\DNS} that summarizes the overall normalization quality.
Section~\ref{sec:dataset} details the construction pipeline, and Section~\ref{sec:three_axis_eval_protocol} defines the evaluation axes and {\DNS}.

\subsection{Dataset Construction}
\label{sec:dataset}
We construct {\DBN} from Spider~\cite{yu-etal-2018-spider} and BIRD~\cite{li2023can}, two \hbox{text-to-SQL} corpora containing real-world relational schemas with primary key (PK) and foreign key (FK) \hbox{annotations}. Using these schemas as a foundation, we introduce controlled, labeled normal-form (NF) violations while preserving key constraints and \hbox{providing} gold decompositions. Three database \hbox{experts} audited the resulting samples, verifying the 1NF-to-BCNF violation patterns, chain-rule labels, and gold decompositions. 
Table~\ref{tab:dataset_stats} summarizes the source DBs and the resulting {\DBN} dataset. It also includes NONE cases with no NF violations to evaluate whether models can recognize normalized schemas.

\begin{table}[t!]
  \centering
  \footnotesize
  \begin{tabular*}{\columnwidth}{@{\extracolsep{\fill}} lccc}
    \toprule
    \textbf{Dataset Attribute} & \textbf{Spider} & \textbf{BIRD} & \textbf{All} \\
    \midrule
    \multicolumn{4}{c}{\textit{Original datasets}} \\
    \midrule
    Number of DBs & 200 & 95 & 295 \\
    Number of domains & 138 & 37 & 175 \\
    Number of tables & 1{,}020 & 694 & 1{,}714 \\
    \midrule
    \multicolumn{4}{c}{\textit{{\DBN} dataset}} \\
    \midrule
    Dev DBs & 108 & 33 & 141 \\
    Test DBs & 74 & 24 & 98 \\
    \midrule
    Dev samples & 1{,}495 & 495 & 1{,}990 \\
    Test samples & 900 & 385 & 1{,}285 \\
    Total samples & 2{,}395 & 880 & 3{,}275 \\
    \bottomrule
  \end{tabular*}
  \vspace{-1.5ex}
  \caption{Descriptive statistics for the source Spider and BIRD datasets and the resulting {\DBN} dataset.}
  \label{tab:dataset_stats}
  \vspace{-3ex}
\end{table}

The {\DBN} construction follows a three-stage \hbox{denormalization} pipeline: (1)~FK-aware subsampling, (2)~hybrid discovery--synthetic violation injection, and (3)~chain-rule labeling.

\paragraph{Stage 1: Foreign Key-aware \hbox{Subsampling.}}
We first downsample each source database while preserving FK constraints. The sampling procedure traverses the FK graph from leaf tables upward, samples up to $n$ tuples from each source table, and expands the selected key sets through FK and self-FK closures. It then applies a global size cap and removes remaining FK violations through cascade cleanup. This produces compact database instances that remain referentially valid. Full details and the complete algorithm are provided in Appendix~\ref{appendix:sampling}.

\vspace{-1ex}
\paragraph{Stage 2: Discovery and Synthesis of \hbox{Violations}.}
For each source schema, we identify NF violation patterns: multivalued cells violating 1NF, partial dependencies violating 2NF, transitive dependencies violating 3NF, and non-superkey determinants violating BCNF.

Because real schemas often contain entangled rather than isolated violations, discovery alone is insufficient for controlled evaluation. We thus complement it with deterministic synthesizers for 2NF, 3NF, and BCNF. These synthesizers inject violations through PK splitting, \hbox{transitive-dependency} injection, and candidate-key augmentation, respectively. We compose synthesized violations in reverse chain-rule order---BCNF, 3NF, 2NF, and 1NF---so that lower-form injections do not erase previously introduced higher-form violations. This construction yields composite samples in which all intended violations remain observable.

\vspace{-1ex}
\paragraph{Stage 3: Chain-rule Labeling.}
Each sample provides two gold decompositions: a \emph{single} target resolving the earliest violation, and a \emph{combined} target that resolves the entire violation chain.

The composite synthesizers in Stage~2 ensure that fixing a lower-form violation does not automatically remove higher-form violations. Thus, each violation must be handled explicitly, making the \emph{combined} target distinct from the \emph{single} target. This enables {\DBN} to test whether a model fixes only the earliest violation or normalizes the full violation chain. Each sample also includes an expected FK list, allowing FK correctness to be scored without re-parsing the generated DDL.


\paragraph{{\DBN} Sample Configuration.} Combining the synthetic violation chains produced by this pipeline with compliant NONE cases, the final {\DBN} benchmark comprises five sample categories: 1NF--2NF--3NF--BCNF, 2NF--3NF--BCNF, 3NF--BCNF, BCNF, and NONE. Details are provided in Appendix~\ref{appendix: Violation Categories by Normal-Form}.


\subsection{Three-axis Evaluation Protocol~\label{sec:three_axis_eval_protocol}}
We evaluate each LLM output along three \hbox{complementary }axes: \emph{semantic equivalence}, \emph{structural accuracy}, and \emph{logical validity}. These axes capture whether a generated decomposition preserves the original data, correctly reconstructs the schema components, and provides correct normalization reasoning. We aggregate the three axes into a unified {\DNS}. The full prompt for \hbox{LLM-as-a-Judge} is provided in Appendix~\ref{appendix:prompt_examples}.

\paragraph{Criterion 1. Semantic Evaluation}
The semantic axis evaluates whether a generated decomposition satisfies the lossless-join property. We formally verify this using the chase algorithm over the functional dependencies. A decomposition is accepted only if the chase derives a row of all distinguished variables. This ensures that the natural join of the generated relations reconstructs the original table without spurious tuples. Any decomposition that drops an attribute of the original schema automatically fails this check. Because a lossless join is a strict requirement for valid normalization, the semantic score acts as a binary gate in {\DNS}, zeroing out the overall score upon failure.

\paragraph{Criterion 2. Structural Evaluation}
The structural axis measures how well the generated DDL recovers the schema elements of the gold decomposition. We parse the \hbox{LLM-generated} DDL into schema objects and compute three metrics:
\squishlist
\item \textbf{Column F1}:
F1 between the generated and gold column sets.
\item \textbf{Primary Key F1}:
F1 between the generated and gold primary-key sets.
\item \textbf{Foreign Key Score}: The product of FK matching and FK validity:
\begingroup
\setlength{\leftmarginii}{1.2em}
\squishlist
\item[$\circ$] \textbf{FK F1}:
F1 between the generated and gold foreign-key sets.
\item[$\circ$] \textbf{FK Connected Score}:
A binary indicator of whether generated FK constraints validly reference \hbox{extant} \hbox{tables} and columns.
\squishend
\endgroup
\squishend
The final structural score is the mean of Column F1, Primary Key F1, and Foreign Key Score.

\paragraph{Criterion 3. Logical Evaluation}
\label{para:logical_eval}
The logical axis evaluates whether the model correctly diagnoses the normalization problem and justifies its decomposition. It combines two components: (i)~an LLM-as-a-Judge~\cite{zheng2023judging} for the generated schema and explanation, and (ii) Violation F1 for the detected NF violation labels.

The judge LLM rates each model output along three dimensions:
\squishlist
\item \textbf{Logical Coherence}:
whether the generated decomposition is consistent with the intended normalization logic and gold DDL.
\item \textbf{Explanation of Schema Alignment}:
whether the explanation is internally consistent with the generated DDL and declared violation labels.
\item \textbf{Explanation Quality}:
whether the explanation uses normalization concepts correctly and presents the reasoning clearly and concisely.
\squishend

We use \texttt{gpt-oss-20b}~\cite{openai2025gptoss120bgptoss20bmodel} as the judge model. We select this model because the Judge's Verdict benchmark identifies it as a human-like judge~\citep{han2025judgesverdictcomprehensiveanalysis}. Specifically, it demonstrates strong agreement with human annotators without exceeding the natural range of human judgment variation. 
We further validate the judge on our task through an expert-agreement study, with details provided in Appendix~\ref{appendix:judge_validation}.

\textbf{Violation F1} compares the violation labels detected by the model against the gold violation labels. For each input table, we compute F1 over violation types, measuring whether the model correctly identifies which violations are present. 

The final logical score is the mean of the average judge score and the Violation F1.

\paragraph{{\DNS}.}

We aggregate the aforementioned three axes as \hbox{follows}:
\begin{equation*}
  \text{{\DNS}} = \text{Semantic} \times \sqrt{\text{Structural} \times \text{Logical}}.
\end{equation*}

Inspired by BLEU~\citep{papineni-etal-2002-bleu}, {\DNS} uses the geometric mean of the structural and logical scores when the semantic score is $1$. This aggregation rewards schemas that satisfy both complementary criteria, preserves the original score scale, and penalizes imbalance more strongly than an arithmetic mean.

\paragraph{Dataset Verification.}

We verify our benchmark dataset quality by scoring its gold decompositions, explanations, and labels using {\DNS}. The resulting score reaches $0.99$, establishing a performance ceiling near the theoretical maximum of $1$ for {\DBN}. As detailed in Appendix~\ref{appendix:gold_score}, this serves as a strong validation of both the gold labels and the evaluator, indicating that all models should be benchmarked against $0.99$ rather than $1$.


\section{Database Normalization Analysis~\label{sec:llm_normal_Anal}}
We evaluate LLMs' database normalization capability using our benchmark dataset and evaluation protocol. Our experiments are designed to measure two core capabilities required for database normalization: (i) identifying and reasoning over functional dependencies (FDs), and (ii) resolving multiple coexisting violations in the same schema.

\subsection{Experimental Setup\label{sec:experiment-construction}}
We use the {\DBN} test split generated by the pipeline in Section~\ref{sec:dataset}. Each sample contains an input table, row samples, gold violation labels, gold decompositions, and schema constraints. We test four representative open-weight LLMs spanning dense and sparse mixture-of-experts (MoE) architectures across different scales. Each setting is evaluated under both \hbox{zero-shot} and few-shot prompting, yielding six configurations. More details on models and prompts are provided in Appendices~\ref {appendix:model_selection} and~\ref{appendix:prompt_examples}.

We design three experimental settings for a \hbox{comprehensive} evaluation under different normalization target scopes and FD availability, as summarized in Table~\ref{tab:experiment_core_settings}.

\subsubsection*{Experiment~1: Single Reasoning}
\label{sec:exp1}
Experiment~1 tests whether a model can resolve the earliest violated normal form given explicit FD evidence. Although an input may contain a chain of violations, the model is asked to fix only the first violation: for example, 1NF in a 1NF--2NF--3NF--BCNF chain, or 2NF in a 2NF--3NF--BCNF chain. This setting isolates the model's ability to apply a single normalization rule.

\begin{table}[!tp]
  \centering
  \footnotesize
  \renewcommand{\arraystretch}{1}
  \setlength{\tabcolsep}{3pt}
  \begin{tabular*}{\columnwidth}{@{\extracolsep{\fill}} l c c}
    \toprule
    \textbf{Experiment} & \textbf{Target Scope} & \textbf{FDs} \\
    \midrule
    Exp.~1: Single & Earliest violation only & Provided \\
    Exp.~2: Complex & All violations & Provided \\
    Exp.~3: FD-withheld & All violations & Not provided \\
    \bottomrule
  \end{tabular*}
  \vspace{-1.5ex}
  \caption{Comparison of the three experimental settings.}
  \label{tab:experiment_core_settings}
  \vspace{-3ex}
\end{table}

\subsubsection*{Experiment~2: Complex Reasoning}
\label{sec:exp2}
Experiment~2 extends the task to the full violation chain while still providing explicit FD evidence. The model must identify all relevant violations and produce a valid multi-step decomposition. This setting evaluates whether the model can handle interactions among multiple normalization rules.

\subsubsection*{Experiment~3: FD-withheld Reasoning}

\label{sec:exp3}
Experiment~3 removes explicit FDs and requires the model to infer them from context. The input includes a table, row samples, and natural-language business rules, but it withholds formal FD annotations. The model must infer latent FDs, identify the full set of violations, and generate a normalized schema. This setting reflects practical normalization scenarios in which dependency information is not directly provided.

\begin{table*}[!tp]
  \centering
  \footnotesize
  \renewcommand{\arraystretch}{0.95}
  \setlength{\tabcolsep}{3pt}
  \newcommand{\pd}{\phantom{$^{\dagger}$}}
  \begin{tabular*}{\textwidth}{@{\extracolsep{\fill}} l ccc ccc ccc}
    \toprule
    \multirow{2}{*}[-0.3em]{\textbf{Model}}
    & \multicolumn{3}{c}{\textbf{Single}}
    & \multicolumn{3}{c}{\textbf{Complex}}
    & \multicolumn{3}{c}{\textbf{FD-withheld}} \\
    \cmidrule(lr){2-4} \cmidrule(lr){5-7} \cmidrule(lr){8-10}
    & Zero-shot & Few-shot & Avg
    & Zero-shot & Few-shot & Avg
    & Zero-shot & Few-shot & Avg \\
    \midrule
    Gemma-3-27B-it                & 0.354\pd              & 0.378\pd              & 0.366\pd              & 0.362$^{\dagger}$     & 0.444$^{\dagger}$     & 0.403$^{\dagger}$     & 0.163\pd              & 0.254$^{\dagger}$     & 0.208\pd              \\
    Qwen3-30B-A3B-Instruct-2507   & \textbf{0.491}\pd     & 0.406$^{\dagger}$     & \textbf{0.449}\pd     & 0.349\pd              & 0.379\pd              & 0.364\pd              & \textbf{0.248}\pd     & \textbf{0.293}\pd     & \textbf{0.270}\pd     \\
    Mixtral-8x7B-Instruct-v0.1    & 0.135\pd              & 0.172\pd              & 0.153\pd              & 0.146\pd              & 0.177\pd              & 0.161\pd              & 0.116\pd              & 0.112\pd              & 0.114\pd              \\
    Llama-3.3-70B-Instruct        & 0.372$^{\dagger}$     & \textbf{0.419}\pd     & 0.396$^{\dagger}$     & \textbf{0.387}\pd     & \textbf{0.458}\pd     & \textbf{0.422}\pd     & 0.207$^{\dagger}$     & 0.247\pd              & 0.227$^{\dagger}$     \\
    \arrayrulecolor{gray!50}\midrule\arrayrulecolor{black}
    Overall                           & 0.338\pd              & 0.344\pd              & 0.341\pd              & 0.311\pd              & 0.365\pd              & 0.338\pd              & 0.184\pd              & 0.226\pd              & 0.205\pd              \\
    \bottomrule
  \end{tabular*}
  \vspace{-1ex}
  \caption{Average {\DNS} ($\uparrow$) on the {\DBN} test split by model and evaluation settings. \textbf{Bold} and $\dagger$ mark the best and second-best scores in each column. Overall row averages the four models evaluated in every setting.}
  \label{tab:dbn_results}
  \vspace{-1ex}
\end{table*}

\begin{table*}[!tp]
  \centering
  \footnotesize
  \renewcommand{\arraystretch}{1}
  \setlength{\tabcolsep}{4pt}
  \newcommand{\pd}{\phantom{$^{\dagger}$}}
  \begin{tabular*}{\textwidth}{@{\extracolsep{\fill}} l c ccc cc}
    \toprule
    \multirow{2}{*}[-0.3em]{\textbf{Model}}
    & \textbf{Semantic}
    & \multicolumn{3}{c}{\textbf{Structural}}
    & \multicolumn{2}{c}{\textbf{Logical}} \\
    \cmidrule(lr){2-2} \cmidrule(lr){3-5} \cmidrule(lr){6-7}
    & lossless join
    & Column F1 & PK F1 & FK Score
    & Violation F1 & LLM Judge \\
    \midrule
    Gemma-3-27B-it                & \textbf{0.524}\pd     & \textbf{0.969}\pd     & 0.656\pd              & 0.183\pd              & 0.376\pd              & 0.225\pd              \\
    Qwen3-30B-A3B-Instruct-2507   & 0.508$^{\dagger}$     & 0.962$^{\dagger}$     & 0.666$^{\dagger}$     & \textbf{0.254}\pd     & \textbf{0.684}\pd     & \textbf{0.262}\pd     \\
    Mixtral-8x7B-Instruct-v0.1    & 0.306\pd              & 0.811\pd              & 0.542\pd              & 0.099\pd              & 0.503\pd              & 0.165\pd              \\
    Llama-3.3-70B-Instruct        & 0.454\pd              & 0.951\pd              & \textbf{0.671}\pd     & 0.241$^{\dagger}$     & 0.584$^{\dagger}$     & 0.254$^{\dagger}$     \\
    \arrayrulecolor{gray!50}\midrule\arrayrulecolor{black}
    Overall                       & 0.448\pd              & 0.923\pd              & 0.634\pd              & 0.194\pd              & 0.537\pd              & 0.227\pd              \\
    \bottomrule
  \end{tabular*}
  \vspace{-1ex}
  \caption{Per-model {\DNS} component breakdown under the \emph{FD-withheld} setting, averaged over the {\DBN} test split with the zero- and few-shot runs pooled. \textbf{Bold} indicates the best score per column, and $\dagger$ the second-best.}
  \label{tab:dbn_components_breakdown}
  \vspace{-2ex}
\end{table*}

\subsection{Baseline Results and Analysis}
Our baseline evaluation yields three main findings regarding LLM-based relational database \hbox{normalization}.

\paragraph{Finding 1. FD extraction is the main bottleneck in \hbox{LLM-based} database normalization.}
As Table~\ref{tab:dbn_results} shows, all four models degrade most sharply in the \emph{FD-withheld} setting, where FDs are not explicitly provided. Averaged across models, {\DNS} drops modestly from $0.341$ in \emph{Single} to $0.338$ in \emph{Complex}, but declines more sharply to $0.205$ in \emph{FD-withheld}. 
This pattern indicates that resolving multiple violations is not the primary bottleneck. Instead, the governing factor for normalization performance is the presence of explicit FDs, as models struggle to infer FDs solely from data samples and natural-language business rules.

\paragraph{Finding 2. LLMs perform better at violation detection than at schema reconstruction.}
When performing normalization, LLMs struggle far more to produce valid global schema structures than to recognize violations. Table~\ref{tab:dbn_components_breakdown} reports the component scores. Violation F1 reaches up to $0.684$, indicating that models can reliably identify normal-form violations. Models preserve local attributes well, achieving an average Column F1 of $0.92$ across four models. However, their performance drops sharply under inter-table constraints, with FK Scores ranging from $0.10$ to $0.25$. LLMs excel at local violation diagnosis but struggle to reconstruct global schemas with valid PK and FK structures.

\vspace{-0.5ex}
\paragraph{Finding 3. LLMs fail at BCNF reasoning and over-normalize valid schemas.}
Models exhibit their weakest performance on BCNF reasoning and frequently over-normalize already valid schemas. Our per-violation analysis, reported in Table~\ref{tab:realworld_by_violation} of Appendix~\ref{appendix:extended_results}, highlights these weaknesses. Although models face difficulties across all normal form violations, BCNF reasoning is the hardest, with an average score of $0.110$ across all four models. \hbox{Interestingly,} even for NONE cases, the average score across the four models reaches only $0.440$. This reveals a strong tendency among models to decompose schemas unconditionally, rather than only when necessary. As reported in Appendix~\ref{appendix:over_normalization}, they introduce unnecessary decompositions in over half of these valid cases.

Overall, current LLMs face three critical challenges in database normalization: (i)~\emph{unreliable FD inference in realistic settings}, (ii)~\emph{weak reconstruction of valid inter-table constraints}, and (iii)~\emph{a tendency to over-normalize already valid schemas}. These findings motivate a structured normalization framework that separates reasoning stages and incorporates explicit verification.

\begin{figure*}[!t]
  \centering

  \includegraphics[width=\textwidth]{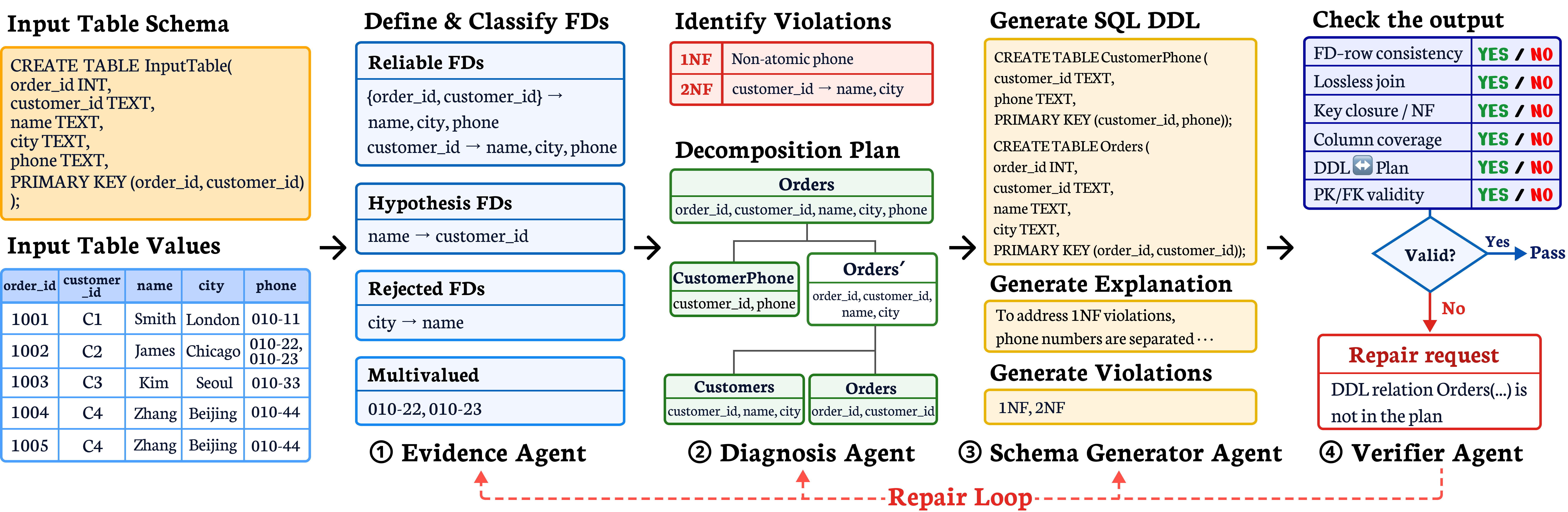}
  \caption{Overview of MARS framework---\emph{Evidence}, \emph{Diagnosis}, \emph{Schema Generator}, and \emph{Verifier} agents---together with the \emph{Repair loop} triggered by verification failures.}
  \label{fig:multi_agent}
  \vspace{-2ex}
\end{figure*}

\section{Multi-Agent Reasoning for Schemas}
\label{sec:multi-agent}



Multi-agent frameworks address this issue by dividing tasks among role-specialized agents. 
There are notable instances of such success, including conversational orchestration in AutoGen~\cite{wu2024autogen}, collaborative software development in MetaGPT~\cite{hong2024metagpt}, and automated coding workflows in ChatDev~\cite{qian-etal-2024-chatdev}.  Motivated by these approaches, we adapt role specialization and verifier feedback to database normalization. In our task, each step satisfies dependency, foreign key, and lossless-join requirements.

\subsection{MARS Framework}
We propose Multi-Agent Reasoning for Schemas (MARS), which decomposes LLM-based database normalization into four stages: (i)~\emph{evidence extraction}, (ii)~\emph{violation diagnosis and decomposition planning}, (iii)~\emph{schema generation}, and~(iv)~\emph{deterministic verification}. Figure~\ref{fig:multi_agent} presents an overview of the MARS framework. The first three stages are LLM calls, and the fourth is rule-based. 
Their prompt templates are provided in Appendix~\ref{appendix:mars_prompts}.

\paragraph{Evidence Agent.}
The Evidence Agent collects FD evidence from the input schema, sample rows, and optional FD annotations. It either adopts the provided FDs directly or infers candidates from the data, sorts them into reliable, hypothetical, and rejected sets, and flags \hbox{multivalued} columns as evidence of 1NF violations.

\vspace{-0.5ex}
\paragraph{Diagnosis Agent.}
Using evidence from the \hbox{Evidence} Agent, the Diagnosis Agent identifies violated normal forms and responsible attributes or FDs, prioritizing reliable FDs over hypothetical ones. It then plans a stepwise decomposition that specifies the driving FD or attribute for each step, along with the resulting relations and keys.

\vspace{-0.5ex}
\paragraph{Schema Generator Agent.}
The Schema Generator Agent generates the planned relations as SQL DDL with PK/FK constraints, along with violation types and explanations.

\vspace{-0.5ex}
\paragraph{Verifier Agent.}
The Verifier Agent deterministically evaluates predictions against the model's own extracted FDs rather than gold dependencies. Specifically, it checks extracted FDs against sample rows, diagnosed violations against the FD closure, the plan for column coverage, per-relation normal forms, lossless join, and multivalued-attribute handling, as well as the DDL for parsing, plan consistency, and PK/FK validity. It reports failures as objective facts without prescribing explicit fixes.  

\vspace{-0.5ex}
\paragraph{Repair Loop.}
When verification fails, MARS performs up to two repair rounds, restarting from the agent whose output failed first. FD conflicts with sample rows return to the Evidence Agent, diagnosis and plan errors to the Diagnosis Agent, and DDL errors to the Schema Generator Agent. The restarted agent receives its prior output and the verifier's findings, and all downstream agents re-run.

\begin{table*}[!t]
  \centering
  \footnotesize
  \renewcommand{\arraystretch}{1}
  \setlength{\tabcolsep}{4pt}
  \newcommand{\pd}{\phantom{$^{\dagger}$}}
  \begin{tabular*}{\textwidth}{@{\extracolsep{\fill}} l l c ccc cc c}
    \toprule
    \multirow{2}{*}[-0.3em]{\textbf{Method}}
    & \multirow{2}{*}[-0.3em]{\textbf{Prompting}}
    & \textbf{Semantic}
    & \multicolumn{3}{c}{\textbf{Structural}}
    & \multicolumn{2}{c}{\textbf{Logical}}
    & \multirow{2}{*}[-0.3em]{\textbf{{\DNS}}} \\
    \cmidrule(lr){3-3} \cmidrule(lr){4-6} \cmidrule(lr){7-8}
    & & lossless join
    & Column F1 & PK F1 & FK Score
    & Violation F1 & LLM Judge
    & \\
    \midrule
    \multirow{2}{*}{Baseline}
    & Zero-shot & 0.482\pd & 0.945\pd & 0.660\pd & 0.223\pd & 0.647\pd & 0.263\pd & 0.248\pd \\
    & Few-shot  & 0.534\pd & 0.978\pd & 0.673\pd & 0.285\pd & \textbf{0.722}\pd & 0.261\pd & 0.293\pd \\
    \midrule
    \multirow{2}{*}{Miffie}
    & Zero-shot & 0.554\pd & 0.953\pd & 0.662\pd & 0.226\pd & 0.650\pd & 0.262\pd & 0.288\pd \\
    & Few-shot  & 0.633\pd & 0.968\pd & 0.661\pd & 0.283\pd & 0.682$^{\dagger}$ & 0.362\pd & 0.381\pd \\
    \midrule
    \multirow{2}{*}{\textbf{MARS} (Ours)}
    & Zero-shot & \textbf{0.891}\pd & \textbf{0.983}\pd & \textbf{0.761}\pd & 0.300$^{\dagger}$ & 0.596\pd & 0.401$^{\dagger}$ & \textbf{0.499}\pd \\
    & Few-shot  & 0.868$^{\dagger}$ & 0.981$^{\dagger}$ & 0.753$^{\dagger}$ & \textbf{0.303}\pd & 0.606\pd & \textbf{0.419}\pd & 0.496$^{\dagger}$ \\
    \bottomrule
  \end{tabular*}
  \vspace{-1ex}
  \caption{Component-level performance breakdown for the \emph{FD-withheld} setting. All scores are computed per test sample and averaged over the {\DBN} dataset. All methods use Qwen3-30B-A3B-Instruct-2507 as the backbone. \textbf{Bold} marks the maximum per column among reported scores, and $\dagger$ marks the second-best.}
  \label{tab:multi-agent_dbn_components_breakdown}
  \vspace{-2ex}
\end{table*}


\subsection{Experiment Setup}
We evaluate all methods in the \emph{FD-withheld} setting, which most closely aligns with practical database normalization, where models must infer FDs from the schema context and business rules rather than relying on given FDs. All methods use Qwen3-30B-A3B-Instruct-2507 as the backbone, which achieved the highest average {\DNS} in this setting among the base LLMs in Table~\ref{tab:dbn_results}. Sharing a single backbone ensures that performance differences reflect the framework design rather than the underlying LLM.

\textbf{Baseline} is Qwen3-30B-A3B-Instruct-2507 configuration from Table~\ref{tab:dbn_results} under \hbox{\emph{FD-withheld}} setting. 

\textbf{Miffie} is the Dual-LLM self-refinement framework proposed by \citet{jo2026databasenormalizationdualllmselfrefinement}. Following the original setup, we run up to three refinement \hbox{iterations.} Although Miffie was designed for \hbox{1NF-to-3NF} normalization, we evaluate it on the full {\DBN} \emph{FD-withheld} setting, including BCNF cases, as an LLM-based refinement baseline. Implementation details are provided in Appendix~\ref{appendix:miffie_reimpl}.

\textbf{MARS} follows Section~\ref{sec:multi-agent}, with one initial generation and up to two repair rounds. This caps it at three generation attempts, matching Miffie's maximum number of refinement iterations.

\subsection{Experiment Results}

\textbf{MARS achieves the highest performance in the \emph{FD-withheld} setting.}
Such a performance demonstrates two key insights: first, that role specialization significantly outperforms self-refinement, and second, that simple generate-and-verify refinement is fundamentally insufficient to improve the schema structure. 
This performance gap can be attributed to the effectiveness of each method in refining the schema structure. For Miffie, metrics such as \hbox{Column} F1, PK F1, and FK Score remain within $0.04$ of the baseline. As shown in \hbox{Table~\ref{tab:miffie_iteration_loop}} in Appendix~\ref{appendix:miffie_artifact_analysis}, its verifier struggles to identify high-quality schemas. This suggests that simple generate-and-verify \hbox{refinement} is insufficient to \hbox{improve} the schema structure. In contrast, MARS effectively improves both semantic and structural components. It raises lossless join to $0.891$. It also achieves the highest FK Score among all methods. Therefore, MARS does not just preserve the original data. Its {\DNS} improvement is driven by the superior generation of valid normalized DDL.

\begin{figure}[!t]
  \centering
  \includegraphics[width=\columnwidth]{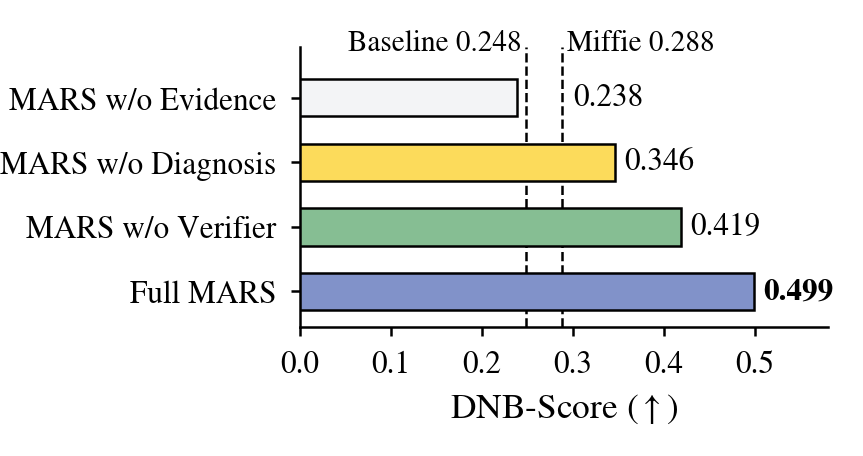}
  \vspace{-4ex}
  \caption{MARS agent-removal ablation under the \hbox{\emph{FD-withheld}} setting with zero-shot prompting, using Qwen3-30B-A3B-Instruct-2507 as the backbone.}
  \label{fig:mars_ablation}
  \vspace{-3ex}
\end{figure}

\textbf{FD evidence extraction is the most critical component, serving as MARS's primary performance bottleneck.}
As shown in Figure~\ref{fig:mars_ablation}, while ablating any agent degrades overall performance, the Evidence Agent emerges as the most critical bottleneck. Removing the Evidence Agent causes the largest drop, to $0.238$, below both the single-prompt Baseline and Miffie. 
Without the Diagnosis Agent or Verifier, MARS scores $0.346$ and $0.419$, respectively---below full MARS but above \hbox{zero-shot} Miffie.
Table~\ref{tab:agent_stage_diagnostics} in Appendix~\ref{appendix:agent_artifact_analysis} relates these results to underlying stage-wise artifacts. Schema generation reliably aligns with diagnosis plans at a rate of $0.97$, so MARS succeeds consistently when provided with a plausible plan. The primary bottlenecks lie upstream in FD evidence extraction and diagnosis, with alignment rates of $0.65$ and $0.60$, respectively. These upstream errors subsequently propagate into key selection and FK reconstruction.

\textbf{The performance gains of MARS are robust, scale-monotonic, and architecturally driven.}
\hbox{Figure~\ref{fig:run_variance}} in Appendix~\ref{appendix:run_variance} details our three repeated runs. MARS exhibits high stability with a coefficient of variation of just $1.68\%$. This is well below the Baseline's $3.20\%$. This low variance confirms that the observed gains are statistically significant. Beyond stability across runs, MARS also restores monotonic scaling across Gemma3 4B, 12B, and 27B, as shown in Figure~\ref{fig:scale_controlled} in Appendix~\ref{appendix:scale_controlled}. Its score rises steadily from $0.305$ to $0.378$, ultimately reaching $0.454$. In contrast, the single-prompt Baseline peaks early at 12B. Regarding inference costs, Figure~\ref{fig:inference_cost} in Appendix~\ref{appendix:inference_cost} shows that MARS requires about 5 model calls per sample, compared to the Baseline's 1. However, this extra compute translates into effective reasoning. MARS in the zero-shot setting actually consumes fewer tokens than Miffie in the few-shot setting. Despite using fewer tokens, it substantially outperforms Miffie, scoring $0.499$ against $0.381$.

\section{Conclusion}
We introduced {\DBN}, a benchmark for assessing LLM-driven database normalization using controlled denormalization data and a three-axis protocol that measures semantic, structural, and logical validity. We demonstrated that LLMs struggle to generate \hbox{information-preserving} schemas with valid \hbox{inter-table} constraints, especially when functional dependencies must be \hbox{inferred.} 

To address these limitations, we additionally proposed MARS, a multi-agent framework that decouples normalization tasks across agents.
MARS outperforms the single-prompt Baseline and Miffie, demonstrating the \hbox{significance} of decomposing the normalization process. We believe {\DBN} and MARS lay the groundwork for future research on \hbox{LLM-driven} database schema design.

\section*{Limitations}


\paragraph{FK reconstruction remains a bottleneck.}
Across every evaluated configuration, foreign-key reconstruction remains a severe bottleneck. Under the \emph{FD-withheld} setting, the single-prompt Baseline achieves lower FK scores. This pattern of low performance persists across more specialized frameworks such as Miffie and MARS. Although MARS achieves the highest FK Score at $0.303$, this absolute value remains critically low. The underlying challenge is that reconstructing the reference structure across decomposed relations is inherently complex. FK reconstruction requires consistent relation boundaries, key choices, and reference targets simultaneously. Consequently, upstream FD or diagnosis errors easily cascade through this process. Recovering valid inter-table constraints after decomposition therefore remains a major open problem.


\paragraph{MARS trades over-normalization for under-detection.}
Averaged over zero- and few-shot prompting, the single-prompt Baseline decomposes $57.0\%$ of already-normalized inputs. MARS almost eliminates this behavior, decomposing only $4.7\%$ of already-normalized inputs, \hbox{because} it decomposes only when its extracted dependencies support a violation. The same conservatism produces the opposite failure, causing MARS to leave $37.1\%$ of genuinely violating inputs unchanged. Deciding whether a decomposition is needed, rather than carrying one out, is thus the dominant remaining error mode.

\section*{Ethical Considerations}
All models in this work are open-weight LLMs used for inference only, and all source data come from the public Spider (CC BY-SA 4.0) and BIRD (CC BY-NC 4.0) benchmarks, which we use for non-commercial research with attribution. DNBench inherits these licenses and will be released under compatible terms. We introduce no new personal data, as all samples are derived from existing benchmark databases. Three database experts voluntarily participated in sample auditing and the judge-agreement study after being informed of the study's purpose. Finally, LLM-generated schemas can silently lose information when lossless join fails, so we recommend that automated normalization be paired with deterministic verification and human review before deployment.

\section*{Use of AI Assistants.}
We used AI assistants solely for language editing, including improving grammar, readability, and conciseness of author-written text, and for formatting bibliographic entries. All research ideas, experimental design, implementation, analyses, and conclusions are our own. The authors reviewed all suggested edits and take full responsibility for the content of this paper.


\clearpage
\bibliography{anthology-1,anthology-2,custom}
\clearpage
\appendix

\section{Dataset Details}
\label{appendix:dataset_details}

This appendix expands on the dataset construction in Section~\ref{sec:dataset}. We describe the Spider and BIRD source datasets, explain why they are suitable source schemas for controlled denormalization, and summarize the expert validation of the generated {\DBN} samples.

\subsection{Source Datasets}

\paragraph{Spider.} Spider~\cite{yu-etal-2018-spider} is a large-scale cross-domain Text-to-SQL dataset built by a research team at Yale University. It consists of 10,181 natural-language queries and 5,693 unique SQL queries, drawn from 200 databases across 138 domains. Each database contains an average of 5.1 tables and foreign-key relationships, covering SQL patterns from simple single-table lookups to nested subqueries and multiple joins.

\paragraph{BIRD.} BIRD~\cite{li2023can} is a benchmark for evaluating Text-to-SQL performance in large-scale, real-world database environments. It consists of 95 databases, 33.4 GB of data, 37 specialized domains, and 12,751 query--SQL pairs.

\paragraph{Dataset Licenses.} The Spider benchmark is distributed under the Creative Commons Attribution-ShareAlike 4.0 International (CC BY-SA 4.0) license. The BIRD benchmark is distributed under the Creative Commons Attribution-NonCommercial 4.0 International (CC BY-NC 4.0) license. Both datasets are used in this work for non-commercial academic research purposes with appropriate attribution.

\begin{algorithm}[!t]
\footnotesize
\caption{\textsc{Structure-Preserving Relational Sampling}}
\label{alg:sampling}
\KwIn{Relational database $D$; max tuples per base table $n$; max rows per table $c$}
\KwOut{Sampled database $D'$}

Extract schema metadata from $D$ and obtain table set $\mathcal{T}$, primary keys, and foreign-key (FK) relations\;
Build the FK graph over $\mathcal{T}$ (edge $u\!\to\!v$ iff $u$ has an FK referencing $v$)\;
Identify base tables $\mathcal{B}\subseteq\mathcal{T}$ (nodes with no outgoing FK edges)\;
Initialize sampled-relation set $\mathcal{R}\leftarrow\emptyset$ and global PK registry $\mathcal{K}\leftarrow\{(t,\emptyset):t\in\mathcal{T}\}$\;

\For{each base table $b \in \mathcal{B}$}{
    Construct join candidates by traversing FK chains from $b$ to related tables\;
    Remove candidates containing null primary keys\;
    Select up to $n$ tuples from $b$ based on its primary key\;
    Initialize local PK registry $\mathcal{K}_b$ from the selected tuples\;
    Expand $\mathcal{K}_b$ via FK/self-FK closure until convergence\;
    Retrieve tuples from related tables using the converged key sets\;
    Merge the retrieved tuples into $\mathcal{R}$ with PK-based deduplication\;
    Update $\mathcal{K}$ using $\mathcal{K}_b$\;
}

\If{some table in $\mathcal{R}$ exceeds $c$ rows}{
    Apply FK-aware global capping while preserving referential integrity\;
}

Create a new SQLite database $D'$ and insert tables in topological order\;
\While{FK violations are detected in $D'$}{
    Remove violating tuples by recursive cascade cleanup\;
}
Remove empty tables from $D'$\;
\Return $D'$
\end{algorithm}

\subsection{Selection Rationale}

We chose Spider and BIRD for three reasons.

First, both datasets provide relational schemas with explicit primary-key and foreign-key annotations. These constraints are essential for FK-aware subsampling, controlled violation injection, and structural evaluation of generated DDL.

Second, both datasets cover diverse domains and schema structures. This allows {\DBN} to evaluate normalization behavior across a wide range of table layouts, key structures, and attribute combinations.

Third, their relational structure makes them suitable source schemas for controlled denormalization. Starting from realistic schemas allows us to generate denormalized samples with known violation patterns while retaining a database-like \hbox{structure.}

\subsection{Structure-Preserving Relational Sampling}
\label{appendix:sampling}
This subsection details Stage~1 (FK-aware Subsampling) of the dataset construction pipeline in Section~\ref{sec:dataset}. The goal of this stage is to downsample each large source database into a compact instance that remains a valid relational database---that is, one that preserves primary-key uniqueness and foreign-key referential integrity---so that the subsequent violation-injection stages operate on realistic, self-consistent schemas.

Algorithm~\ref{alg:sampling} formalizes this procedure. Starting from the schema metadata, we build a foreign-key graph over the tables and identify \emph{base tables} (nodes with no outgoing FK edges). For each base table, we select up to $n$ tuples and expand the selected key sets through FK and self-FK closures until convergence, ensuring that every referenced tuple is also retrieved. After merging the retrieved tuples with primary-key-based deduplication, we apply a global per-table row cap and insert the tables into a new SQLite database in topological order. Finally, any residual FK violations are removed through recursive cascade cleanup, and empty tables are dropped. The result $D'$ is a structurally faithful subsample on which controlled denormalization can be applied without introducing spurious integrity errors.

\subsection{Expert Validation of {\DBN} Samples}

After constructing the {\DBN} dataset, three database experts reviewed the generated {\DBN} samples. The review focused on whether the denormalized tables contain the 1NF--to--BCNF violation patterns and whether the chain-rule labels, gold decompositions, and expected schema constraints are consistent with the construction rules. Additionally, the experts confirmed that the NONE samples are fully compliant and exhibit no normal-form violations. This validation concerns the {\DBN} samples generated for our experiments.

\subsection{{\DBN} Sample Configuration}
\label{appendix: Violation Categories by Normal-Form}

The {\DBN} benchmark comprises samples constructed across five distinct structural categories. Each category represents a specific pattern of normal-form violations or a fully compliant state, defined as follows:

\squishlist
\item 1\_2\_3\_BCNF: violates 1NF, 2NF, 3NF, and BCNF;
\item 2\_3\_BCNF: violates 2NF, 3NF, and BCNF;
\item 3\_BCNF: violates 3NF and BCNF;
\item BCNF: violates BCNF only;
\item NONE: already normalized, with no violations to resolve.
\noindent
\squishend

The dev split includes 299 Spider and 99 BIRD samples per category, and the test split includes 180 Spider and 77 BIRD samples per category.
\section{Model Selection}
\label{appendix:model_selection}

We restrict our evaluation to four open-weight language models widely adopted in the contemporary LLM ecosystem, which together cover the dominant architectural paradigms in modern LLM design. By choosing models that are openly released, reproducible, and actively used across research and industry, we ensure that the failure modes uncovered by {\DBN} are not artifacts of a single proprietary system but reflect properties of the broader class of modern LLMs. The four models span two complementary axes that are known to influence reasoning behavior, namely model scale (ranging from $27$B to $70$B total parameters) and architectural family (dense decoder-only models against sparse mixture-of-experts models), allowing the benchmark to probe whether normalization difficulty correlates with capacity, sparsity, or instruction-tuning recipe. In addition to these four evaluated models, we use a separate open-weight model as the LLM judge and two smaller members of the Gemma~3 family for the scale-controlled analysis in Appendix~\ref{appendix:scale_controlled}.
\vspace{2mm}
\begin{itemize}
    \item \textbf{Gemma-3-27B-it}~\cite{gemmateam2025gemma3technicalreport} is Google's open-weight dense model designed for efficient deployment at moderate scale. It represents the class of mid-sized dense models that trade raw capacity for inference cost while retaining competitive instruction-following ability. Its smaller siblings, \textbf{Gemma-3-4B-it} and \textbf{Gemma-3-12B-it}, share the same architecture and post-training recipe and are used only in the scale-controlled comparison of Appendix~\ref{appendix:scale_controlled}, where parameter count is the sole varying factor.
    \item \textbf{Qwen3-30B-A3B-Instruct-2507}~\cite{yang2025qwen3technicalreport} is Alibaba's mixture-of-experts model from the Qwen3 family with $30.5$B total and $3.3$B activated parameters. By activating a subset of its experts per token, it isolates the effect of sparse expert routing on normalization reasoning while remaining comparable in active-parameter footprint to mid-sized dense \hbox{models.}
    \item \textbf{Mixtral-8x7B-Instruct-v0.1}~\cite{jiang2024mixtralexperts} is Mistral's sparse mixture-of-experts model that routes each token through two of eight $7$B experts. It is included as the most established open-weight MoE baseline and as a larger-scale counterpart to Qwen3-30B-A3B-Instruct-2507, enabling comparison across two MoE designs.
    \item \textbf{Llama-3.3-70B-Instruct}~\cite{grattafiori2024llama3herdmodels} is a $70$B-parameter dense decoder-only model released by Meta and serves as a strong reference point for instruction-tuned dense LLMs. Its broad availability and extensive evaluation across reasoning benchmarks make it the standard baseline for open-weight large-scale models.
    \item \textbf{gpt-oss-20b}~\cite{openai2025gptoss120bgptoss20bmodel} is \hbox{OpenAI's} open-weight mixture-of-experts model with $21$B total and $3.6$B activated parameters. It is not an evaluated model; we use it exclusively as the LLM judge for the logical axis evaluation. Choosing a judge from a different model family than any of the evaluated models avoids self-preference bias, and the judge's open weights keep the judging procedure reproducible.
\end{itemize}


\paragraph{Model Licenses.}
All pre-trained models are used in accordance with their respective licenses and employed as-is for evaluation only, without any modifications, fine-tuning, or parameter updates. 
For quantized variants, we use GGUF builds redistributed by Unsloth~\cite{unsloth} 
under the original model licenses, with Q4\_K\_S quantization applied only to the model weights.

\begin{itemize}
    \item \textbf{Gemma-3-27B-it}, \textbf{Gemma-3-12B-it}, and \textbf{Gemma-3-4B-it} are used under the Gemma Terms of Use.
    \vspace{-1mm}
    \item \textbf{Qwen3-30B-A3B-Instruct-2507} is used under the Apache License 2.0.
    \vspace{-1mm}
    \item \textbf{Mixtral-8x7B-Instruct-v0.1} is used under the Apache License 2.0.
    \vspace{-1mm}
    \item \textbf{Llama-3.3-70B-Instruct} is used under the Llama 3.3 Community License Agreement, \textcopyright\ Meta Platforms, Inc.
    \vspace{-1mm}
    \item \textbf{gpt-oss-20b}, used only as the judge model (Section~\ref{sec:three_axis_eval_protocol}), is released under the Apache License 2.0 and the gpt-oss usage policy.
\end{itemize}
\section{Experiment Reproducibility}~\label{app:reproducibility}
  All experiments were conducted as inference-only evaluations on a high-performance computing cluster. 
  The hardware and software configurations used for our empirical evaluations are as follows:
  \begin{itemize}
      \item \textbf{Hardware Resources.} We used an NVIDIA DGX-class system equipped with NVIDIA A100-SXM4-40GB GPUs. We used 1$\times$A100 for Gemma-3-27B-it, Qwen3-30B-A3B-Instruct-2507, Mixtral-8x7B-Instruct-v0.1, and the MARS agent (based on Qwen3-30B-A3B-Instruct-2507), and 2$\times$A100 for Llama-3.3-70B-Instruct.

      \item \textbf{Software Environment.} The system operated with CUDA 12.2 and NVIDIA Driver 535.161.08. Model inference was served through an OpenAI-compatible vLLM server, and all evaluation scripts were executed in the same software environment.

      
      \item \textbf{Results.} All experimental results, including MARS and Miffie, were described based on a single run.
  \end{itemize}
\section{Gold Decomposition Verification}
\label{appendix:gold_score}

\begin{table}[!t]
  \centering
  \footnotesize
  \setlength{\tabcolsep}{4pt}
  \begin{tabular*}{\columnwidth}{@{\extracolsep{\fill}} lccc}
    \toprule
    & \textbf{Spider} & \textbf{BIRD} & \textbf{All} \\
    \midrule
    \textbf{{\DNS}} & 0.99 & 0.99 & 0.99 \\
    \midrule
    \multicolumn{4}{c}{\textit{Components}} \\
    \midrule
    Lossless Join & 1.00 & 1.00 & 1.00 \\
    Column F1 & 1.00 & 1.00 & 1.00 \\
    PK F1 & 1.00 & 1.00 & 1.00 \\
    FK Score & 1.00 & 1.00 & 1.00 \\
    Violation F1 & 1.00 & 1.00 & 1.00 \\
    LLM-as-a-Judge & 0.97 & 0.97 & 0.97 \\
    \quad $\hookrightarrow$ Logical Coherence & 0.97 & 0.97 & 0.97 \\
    \quad $\hookrightarrow$ Expl.\ of Schema Align. & 0.95 & 0.95 & 0.95 \\
    \quad $\hookrightarrow$ Explanation Quality & 0.99 & 0.98 & 0.99 \\
    \bottomrule
  \end{tabular*}
  \caption{{\DNS} and its components obtained by scoring the gold \emph{combined} decompositions of the {\DBN} test split as model outputs.}
  \label{tab:gold_score}
\end{table}

\begin{figure*}[tp]
  \centering
  \includegraphics[width=\textwidth]{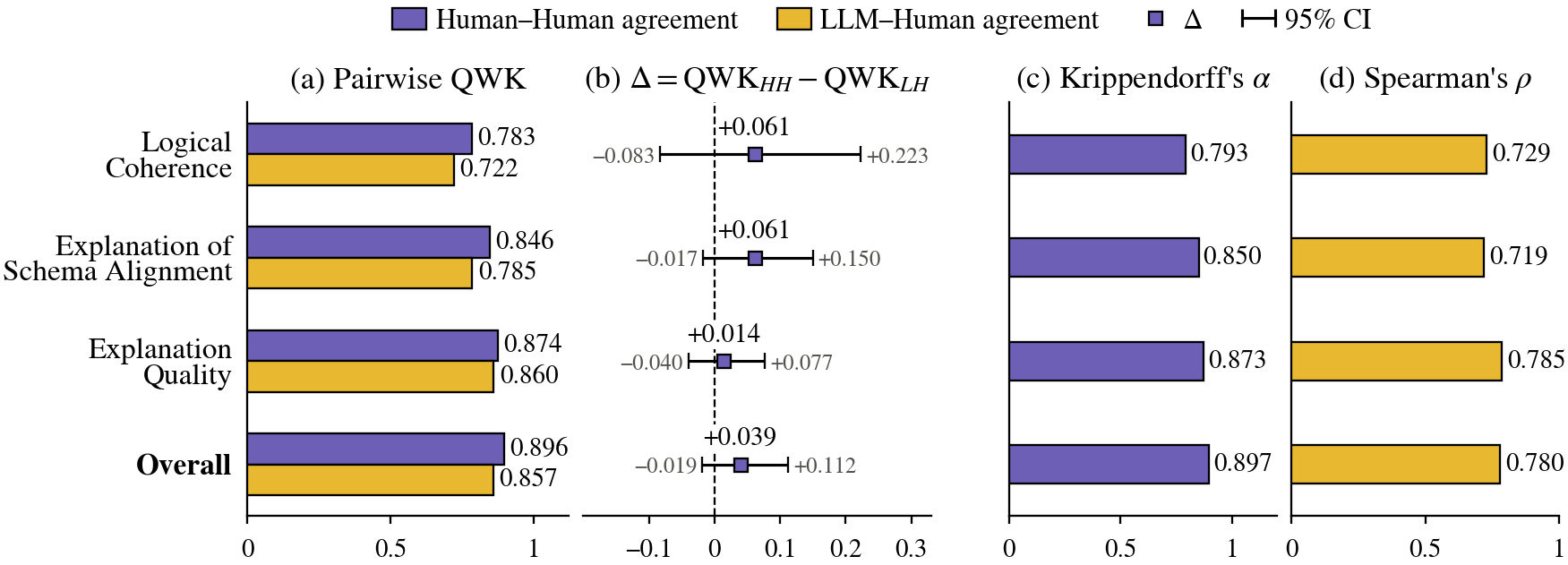}
  \vspace{-4ex}
  \caption{Human–Human and LLM–Human agreement statistics on each LLM-as-a-Judge dimension and the overall score. (a) Pairwise QWK of Human--Human and LLM--Human agreement. (b) $\Delta = \mathrm{QWK}_{HH} - \mathrm{QWK}_{LH}$ with its $95\%$ paired-bootstrap CI. (c) Krippendorff's $\alpha$ among the human experts. (d) Spearman's $\rho$ between the judge and the human mean.}
  \label{fig:judge_validation}
  \vspace{-2ex}
\end{figure*}

To establish the performance ceiling under our evaluator, we test the entire dataset test split. For each sample, we submit its gold target, which comprises the normalized DDL, rationale explanation, and violation labels, to the evaluator as a model prediction. Evaluation follows Section~\ref{sec:three_axis_eval_protocol} using the same judge model configuration as all other runs.


As presented in Table~\ref{tab:gold_score}, the gold decompositions achieve a {\DNS} of $0.99$ across both Spider and BIRD. All rule-based components, including the lossless-join check, Column F1, PK F1, FK Score, and Violation F1, reach a perfect score of $1.0$.  The minor gap to a perfect score stems almost entirely from the \hbox{LLM-as-a-Judge} component, which yields $0.97$. Crucially, our \hbox{human-agreement} validation, as shown in \hbox{Appendix~\ref{appendix:judge_validation}}, demonstrates that the LLM judge's evaluations are statistically indistinguishable from human expert judgments ($\mathrm{QWK} = 0.857$).




\section{Validation of LLM-as-a-Judge}
\label{appendix:judge_validation}


To validate the LLM-as-a-Judge, we conduct a human-agreement study. We measure whether the judge's agreement with humans is statistically indistinguishable from inter-human agreement.

We sample $50$ instances (10 databases from \hbox{Spider} and BIRD $\times$ 5 samples each) and ask three database experts to score them independently under the same rubric used by the judge. The LLM judge evaluates the same instances with identical prompts. Scores are assigned on a 0--5 scale normalized to $\{0, 0.2, 0.4, 0.6, 0.8, 1.0\}$. The overall score is the mean of the three LLM-as-a-Judge dimensions (Logical Coherence, Explanation of Schema Alignment, and Explanation Quality) defined in Section~\ref{sec:three_axis_eval_protocol}.

As in Figure~\ref{fig:judge_validation}, we evaluate agreement using four metrics. Quadratic weighted Cohen's $\kappa$, or QWK, measures pairwise agreement on the ordinal scale~\cite{cohen1968weighted}. Krippendorff's $\alpha$ with the interval metric captures agreement among three experts ~\cite{krippendorff2011computing}. Spearman's $\rho$ assesses rank agreement between the judge and the human mean. To evaluate significance, a paired bootstrap comparison computes $\Delta = \text{QWK}_{HH} - \text{QWK}_{LH}$ from $2{,}000$ resamples at a $95\%$ confidence interval~\cite{efron1994bootstrap}.

The empirical findings confirm strong alignment between the judge and human raters. On the overall score, the LLM--human QWK is $0.857$ against a human--human QWK of $0.896$, and the judge's ranking correlates with the human mean at $\rho = 0.780$. The experts themselves reach an interval-level $\alpha$ of $0.897$, confirming a reliable human reference. The paired bootstrap comparison gives $\Delta = +0.039$ with $95\%$ CI $[-0.019, +0.112]$. Because the interval contains $0$, LLM--human agreement is statistically indistinguishable from human--human agreement, supporting the LLM judge as a valid proxy across all three dimensions.


\section{Miffie Construction}
\label{appendix:miffie_reimpl}

We implement Miffie as described by \citet{jo2026databasenormalizationdualllmselfrefinement}. The method uses a generator LLM and a verifier LLM in an iterative refinement loop. To ensure a controlled comparison, both roles use the same model in the {\DBN} \emph{FD-withheld} setting. We adapt the output format so that Miffie produces the fields required by the {\DBN} evaluator, including normalized DDL, violation labels, and explanations. The original generator--verifier refinement structure is otherwise preserved.


\section{Additional Analyses}
\label{appendix:additional_analyses}

\subsection{Scale-Controlled Comparison}
\label{appendix:scale_controlled}

\begin{figure}[!t]
  \centering
  \includegraphics[width=\columnwidth]{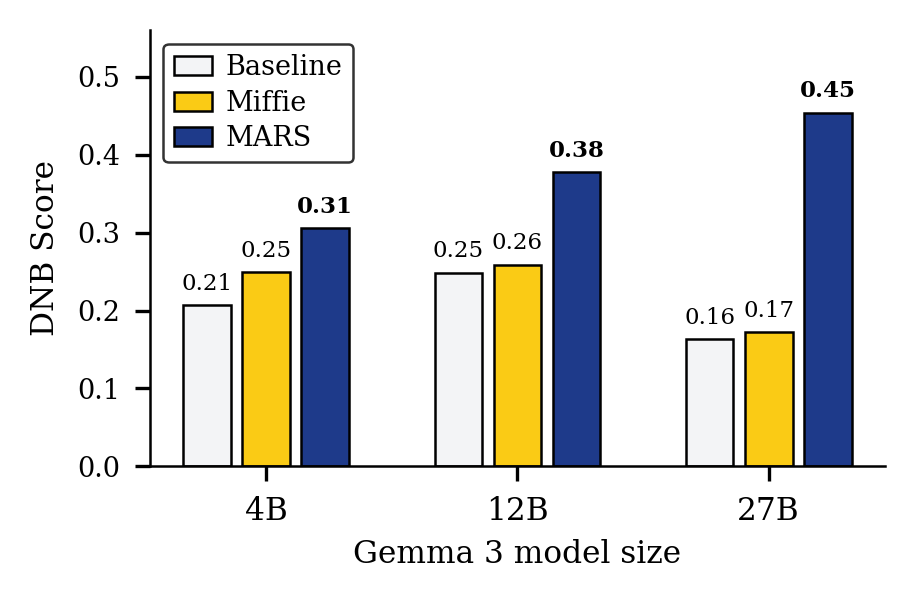}
  \vspace{-2em}
  \caption{Scale-controlled comparison within the Gemma~3 family under the \emph{FD-withheld} setting with zero-shot prompting, pooled over test samples.}
  \label{fig:scale_controlled}
  \vspace{-2ex}
\end{figure}


The model selection in Appendix~\ref{appendix:model_selection} varies architectural family and parameter count simultaneously, so a difference between any two models cannot be attributed to either factor alone. To isolate scale, we additionally compare two models that differ only in parameter count within the same family, under the \emph{FD-withheld} setting with zero-shot prompting.

As Figure~\ref{fig:scale_controlled} shows, the single-prompt Baseline fails to improve monotonically with scale. The 12B model reaches $0.2485$, but the 27B model drops to $0.1631$, representing a decrease of $0.0854$ despite having more than twice the parameter count. Parameter count alone, therefore, does not determine {\DNS}. In contrast, scores under MARS rise consistently with model size, increasing from $0.3054$ at 4B to $0.3776$ at 12B and reaching $0.4539$ at 27B, with every model outperforming its corresponding single-prompt Baseline and Miffie.

\subsection{Variance Across Repeated Runs}
\label{appendix:run_variance}

\begin{figure}[!t]
  \centering
  \includegraphics[width=\columnwidth]{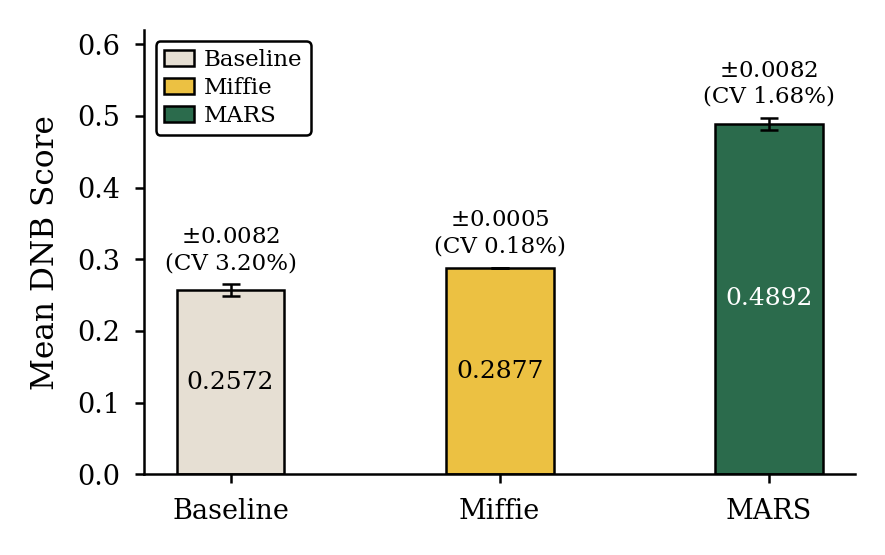}
  \vspace{-2em}
  \caption{Mean, standard deviation, and coefficient of variation (CV) of {\DNS} over three independent runs per method, with Qwen3-30B-A3B-Instruct-2507 in the zero-shot \emph{FD-withheld} setting.}
  \label{fig:run_variance}
  \vspace{-1ex}
\end{figure}

As reported in Figure~\ref{fig:run_variance}, run-to-run variance remains consistently low across all three methods, with coefficients of variation staying within $3.20\%$, confirming that our experimental results are statistically reliable and robust.

\subsection{Under- and Over-Normalization}
\label{appendix:over_normalization}

\begin{table}[!t]
  \centering
  \small
  \setlength{\tabcolsep}{3pt}
  \begin{tabular*}{\columnwidth}{@{\extracolsep{\fill}} llcc}
    \toprule
    \multirow{2}{*}{\textbf{Method}} & \multirow{2}{*}{\textbf{Shot}}
    & \textbf{Under-} & \textbf{Over-} \\
    & & \textbf{Normalization} ($\downarrow$) & \textbf{Normalization} ($\downarrow$) \\
    \midrule
    \multirow{2}{*}{Baseline} & Zero & 9/1028 (0.88\%)   & 155/257 (60.31\%) \\
    & Few  & 3/1028 (0.29\%)   & 138/257 (53.70\%) \\
    \midrule
    \multirow{2}{*}{Miffie}   & Zero & 25/1028 (2.43\%)   & 144/257 (56.03\%) \\
    & Few  & 37/1028 (3.60\%)   & 52/257 (20.23\%) \\
    \midrule
    \multirow{2}{*}{\textbf{MARS}} & Zero & 402/1028 (39.11\%) & 12/257 (4.67\%) \\
    & Few  & 361/1028 (35.12\%) & 12/257 (4.67\%) \\
    \bottomrule
  \end{tabular*}
  \caption{Decomposition errors under the \emph{FD-withheld} setting (Qwen3-30B-A3B-Instruct-2507).}
  \label{tab:over_normalization}
  \vspace{-1ex}
\end{table}

A model can inflate its apparent normalization ability by decomposing every input, including schemas that are already in the target normal form. We therefore separate two error types in the \emph{FD-withheld} setting with Qwen3-30B-A3B-Instruct-2507: unnecessary decomposition of NONE inputs (over-normalization) and missed decomposition of inputs violating a normal form (under-normalization), as shown in Table~\ref{tab:over_normalization}. 

While the single-prompt Baseline successfully decomposes most violating inputs, it suffers from severe over-normalization by splitting over half of already-normalized inputs. In contrast, MARS decomposes inputs only when its extracted dependencies support a violation, thereby reducing unnecessary decomposition to $4.7\%$ across both settings and largely eliminating over-normalization. The primary tradeoff is under-detection: when the Evidence Agent misses the dependency behind a violation, MARS leaves $35.1\%$--$39.1\%$ of violating inputs unchanged. Consequently, recognizing violations whose supporting dependencies are only weakly evidenced in sample rows remains the main open problem for MARS.

\begin{figure}[!t]
  \centering
  \includegraphics[width=\columnwidth]{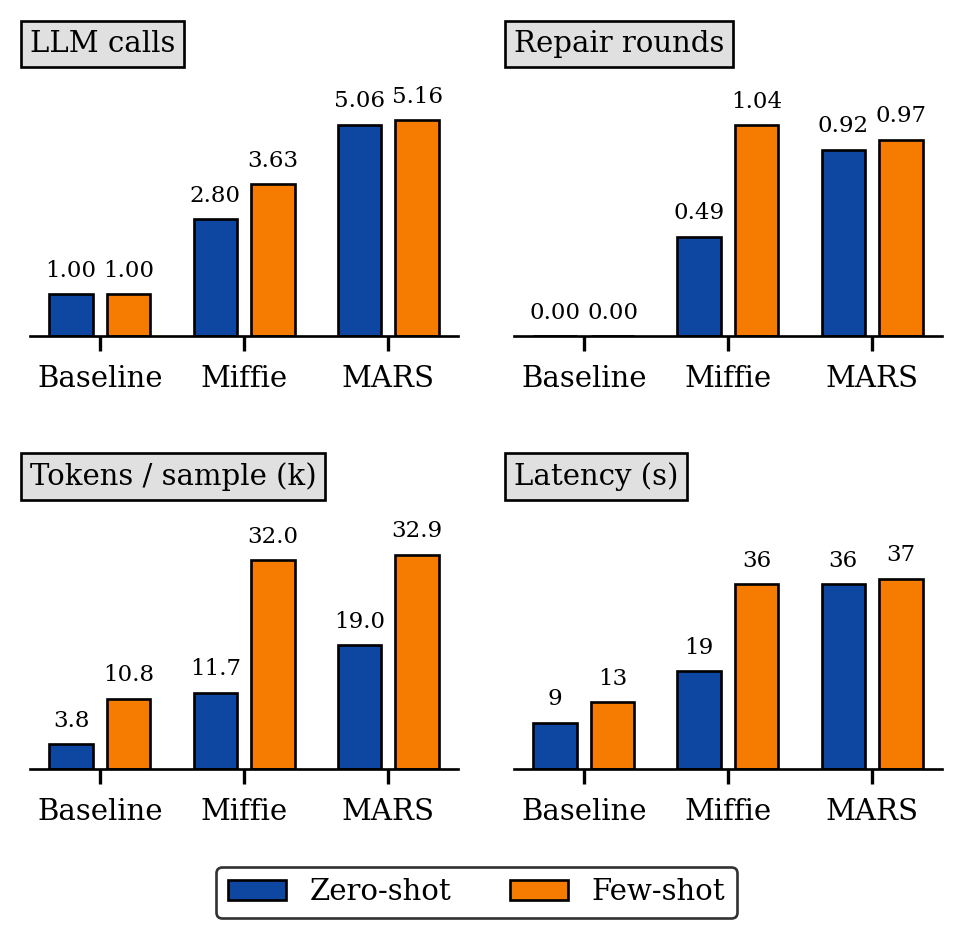}
  \vspace{-2em}
  \caption{Per-sample inference cost under the \emph{FD-withheld} setting, measured on the same backbone (Qwen3-30B-A3B-Instruct-2507). Repair Rounds count verification-triggered regeneration passes.}
  \label{fig:inference_cost}
\end{figure}

\subsection{Inference Cost of MARS}
\label{appendix:inference_cost}

Figure~\ref{fig:inference_cost} reports the per-sample cost for all three methods. MARS requires more calls, tokens, and latency than both single-prompt Baseline and Miffie, and we regard this as genuine cost of the approach rather than negligible overhead. However, the performance gain is not explained by the token budget alone. MARS with zero-shot prompting consumes fewer tokens than Miffie with few-shot prompting while still reaching a higher {\DNS} as shown in Table~\ref{tab:multi-agent_dbn_components_breakdown}. Under a comparable token budget, the role-specialized decomposition therefore contributes more than simply spending more inference compute.
\section{Artifact and Loop Analysis}

\subsection{Miffie Artifact Analysis}
\label{appendix:miffie_artifact_analysis}

Table~\ref{tab:miffie_iteration_loop} summarizes the iteration-loop behavior of Miffie, where each iteration corresponds to one generator--verifier refinement step. The group that terminates after the first iteration achieves an average {\DNS} of $0.3826$. However, undergoing additional iterations causes scores to drop to around $0.27$, regardless of whether samples passed or failed the verifier. This pattern suggests that repeated LLM-based refinements are insufficient to recover from difficult normalization cases, demonstrating that the LLM-based verifier fails to reliably distinguish schema-level normalization quality under {\DBN}'s evaluation.

\subsection{MARS Artifact Analysis}\label{appendix:agent_artifact_analysis}
Table~\ref{tab:repair_loop_effect} summarizes how final {\DNS} changes across MARS repair groups. Outputs passing verification initially achieve the highest {\DNS}. Samples requiring one repair round drop to $0.4759$, while those needing two repair rounds achieve the lowest average score. Among the $1{,}039$ samples requiring two repairs, only $166$ pass final verification with a score of $0.4812$, comparable to the one-repair group, while the remaining $873$ remain unresolved.

Table~\ref{tab:agent_stage_diagnostics} summarizes stage-wise artifact diagnostics for MARS. Exact-match FD recall evaluates whether the Evidence stage recovers gold atomic FDs, while violation-type F1 measures the correctness of the Diagnosis stage against gold violation labels. Plan-to-DDL exact-match checks whether the generated DDL matches the diagnosis plan in terms of relation counts, relation names, columns, and primary keys. Verifier pass rate reports the fraction of final outputs that pass deterministic verification after repair.

The moderate FD recall and violation-type F1 indicate that evidence extraction and diagnosis remain upstream bottlenecks. The Evidence stage is precise but incomplete: FD precision reaches $0.967$, so the FDs it extracts are largely correct, yet recall is only $0.654$. Violation-type F1 is the weakest stage at $0.5999$, consistent with the over-normalization analysis in Appendix~\ref{appendix:over_normalization}: MARS tends to under-report violations and answer NONE. By contrast, the plan-to-DDL exact match of $0.9681$ indicates that the Schema Generator implements a given diagnosis plan with near-perfect reliability. Thus, MARS mainly improves the conversion of a normalization plan into executable DDL, while reliable FD inference remains difficult.

\begin{table}[!t]
  \centering
  \footnotesize
  \renewcommand{\arraystretch}{1.12}
  \setlength{\tabcolsep}{4pt}
  \begin{tabular*}{\columnwidth}{@{\extracolsep{\fill}} l r r r}
    \toprule
    \textbf{Iteration group} & \textbf{Samples} & \textbf{\% of dataset} & \textbf{{\DNS}} \\
    \midrule
    1 iteration                            & 1436 &  55.9\% & 0.3826 \\
    2 iterations                           &  287 &  11.2\% & 0.2684 \\
    3 iterations                           &  847 &  33.0\% & 0.2759 \\
    \quad$\hookrightarrow$\ passed         &  167 &   6.5\% & 0.2655 \\
    \quad$\hookrightarrow$\ failed         &  680 &  26.5\% & 0.2785 \\
    \bottomrule
  \end{tabular*}
  \vspace{-1ex}
  \caption{Iteration-loop behavior of Miffie (Qwen3-30B-A3B-Instruct-2507, \emph{FD-withheld}, zero- and few-shot; $2{,}570$ samples).}
  \label{tab:miffie_iteration_loop}
  \vspace{-1ex}
\end{table}

\begin{table}[!t]
  \centering
  \footnotesize
  \renewcommand{\arraystretch}{1.12}
  \setlength{\tabcolsep}{4pt}
  \begin{tabular*}{\columnwidth}{@{\extracolsep{\fill}} l r r r}
    \toprule
    \textbf{Repair group} & \textbf{Samples} & \textbf{\% of dataset} & \textbf{{\DNS}} \\
    \midrule
    No repair needed & 1177 & 45.8\% & 0.6084 \\
    One repair & 354 & 13.8\% & 0.4759 \\
    Two repairs & 1039 & 40.4\% & 0.3789 \\
    \quad$\hookrightarrow$\ passed & 166 & 6.5\% & 0.4812 \\
    \quad$\hookrightarrow$\ failed & 873\ & 34.0\% & 0.3595 \\
    \bottomrule
  \end{tabular*}
  \vspace{-1ex}
  \caption{Repair-loop behavior of MARS (Qwen3-30B-A3B-Instruct-2507, \emph{FD-withheld}, zero- and few-shot; $2{,}570$ samples).}
  \label{tab:repair_loop_effect}
  \vspace{-1ex}
\end{table}




\begin{table}[!t]
  \centering
  \footnotesize
  \renewcommand{\arraystretch}{1.12}
  \setlength{\tabcolsep}{4pt}
  \begin{tabular*}{\columnwidth}{@{\extracolsep{\fill}} l l r}
    \toprule
    \textbf{Stage} & \textbf{Metric} & \textbf{Score} \\
    \midrule
    Evidence & Exact-match FD recall & 0.6539 \\
    Diagnosis & Violation-type F1 & 0.5999 \\
    Schema generation & Plan-to-DDL exact match & 0.9681 \\
    Verification & Verifier pass rate & 0.6603 \\
    \bottomrule
  \end{tabular*}
  \vspace{-1ex}
  \caption{Stage-wise artifact diagnostics for MARS.}
  \label{tab:agent_stage_diagnostics}
  \vspace{-1ex}
\end{table}
\section{Extended Result Tables}
\label{appendix:extended_results}

This appendix collects the per-cell result tables underlying
Tables~\ref{tab:dbn_results}--\ref{tab:multi-agent_dbn_components_breakdown} of the main paper, covering the single-prompt Baseline, Miffie, and MARS.

Tables~\ref{tab:dbn_per_dataset} and \ref{tab:realworld_by_violation} report {\DNS} values for the base models, broken down by dataset and by violation path, respectively.

Table~\ref{tab:full_matrix} provides the full per-cell results for the base models: {\DNS} and their three component scores for every (dataset, model, prompting setting, violation path) configuration.

Tables~\ref{tab:multi-agent_full_matrix} and \ref{tab:miffie_full_matrix} report the corresponding per-cell breakdowns for MARS and Miffie, respectively, applied to each of the four base models under the \emph{FD-withheld} setting.



\clearpage
\onecolumn
\begin{table*}[htb]
  \centering
  \footnotesize
  \begin{tabular*}{\textwidth}{@{\extracolsep{\fill}} llcccccc}
    \toprule
    \multirow{2}{*}[-0.3em]{\textbf{Dataset}} & \multirow{2}{*}[-0.3em]{\textbf{Model}} & \multicolumn{2}{c}{\textbf{Single}} & \multicolumn{2}{c}{\textbf{Complex}} & \multicolumn{2}{c}{\textbf{FD-withheld}} \\
    \cmidrule(lr){3-4} \cmidrule(lr){5-6} \cmidrule(lr){7-8}
    & & Zero-shot & Few-shot & Zero-shot & Few-shot & Zero-shot & Few-shot \\
    \midrule
    \multirow{4}{*}{BIRD} & Gemma-3-27B-it               & 0.355          & 0.337          & \textbf{0.357} & \textbf{0.404} & 0.184          & 0.238          \\
    & Qwen3-30B-A3B-Instruct-2507  & \textbf{0.494} & \textbf{0.371} & 0.342          & 0.344          & \textbf{0.225} & \textbf{0.280} \\
    & Mixtral-8x7B-Instruct-v0.1   & 0.138          & 0.154          & 0.157          & 0.185          & 0.101          & 0.106          \\
    & Llama-3.3-70B-Instruct       & 0.325          & 0.356          & 0.357          & 0.396          & 0.191          & 0.235          \\
    \midrule
    \multirow{4}{*}{Spider} & Gemma-3-27B-it               & 0.354          & 0.395          & 0.364          & 0.461          & 0.154          & 0.261          \\
    & Qwen3-30B-A3B-Instruct-2507  & \textbf{0.489} & 0.422          & 0.352          & 0.394          & \textbf{0.257} & \textbf{0.299} \\
    & Mixtral-8x7B-Instruct-v0.1   & 0.133          & 0.180          & 0.141          & 0.173          & 0.123          & 0.115          \\
    & Llama-3.3-70B-Instruct       & 0.393          & \textbf{0.446} & \textbf{0.399} & \textbf{0.485} & 0.214          & 0.252          \\
    \bottomrule
  \end{tabular*}
  \caption{Single-prompt Baseline {\DNS} values per model on each test split. Each cell averages the five violation-path categories (NONE, BCNF only, 3NF+BCNF, 2NF+3NF+BCNF, 1NF+2NF+3NF+BCNF) for the given (dataset, model, task, prompt) configuration. \textbf{Bold} marks the best model in each column within a dataset.}
  \label{tab:dbn_per_dataset}
\end{table*}

\begin{table*}[htb]
  \centering
  \small
  \setlength{\tabcolsep}{3pt}
  \begin{tabular*}{\textwidth}{@{\extracolsep{\fill}} ll ccc ccc}
    \toprule
    \multirow{2}{*}[-0.3em]{\textbf{Model}} & \multirow{2}{*}[-0.3em]{\textbf{Violation Path}}
    & \multicolumn{3}{c}{\textbf{BIRD}}
    & \multicolumn{3}{c}{\textbf{Spider}} \\
    \cmidrule(lr){3-5} \cmidrule(lr){6-8}
    & & Zero-shot & Few-shot & Avg
    & Zero-shot & Few-shot & Avg \\
    \midrule
    \multirow{5}{*}{Gemma-3-27B-it} & 1NF-2NF-3NF-BCNF   & 0.173          & 0.205          & 0.189          & 0.122          & 0.229          & 0.175          \\
    & 2NF-3NF-BCNF       & 0.181          & 0.173          & 0.177          & 0.176          & 0.275          & 0.225          \\
    & 3NF-BCNF           & 0.124          & 0.211          & 0.168          & 0.089          & 0.180          & 0.134          \\
    & BCNF               & 0.111          & 0.116          & 0.113          & 0.055          & 0.084          & 0.070          \\
    & NONE               & \textbf{0.334} & \textbf{0.483} & \textbf{0.408} & \textbf{0.328} & \textbf{0.535} & \textbf{0.431} \\
    \midrule
    \multirow{5}{*}{Qwen3-30B-A3B-Instruct-2507} & 1NF-2NF-3NF-BCNF   & 0.139          & 0.190          & 0.164          & 0.152          & 0.172          & 0.162          \\
    & 2NF-3NF-BCNF       & 0.151          & 0.175          & 0.163          & 0.144          & 0.182          & 0.163          \\
    & 3NF-BCNF           & 0.187          & 0.351          & 0.269          & 0.248          & 0.356          & 0.302          \\
    & BCNF               & 0.166          & 0.153          & 0.159          & 0.137          & 0.178          & 0.158          \\
    & NONE               & \textbf{0.481} & \textbf{0.533} & \textbf{0.507} & \textbf{0.606} & \textbf{0.605} & \textbf{0.606} \\
    \midrule
    \multirow{5}{*}{Mixtral-8x7B-Instruct-v0.1} & 1NF-2NF-3NF-BCNF   & 0.027          & 0.020          & 0.024          & 0.044          & 0.040          & 0.042          \\
    & 2NF-3NF-BCNF       & 0.036          & 0.016          & 0.026          & 0.028          & 0.047          & 0.037          \\
    & 3NF-BCNF           & 0.089          & 0.076          & 0.083          & 0.115          & 0.138          & 0.126          \\
    & BCNF               & 0.059          & 0.125          & 0.092          & 0.093          & 0.070          & 0.081          \\
    & NONE               & \textbf{0.292} & \textbf{0.294} & \textbf{0.293} & \textbf{0.336} & \textbf{0.278} & \textbf{0.307} \\
    \midrule
    \multirow{5}{*}{Llama-3.3-70B-Instruct} & 1NF-2NF-3NF-BCNF   & 0.120          & 0.149          & 0.134          & 0.136          & 0.229          & 0.183          \\
    & 2NF-3NF-BCNF       & 0.107          & 0.210          & 0.158          & 0.195          & 0.275          & 0.235          \\
    & 3NF-BCNF           & 0.104          & 0.214          & 0.159          & 0.125          & 0.267          & 0.196          \\
    & BCNF               & 0.098          & 0.162          & 0.130          & 0.081          & 0.132          & 0.106          \\
    & NONE               & \textbf{0.527} & \textbf{0.440} & \textbf{0.483} & \textbf{0.534} & \textbf{0.358} & \textbf{0.446} \\
    \bottomrule
  \end{tabular*}
  \caption{Single-prompt Baseline FD-withheld {\DNS} values ($\uparrow$) by model, violation path, and dataset. The five violation paths correspond to mixed violations and already-normalized cases with no violation (NONE). \textbf{Bold} marks the best score per column within each model among the five violation paths.}
  \label{tab:realworld_by_violation}
  \vspace{-3mm}
\end{table*}

\clearpage

{\scriptsize
  \setlength{\tabcolsep}{2pt}
  \renewcommand{\arraystretch}{1.15}
  \setlength{\LTleft}{0pt}
  \setlength{\LTright}{0pt}
  \setlength{\LTcapwidth}{\linewidth}
  \captionsetup{hypcap=false,position=below,skip=6pt}
  \begin{xltabular}{\linewidth}{|l|l|l|l|l|>{\centering\arraybackslash}X|>{\centering\arraybackslash}X|>{\centering\arraybackslash}X|>{\centering\arraybackslash}X|>{\centering\arraybackslash}X|}
    \hline
    \multirow{2}{*}{\textbf{Dataset}} & \multirow{2}{*}{\textbf{Model}} & \multirow{2}{*}{\textbf{Shot}} & \multirow{2}{*}{\textbf{Experiment}} & \multirow{2}{*}{\textbf{Violation Path}} & \multirow{2}{*}{\textbf{Num.}} & \multirow{2}{*}{\textbf{{\DNS}}} & \multicolumn{3}{c|}{\textbf{Components}} \\
    \cline{8-10}
    & & & & & & & \textbf{Semantic} & \textbf{Structural} & \textbf{Logical} \\
    \hline
    \endfirsthead

    \hline
    \multirow{2}{*}{\textbf{Dataset}} & \multirow{2}{*}{\textbf{Model}} & \multirow{2}{*}{\textbf{Shot}} & \multirow{2}{*}{\textbf{Experiment}} & \multirow{2}{*}{\textbf{Violation Path}} & \multirow{2}{*}{\textbf{Num.}} & \multirow{2}{*}{\textbf{{\DNS}}} & \multicolumn{3}{c|}{\textbf{Components}} \\
    \cline{8-10}
    & & & & & & & \textbf{Semantic} & \textbf{Structural} & \textbf{Logical} \\
    \hline
    \endhead

    \endfoot

    \noalign{\vskip 6pt}
    \caption{Single-prompt Baseline full evaluation matrix: {\DNS} and its three components for every (dataset, model, prompting setting, violation path) cell. Results are reported separately by \textbf{Shot} (zero/few) and \textbf{Experiment} (\emph{Single}/\emph{Complex}/\emph{FD-withheld}).}\label{tab:full_matrix}\\
    \endlastfoot

    BIRD & Gemma-3-27B-it & Zero-shot & Single & 1\_2\_3\_BCNF & 77 & 0.4896 & 0.8442 & 0.5964 & 0.5939 \\\arrayrulecolor{gray!30}\cline{5-10}
    &  &  &  & 2\_3\_BCNF & 77 & 0.1940 & 0.8571 & 0.6539 & 0.1554 \\\arrayrulecolor{gray!30}\cline{5-10}
    &  &  &  & 3\_BCNF & 77 & 0.1509 & 0.8571 & 0.5904 & 0.0727 \\\arrayrulecolor{gray!30}\cline{5-10}
    &  &  &  & BCNF & 77 & 0.1414 & 0.8831 & 0.5443 & 0.0649 \\\arrayrulecolor{gray!30}\cline{5-10}
    &  &  &  & NONE & 77 & 0.7986 & 0.9740 & 0.9416 & 0.7468 \\\arrayrulecolor{gray!40}\cline{4-10}
    &  &  & Complex & 1\_2\_3\_BCNF & 77 & 0.3331 & 0.5974 & 0.5444 & 0.5997 \\\arrayrulecolor{gray!30}\cline{5-10}
    &  &  &  & 2\_3\_BCNF & 77 & 0.4048 & 0.7273 & 0.5953 & 0.5263 \\\arrayrulecolor{gray!30}\cline{5-10}
    &  &  &  & 3\_BCNF & 77 & 0.3029 & 0.5844 & 0.6117 & 0.4242 \\\arrayrulecolor{gray!30}\cline{5-10}
    &  &  &  & BCNF & 77 & 0.2733 & 0.7143 & 0.5500 & 0.2801 \\\arrayrulecolor{gray!30}\cline{5-10}
    &  &  &  & NONE & 77 & 0.4708 & 0.9610 & 0.7807 & 0.3948 \\\arrayrulecolor{gray!40}\cline{4-10}
    &  &  & FD-withheld & 1\_2\_3\_BCNF & 77 & 0.1729 & 0.3506 & 0.5455 & 0.4905 \\\arrayrulecolor{gray!30}\cline{5-10}
    &  &  &  & 2\_3\_BCNF & 77 & 0.1811 & 0.3636 & 0.6026 & 0.4239 \\\arrayrulecolor{gray!30}\cline{5-10}
    &  &  &  & 3\_BCNF & 77 & 0.1241 & 0.4935 & 0.5614 & 0.1751 \\\arrayrulecolor{gray!30}\cline{5-10}
    &  &  &  & BCNF & 77 & 0.1106 & 0.4935 & 0.5455 & 0.1011 \\\arrayrulecolor{gray!30}\cline{5-10}
    &  &  &  & NONE & 77 & 0.3338 & 0.8571 & 0.7131 & 0.2537 \\\arrayrulecolor{gray!50}\cline{3-10}
    &  & Few-shot & Single & 1\_2\_3\_BCNF & 77 & 0.1788 & 0.5974 & 0.6077 & 0.2035 \\\arrayrulecolor{gray!30}\cline{5-10}
    &  &  &  & 2\_3\_BCNF & 77 & 0.3061 & 0.6104 & 0.5942 & 0.4654 \\\arrayrulecolor{gray!30}\cline{5-10}
    &  &  &  & 3\_BCNF & 77 & 0.2549 & 0.7532 & 0.5827 & 0.2437 \\\arrayrulecolor{gray!30}\cline{5-10}
    &  &  &  & BCNF & 77 & 0.2163 & 0.8052 & 0.5316 & 0.2108 \\\arrayrulecolor{gray!30}\cline{5-10}
    &  &  &  & NONE & 77 & 0.7293 & 0.9481 & 0.8776 & 0.7095 \\\arrayrulecolor{gray!40}\cline{4-10}
    &  &  & Complex & 1\_2\_3\_BCNF & 77 & 0.2914 & 0.5325 & 0.5688 & 0.4889 \\\arrayrulecolor{gray!30}\cline{5-10}
    &  &  &  & 2\_3\_BCNF & 77 & 0.3275 & 0.5325 & 0.6228 & 0.5370 \\\arrayrulecolor{gray!30}\cline{5-10}
    &  &  &  & 3\_BCNF & 77 & 0.3804 & 0.7403 & 0.6246 & 0.4355 \\\arrayrulecolor{gray!30}\cline{5-10}
    &  &  &  & BCNF & 77 & 0.3093 & 0.7792 & 0.5206 & 0.3814 \\\arrayrulecolor{gray!30}\cline{5-10}
    &  &  &  & NONE & 77 & 0.7102 & 0.8831 & 0.8651 & 0.7229 \\\arrayrulecolor{gray!40}\cline{4-10}
    &  &  & FD-withheld & 1\_2\_3\_BCNF & 77 & 0.2046 & 0.4805 & 0.5467 & 0.3506 \\\arrayrulecolor{gray!30}\cline{5-10}
    &  &  &  & 2\_3\_BCNF & 77 & 0.1732 & 0.3636 & 0.5990 & 0.3883 \\\arrayrulecolor{gray!30}\cline{5-10}
    &  &  &  & 3\_BCNF & 77 & 0.2109 & 0.5974 & 0.6266 & 0.2433 \\\arrayrulecolor{gray!30}\cline{5-10}
    &  &  &  & BCNF & 77 & 0.1158 & 0.5584 & 0.5060 & 0.1649 \\\arrayrulecolor{gray!30}\cline{5-10}
    &  &  &  & NONE & 77 & 0.4831 & 0.7403 & 0.7617 & 0.5351 \\\arrayrulecolor{gray!60}\cline{2-10}
    & Qwen3-30B-A3B-Instruct-2507 & Zero-shot & Single & 1\_2\_3\_BCNF & 77 & 0.5782 & 1.0000 & 0.6543 & 0.5684 \\\arrayrulecolor{gray!30}\cline{5-10}
    &  &  &  & 2\_3\_BCNF & 77 & 0.5723 & 1.0000 & 0.6573 & 0.5528 \\\arrayrulecolor{gray!30}\cline{5-10}
    &  &  &  & 3\_BCNF & 77 & 0.2104 & 0.9481 & 0.5445 & 0.1593 \\\arrayrulecolor{gray!30}\cline{5-10}
    &  &  &  & BCNF & 77 & 0.1523 & 0.9740 & 0.5468 & 0.0913 \\\arrayrulecolor{gray!30}\cline{5-10}
    &  &  &  & NONE & 77 & 0.9546 & 1.0000 & 1.0000 & 0.9190 \\\arrayrulecolor{gray!40}\cline{4-10}
    &  &  & Complex & 1\_2\_3\_BCNF & 77 & 0.2304 & 0.4026 & 0.5539 & 0.6652 \\\arrayrulecolor{gray!30}\cline{5-10}
    &  &  &  & 2\_3\_BCNF & 77 & 0.2896 & 0.4805 & 0.6497 & 0.5541 \\\arrayrulecolor{gray!30}\cline{5-10}
    &  &  &  & 3\_BCNF & 77 & 0.3453 & 0.6753 & 0.6099 & 0.4286 \\\arrayrulecolor{gray!30}\cline{5-10}
    &  &  &  & BCNF & 77 & 0.2537 & 0.6753 & 0.5287 & 0.2766 \\\arrayrulecolor{gray!30}\cline{5-10}
    &  &  &  & NONE & 77 & 0.5913 & 0.9091 & 0.7499 & 0.5524 \\\arrayrulecolor{gray!40}\cline{4-10}
    &  &  & FD-withheld & 1\_2\_3\_BCNF & 77 & 0.1389 & 0.2857 & 0.5523 & 0.6047 \\\arrayrulecolor{gray!30}\cline{5-10}
    &  &  &  & 2\_3\_BCNF & 77 & 0.1505 & 0.3117 & 0.5968 & 0.5227 \\\arrayrulecolor{gray!30}\cline{5-10}
    &  &  &  & 3\_BCNF & 77 & 0.1869 & 0.4156 & 0.5678 & 0.3732 \\\arrayrulecolor{gray!30}\cline{5-10}
    &  &  &  & BCNF & 77 & 0.1661 & 0.4805 & 0.5196 & 0.2426 \\\arrayrulecolor{gray!30}\cline{5-10}
    &  &  &  & NONE & 77 & 0.4813 & 0.8312 & 0.7292 & 0.4571 \\\arrayrulecolor{gray!50}\cline{3-10}
    &  & Few-shot & Single & 1\_2\_3\_BCNF & 77 & 0.1799 & 0.5844 & 0.4763 & 0.2351 \\\arrayrulecolor{gray!30}\cline{5-10}
    &  &  &  & 2\_3\_BCNF & 77 & 0.5856 & 0.8052 & 0.7692 & 0.7740 \\\arrayrulecolor{gray!30}\cline{5-10}
    &  &  &  & 3\_BCNF & 77 & 0.2413 & 0.7662 & 0.4721 & 0.3615 \\\arrayrulecolor{gray!30}\cline{5-10}
    &  &  &  & BCNF & 77 & 0.1687 & 0.7922 & 0.4641 & 0.2372 \\\arrayrulecolor{gray!30}\cline{5-10}
    &  &  &  & NONE & 77 & 0.6787 & 0.9351 & 0.8463 & 0.6359 \\\arrayrulecolor{gray!40}\cline{4-10}
    &  &  & Complex & 1\_2\_3\_BCNF & 77 & 0.1853 & 0.2468 & 0.6683 & 0.6783 \\\arrayrulecolor{gray!30}\cline{5-10}
    &  &  &  & 2\_3\_BCNF & 77 & 0.2125 & 0.2857 & 0.7670 & 0.6211 \\\arrayrulecolor{gray!30}\cline{5-10}
    &  &  &  & 3\_BCNF & 77 & 0.3756 & 0.5974 & 0.7358 & 0.4974 \\\arrayrulecolor{gray!30}\cline{5-10}
    &  &  &  & BCNF & 77 & 0.3262 & 0.6494 & 0.6611 & 0.3452 \\\arrayrulecolor{gray!30}\cline{5-10}
    &  &  &  & NONE & 77 & 0.6181 & 0.9351 & 0.7719 & 0.5810 \\\arrayrulecolor{gray!40}\cline{4-10}
    &  &  & FD-withheld & 1\_2\_3\_BCNF & 77 & 0.1897 & 0.3117 & 0.5922 & 0.5998 \\\arrayrulecolor{gray!30}\cline{5-10}
    &  &  &  & 2\_3\_BCNF & 77 & 0.1747 & 0.2727 & 0.6739 & 0.5565 \\\arrayrulecolor{gray!30}\cline{5-10}
    &  &  &  & 3\_BCNF & 77 & 0.3510 & 0.5974 & 0.6774 & 0.4684 \\\arrayrulecolor{gray!30}\cline{5-10}
    &  &  &  & BCNF & 77 & 0.1528 & 0.5844 & 0.5159 & 0.2433 \\\arrayrulecolor{gray!30}\cline{5-10}
    &  &  &  & NONE & 77 & 0.5331 & 0.7662 & 0.7551 & 0.5199 \\\arrayrulecolor{gray!60}\cline{2-10}
    & Mixtral-8x7B-Instruct-v0.1 & Zero-shot & Single & 1\_2\_3\_BCNF & 77 & 0.0732 & 0.5455 & 0.3517 & 0.0680 \\\arrayrulecolor{gray!30}\cline{5-10}
    &  &  &  & 2\_3\_BCNF & 77 & 0.1621 & 0.3896 & 0.3716 & 0.4156 \\\arrayrulecolor{gray!30}\cline{5-10}
    &  &  &  & 3\_BCNF & 77 & 0.1274 & 0.5844 & 0.4974 & 0.1351 \\\arrayrulecolor{gray!30}\cline{5-10}
    &  &  &  & BCNF & 77 & 0.1451 & 0.6234 & 0.4396 & 0.2065 \\\arrayrulecolor{gray!30}\cline{5-10}
    &  &  &  & NONE & 77 & 0.1812 & 0.7403 & 0.5122 & 0.1563 \\\arrayrulecolor{gray!40}\cline{4-10}
    &  &  & Complex & 1\_2\_3\_BCNF & 77 & 0.0945 & 0.2727 & 0.2301 & 0.3715 \\\arrayrulecolor{gray!30}\cline{5-10}
    &  &  &  & 2\_3\_BCNF & 77 & 0.1206 & 0.3377 & 0.2740 & 0.3504 \\\arrayrulecolor{gray!30}\cline{5-10}
    &  &  &  & 3\_BCNF & 77 & 0.1794 & 0.5455 & 0.4009 & 0.2695 \\\arrayrulecolor{gray!30}\cline{5-10}
    &  &  &  & BCNF & 77 & 0.1660 & 0.5325 & 0.3856 & 0.2823 \\\arrayrulecolor{gray!30}\cline{5-10}
    &  &  &  & NONE & 77 & 0.2265 & 0.7403 & 0.4868 & 0.2171 \\\arrayrulecolor{gray!40}\cline{4-10}
    &  &  & FD-withheld & 1\_2\_3\_BCNF & 77 & 0.0273 & 0.1039 & 0.3098 & 0.2957 \\\arrayrulecolor{gray!30}\cline{5-10}
    &  &  &  & 2\_3\_BCNF & 77 & 0.0364 & 0.1299 & 0.2825 & 0.2489 \\\arrayrulecolor{gray!30}\cline{5-10}
    &  &  &  & 3\_BCNF & 77 & 0.0894 & 0.3506 & 0.4752 & 0.1537 \\\arrayrulecolor{gray!30}\cline{5-10}
    &  &  &  & BCNF & 77 & 0.0588 & 0.2727 & 0.4155 & 0.1506 \\\arrayrulecolor{gray!30}\cline{5-10}
    &  &  &  & NONE & 77 & 0.2919 & 0.6753 & 0.5870 & 0.3411 \\\arrayrulecolor{gray!50}\cline{3-10}
    &  & Few-shot & Single & 1\_2\_3\_BCNF & 77 & 0.0841 & 0.3377 & 0.4907 & 0.1102 \\\arrayrulecolor{gray!30}\cline{5-10}
    &  &  &  & 2\_3\_BCNF & 77 & 0.1882 & 0.3636 & 0.6163 & 0.3894 \\\arrayrulecolor{gray!30}\cline{5-10}
    &  &  &  & 3\_BCNF & 77 & 0.0652 & 0.2597 & 0.5603 & 0.1786 \\\arrayrulecolor{gray!30}\cline{5-10}
    &  &  &  & BCNF & 77 & 0.1523 & 0.4545 & 0.4676 & 0.2866 \\\arrayrulecolor{gray!30}\cline{5-10}
    &  &  &  & NONE & 77 & 0.2812 & 0.7532 & 0.6451 & 0.2494 \\\arrayrulecolor{gray!40}\cline{4-10}
    &  &  & Complex & 1\_2\_3\_BCNF & 77 & 0.1045 & 0.2078 & 0.4994 & 0.4579 \\\arrayrulecolor{gray!30}\cline{5-10}
    &  &  &  & 2\_3\_BCNF & 77 & 0.0588 & 0.1169 & 0.5187 & 0.4439 \\\arrayrulecolor{gray!30}\cline{5-10}
    &  &  &  & 3\_BCNF & 77 & 0.1134 & 0.2597 & 0.5177 & 0.4506 \\\arrayrulecolor{gray!30}\cline{5-10}
    &  &  &  & BCNF & 77 & 0.1833 & 0.3377 & 0.4916 & 0.5123 \\\arrayrulecolor{gray!30}\cline{5-10}
    &  &  &  & NONE & 77 & 0.4662 & 0.8571 & 0.7556 & 0.4450 \\\arrayrulecolor{gray!40}\cline{4-10}
    &  &  & FD-withheld & 1\_2\_3\_BCNF & 77 & 0.0197 & 0.0390 & 0.4604 & 0.4438 \\\arrayrulecolor{gray!30}\cline{5-10}
    &  &  &  & 2\_3\_BCNF & 77 & 0.0157 & 0.0260 & 0.4738 & 0.4936 \\\arrayrulecolor{gray!30}\cline{5-10}
    &  &  &  & 3\_BCNF & 77 & 0.0760 & 0.1558 & 0.4980 & 0.4656 \\\arrayrulecolor{gray!30}\cline{5-10}
    &  &  &  & BCNF & 77 & 0.1254 & 0.3117 & 0.4329 & 0.3656 \\\arrayrulecolor{gray!30}\cline{5-10}
    &  &  &  & NONE & 77 & 0.2939 & 0.7273 & 0.6492 & 0.2606 \\\arrayrulecolor{gray!60}\cline{2-10}
    & Llama-3.3-70B-Instruct & Zero-shot & Single & 1\_2\_3\_BCNF & 77 & 0.2224 & 0.5455 & 0.5847 & 0.3697 \\\arrayrulecolor{gray!30}\cline{5-10}
    &  &  &  & 2\_3\_BCNF & 77 & 0.4259 & 0.6623 & 0.7208 & 0.4935 \\\arrayrulecolor{gray!30}\cline{5-10}
    &  &  &  & 3\_BCNF & 77 & 0.2007 & 0.6753 & 0.6004 & 0.1450 \\\arrayrulecolor{gray!30}\cline{5-10}
    &  &  &  & BCNF & 77 & 0.1558 & 0.5974 & 0.5340 & 0.1892 \\\arrayrulecolor{gray!30}\cline{5-10}
    &  &  &  & NONE & 77 & 0.6203 & 0.7792 & 0.9091 & 0.7615 \\\arrayrulecolor{gray!40}\cline{4-10}
    &  &  & Complex & 1\_2\_3\_BCNF & 77 & 0.2479 & 0.4805 & 0.5755 & 0.4818 \\\arrayrulecolor{gray!30}\cline{5-10}
    &  &  &  & 2\_3\_BCNF & 77 & 0.3804 & 0.6753 & 0.6365 & 0.4917 \\\arrayrulecolor{gray!30}\cline{5-10}
    &  &  &  & 3\_BCNF & 77 & 0.3229 & 0.6623 & 0.6008 & 0.4002 \\\arrayrulecolor{gray!30}\cline{5-10}
    &  &  &  & BCNF & 77 & 0.2104 & 0.5844 & 0.5398 & 0.3084 \\\arrayrulecolor{gray!30}\cline{5-10}
    &  &  &  & NONE & 77 & 0.6231 & 0.7532 & 0.9055 & 0.8022 \\\arrayrulecolor{gray!40}\cline{4-10}
    &  &  & FD-withheld & 1\_2\_3\_BCNF & 77 & 0.1203 & 0.2857 & 0.5677 & 0.3971 \\\arrayrulecolor{gray!30}\cline{5-10}
    &  &  &  & 2\_3\_BCNF & 77 & 0.1072 & 0.2857 & 0.6109 & 0.3214 \\\arrayrulecolor{gray!30}\cline{5-10}
    &  &  &  & 3\_BCNF & 77 & 0.1037 & 0.3766 & 0.5639 & 0.2206 \\\arrayrulecolor{gray!30}\cline{5-10}
    &  &  &  & BCNF & 77 & 0.0985 & 0.3896 & 0.4840 & 0.1481 \\\arrayrulecolor{gray!30}\cline{5-10}
    &  &  &  & NONE & 77 & 0.5267 & 0.8442 & 0.7218 & 0.5385 \\\arrayrulecolor{gray!50}\cline{3-10}
    &  & Few-shot & Single & 1\_2\_3\_BCNF & 77 & 0.1833 & 0.4805 & 0.4323 & 0.5312 \\\arrayrulecolor{gray!30}\cline{5-10}
    &  &  &  & 2\_3\_BCNF & 77 & 0.4262 & 0.7143 & 0.5968 & 0.6268 \\\arrayrulecolor{gray!30}\cline{5-10}
    &  &  &  & 3\_BCNF & 77 & 0.2331 & 0.7013 & 0.6219 & 0.2446 \\\arrayrulecolor{gray!30}\cline{5-10}
    &  &  &  & BCNF & 77 & 0.1525 & 0.7013 & 0.5002 & 0.2117 \\\arrayrulecolor{gray!30}\cline{5-10}
    &  &  &  & NONE & 77 & 0.7830 & 0.9221 & 0.9063 & 0.8364 \\\arrayrulecolor{gray!40}\cline{4-10}
    &  &  & Complex & 1\_2\_3\_BCNF & 77 & 0.2801 & 0.4286 & 0.6492 & 0.6310 \\\arrayrulecolor{gray!30}\cline{5-10}
    &  &  &  & 2\_3\_BCNF & 77 & 0.4001 & 0.5584 & 0.7150 & 0.6364 \\\arrayrulecolor{gray!30}\cline{5-10}
    &  &  &  & 3\_BCNF & 77 & 0.3105 & 0.5065 & 0.6696 & 0.5167 \\\arrayrulecolor{gray!30}\cline{5-10}
    &  &  &  & BCNF & 77 & 0.2459 & 0.5974 & 0.5265 & 0.3338 \\\arrayrulecolor{gray!30}\cline{5-10}
    &  &  &  & NONE & 77 & 0.7422 & 0.8831 & 0.8965 & 0.8052 \\\arrayrulecolor{gray!40}\cline{4-10}
    &  &  & FD-withheld & 1\_2\_3\_BCNF & 77 & 0.1486 & 0.2987 & 0.5255 & 0.5627 \\\arrayrulecolor{gray!30}\cline{5-10}
    &  &  &  & 2\_3\_BCNF & 77 & 0.2098 & 0.3896 & 0.6211 & 0.5275 \\\arrayrulecolor{gray!30}\cline{5-10}
    &  &  &  & 3\_BCNF & 77 & 0.2139 & 0.4416 & 0.6104 & 0.4305 \\\arrayrulecolor{gray!30}\cline{5-10}
    &  &  &  & BCNF & 77 & 0.1619 & 0.5195 & 0.5002 & 0.2398 \\\arrayrulecolor{gray!30}\cline{5-10}
    &  &  &  & NONE & 77 & 0.4401 & 0.6494 & 0.8135 & 0.6983 \\\arrayrulecolor{black}\hline
    Spider & Gemma-3-27B-it & Zero-shot & Single & 1\_2\_3\_BCNF & 180 & 0.5033 & 0.8000 & 0.6055 & 0.6770 \\\arrayrulecolor{gray!30}\cline{5-10}
    &  &  &  & 2\_3\_BCNF & 180 & 0.2979 & 0.7444 & 0.7036 & 0.2778 \\\arrayrulecolor{gray!30}\cline{5-10}
    &  &  &  & 3\_BCNF & 180 & 0.1598 & 0.6778 & 0.6023 & 0.0993 \\\arrayrulecolor{gray!30}\cline{5-10}
    &  &  &  & BCNF & 180 & 0.1421 & 0.7778 & 0.5220 & 0.0833 \\\arrayrulecolor{gray!30}\cline{5-10}
    &  &  &  & NONE & 180 & 0.6648 & 0.9389 & 0.8635 & 0.6133 \\\arrayrulecolor{gray!40}\cline{4-10}
    &  &  & Complex & 1\_2\_3\_BCNF & 180 & 0.4262 & 0.7333 & 0.5722 & 0.6043 \\\arrayrulecolor{gray!30}\cline{5-10}
    &  &  &  & 2\_3\_BCNF & 180 & 0.3883 & 0.6833 & 0.6212 & 0.5311 \\\arrayrulecolor{gray!30}\cline{5-10}
    &  &  &  & 3\_BCNF & 180 & 0.3460 & 0.6611 & 0.5993 & 0.4469 \\\arrayrulecolor{gray!30}\cline{5-10}
    &  &  &  & BCNF & 180 & 0.2786 & 0.6889 & 0.5449 & 0.3029 \\\arrayrulecolor{gray!30}\cline{5-10}
    &  &  &  & NONE & 180 & 0.3824 & 0.9500 & 0.7026 & 0.2904 \\\arrayrulecolor{gray!40}\cline{4-10}
    &  &  & FD-withheld & 1\_2\_3\_BCNF & 180 & 0.1219 & 0.2278 & 0.5754 & 0.5125 \\\arrayrulecolor{gray!30}\cline{5-10}
    &  &  &  & 2\_3\_BCNF & 180 & 0.1760 & 0.3389 & 0.6224 & 0.4256 \\\arrayrulecolor{gray!30}\cline{5-10}
    &  &  &  & 3\_BCNF & 180 & 0.0888 & 0.3333 & 0.5592 & 0.1534 \\\arrayrulecolor{gray!30}\cline{5-10}
    &  &  &  & BCNF & 180 & 0.0554 & 0.3000 & 0.5075 & 0.0841 \\\arrayrulecolor{gray!30}\cline{5-10}
    &  &  &  & NONE & 180 & 0.3280 & 0.8111 & 0.6672 & 0.2754 \\\arrayrulecolor{gray!50}\cline{3-10}
    &  & Few-shot & Single & 1\_2\_3\_BCNF & 180 & 0.2558 & 0.7444 & 0.6352 & 0.2507 \\\arrayrulecolor{gray!30}\cline{5-10}
    &  &  &  & 2\_3\_BCNF & 180 & 0.4347 & 0.7667 & 0.6759 & 0.5346 \\\arrayrulecolor{gray!30}\cline{5-10}
    &  &  &  & 3\_BCNF & 180 & 0.2818 & 0.9222 & 0.6091 & 0.2080 \\\arrayrulecolor{gray!30}\cline{5-10}
    &  &  &  & BCNF & 180 & 0.3265 & 0.8556 & 0.5341 & 0.3641 \\\arrayrulecolor{gray!30}\cline{5-10}
    &  &  &  & NONE & 180 & 0.6783 & 0.9556 & 0.8447 & 0.6494 \\\arrayrulecolor{gray!40}\cline{4-10}
    &  &  & Complex & 1\_2\_3\_BCNF & 180 & 0.3984 & 0.6778 & 0.6155 & 0.5644 \\\arrayrulecolor{gray!30}\cline{5-10}
    &  &  &  & 2\_3\_BCNF & 180 & 0.4332 & 0.6944 & 0.6709 & 0.5767 \\\arrayrulecolor{gray!30}\cline{5-10}
    &  &  &  & 3\_BCNF & 180 & 0.4559 & 0.8556 & 0.6652 & 0.4356 \\\arrayrulecolor{gray!30}\cline{5-10}
    &  &  &  & BCNF & 180 & 0.3455 & 0.8333 & 0.5144 & 0.3931 \\\arrayrulecolor{gray!30}\cline{5-10}
    &  &  &  & NONE & 180 & 0.6715 & 0.9500 & 0.8286 & 0.6356 \\\arrayrulecolor{gray!40}\cline{4-10}
    &  &  & FD-withheld & 1\_2\_3\_BCNF & 180 & 0.2289 & 0.5056 & 0.5827 & 0.3787 \\\arrayrulecolor{gray!30}\cline{5-10}
    &  &  &  & 2\_3\_BCNF & 180 & 0.2749 & 0.5778 & 0.6399 & 0.3883 \\\arrayrulecolor{gray!30}\cline{5-10}
    &  &  &  & 3\_BCNF & 180 & 0.1799 & 0.6722 & 0.6034 & 0.1506 \\\arrayrulecolor{gray!30}\cline{5-10}
    &  &  &  & BCNF & 180 & 0.0845 & 0.5444 & 0.4994 & 0.0806 \\\arrayrulecolor{gray!30}\cline{5-10}
    &  &  &  & NONE & 180 & 0.5346 & 0.9056 & 0.7808 & 0.5037 \\\arrayrulecolor{gray!60}\cline{2-10}
    & Qwen3-30B-A3B-Instruct-2507 & Zero-shot & Single & 1\_2\_3\_BCNF & 180 & 0.5777 & 0.9889 & 0.6425 & 0.5854 \\\arrayrulecolor{gray!30}\cline{5-10}
    &  &  &  & 2\_3\_BCNF & 180 & 0.5832 & 0.9833 & 0.6577 & 0.5670 \\\arrayrulecolor{gray!30}\cline{5-10}
    &  &  &  & 3\_BCNF & 180 & 0.1758 & 0.9944 & 0.5577 & 0.1202 \\\arrayrulecolor{gray!30}\cline{5-10}
    &  &  &  & BCNF & 180 & 0.1561 & 1.0000 & 0.5517 & 0.1035 \\\arrayrulecolor{gray!30}\cline{5-10}
    &  &  &  & NONE & 180 & 0.9540 & 1.0000 & 0.9967 & 0.9159 \\\arrayrulecolor{gray!40}\cline{4-10}
    &  &  & Complex & 1\_2\_3\_BCNF & 180 & 0.2493 & 0.4111 & 0.5950 & 0.6822 \\\arrayrulecolor{gray!30}\cline{5-10}
    &  &  &  & 2\_3\_BCNF & 180 & 0.2716 & 0.4778 & 0.6238 & 0.5782 \\\arrayrulecolor{gray!30}\cline{5-10}
    &  &  &  & 3\_BCNF & 180 & 0.3368 & 0.6944 & 0.5867 & 0.4211 \\\arrayrulecolor{gray!30}\cline{5-10}
    &  &  &  & BCNF & 180 & 0.2481 & 0.6556 & 0.5474 & 0.2781 \\\arrayrulecolor{gray!30}\cline{5-10}
    &  &  &  & NONE & 180 & 0.6537 & 0.9167 & 0.7991 & 0.6263 \\\arrayrulecolor{gray!40}\cline{4-10}
    &  &  & FD-withheld & 1\_2\_3\_BCNF & 180 & 0.1519 & 0.2944 & 0.5711 & 0.5760 \\\arrayrulecolor{gray!30}\cline{5-10}
    &  &  &  & 2\_3\_BCNF & 180 & 0.1442 & 0.2611 & 0.6185 & 0.5311 \\\arrayrulecolor{gray!30}\cline{5-10}
    &  &  &  & 3\_BCNF & 180 & 0.2476 & 0.5167 & 0.5845 & 0.3968 \\\arrayrulecolor{gray!30}\cline{5-10}
    &  &  &  & BCNF & 180 & 0.1375 & 0.4611 & 0.5093 & 0.2348 \\\arrayrulecolor{gray!30}\cline{5-10}
    &  &  &  & NONE & 180 & 0.6062 & 0.9111 & 0.7974 & 0.5707 \\\arrayrulecolor{gray!50}\cline{3-10}
    &  & Few-shot & Single & 1\_2\_3\_BCNF & 180 & 0.2653 & 0.8556 & 0.5327 & 0.2565 \\\arrayrulecolor{gray!30}\cline{5-10}
    &  &  &  & 2\_3\_BCNF & 180 & 0.7130 & 0.8778 & 0.8164 & 0.8680 \\\arrayrulecolor{gray!30}\cline{5-10}
    &  &  &  & 3\_BCNF & 180 & 0.2848 & 0.7222 & 0.5312 & 0.3913 \\\arrayrulecolor{gray!30}\cline{5-10}
    &  &  &  & BCNF & 180 & 0.1546 & 0.7278 & 0.4176 & 0.2776 \\\arrayrulecolor{gray!30}\cline{5-10}
    &  &  &  & NONE & 180 & 0.6909 & 0.9611 & 0.8282 & 0.6683 \\\arrayrulecolor{gray!40}\cline{4-10}
    &  &  & Complex & 1\_2\_3\_BCNF & 180 & 0.2535 & 0.3389 & 0.7216 & 0.6893 \\\arrayrulecolor{gray!30}\cline{5-10}
    &  &  &  & 2\_3\_BCNF & 180 & 0.2793 & 0.3833 & 0.8101 & 0.6039 \\\arrayrulecolor{gray!30}\cline{5-10}
    &  &  &  & 3\_BCNF & 180 & 0.4625 & 0.7278 & 0.7792 & 0.4946 \\\arrayrulecolor{gray!30}\cline{5-10}
    &  &  &  & BCNF & 180 & 0.3002 & 0.6278 & 0.6160 & 0.3501 \\\arrayrulecolor{gray!30}\cline{5-10}
    &  &  &  & NONE & 180 & 0.6758 & 0.9167 & 0.8031 & 0.6663 \\\arrayrulecolor{gray!40}\cline{4-10}
    &  &  & FD-withheld & 1\_2\_3\_BCNF & 180 & 0.1720 & 0.2722 & 0.6032 & 0.6201 \\\arrayrulecolor{gray!30}\cline{5-10}
    &  &  &  & 2\_3\_BCNF & 180 & 0.1820 & 0.2778 & 0.7107 & 0.5560 \\\arrayrulecolor{gray!30}\cline{5-10}
    &  &  &  & 3\_BCNF & 180 & 0.3556 & 0.6222 & 0.6417 & 0.4923 \\\arrayrulecolor{gray!30}\cline{5-10}
    &  &  &  & BCNF & 180 & 0.1777 & 0.6389 & 0.4899 & 0.2398 \\\arrayrulecolor{gray!30}\cline{5-10}
    &  &  &  & NONE & 180 & 0.6052 & 0.9167 & 0.7858 & 0.5778 \\\arrayrulecolor{gray!60}\cline{2-10}
    & Mixtral-8x7B-Instruct-v0.1 & Zero-shot & Single & 1\_2\_3\_BCNF & 180 & 0.0544 & 0.3500 & 0.3868 & 0.0894 \\\arrayrulecolor{gray!30}\cline{5-10}
    &  &  &  & 2\_3\_BCNF & 180 & 0.1334 & 0.3500 & 0.4471 & 0.3052 \\\arrayrulecolor{gray!30}\cline{5-10}
    &  &  &  & 3\_BCNF & 180 & 0.1373 & 0.5389 & 0.4804 & 0.1292 \\\arrayrulecolor{gray!30}\cline{5-10}
    &  &  &  & BCNF & 180 & 0.1917 & 0.6389 & 0.4340 & 0.2701 \\\arrayrulecolor{gray!30}\cline{5-10}
    &  &  &  & NONE & 180 & 0.1502 & 0.5278 & 0.3812 & 0.2048 \\\arrayrulecolor{gray!40}\cline{4-10}
    &  &  & Complex & 1\_2\_3\_BCNF & 180 & 0.1428 & 0.3167 & 0.3015 & 0.4559 \\\arrayrulecolor{gray!30}\cline{5-10}
    &  &  &  & 2\_3\_BCNF & 180 & 0.1257 & 0.3056 & 0.3580 & 0.3869 \\\arrayrulecolor{gray!30}\cline{5-10}
    &  &  &  & 3\_BCNF & 180 & 0.1696 & 0.4444 & 0.4540 & 0.3086 \\\arrayrulecolor{gray!30}\cline{5-10}
    &  &  &  & BCNF & 180 & 0.1404 & 0.4833 & 0.3460 & 0.2381 \\\arrayrulecolor{gray!30}\cline{5-10}
    &  &  &  & NONE & 180 & 0.1276 & 0.5000 & 0.3096 & 0.2463 \\\arrayrulecolor{gray!40}\cline{4-10}
    &  &  & FD-withheld & 1\_2\_3\_BCNF & 180 & 0.0438 & 0.1278 & 0.3355 & 0.2994 \\\arrayrulecolor{gray!30}\cline{5-10}
    &  &  &  & 2\_3\_BCNF & 180 & 0.0278 & 0.0889 & 0.3797 & 0.2492 \\\arrayrulecolor{gray!30}\cline{5-10}
    &  &  &  & 3\_BCNF & 180 & 0.1146 & 0.3778 & 0.4780 & 0.1922 \\\arrayrulecolor{gray!30}\cline{5-10}
    &  &  &  & BCNF & 180 & 0.0927 & 0.3944 & 0.4051 & 0.1691 \\\arrayrulecolor{gray!30}\cline{5-10}
    &  &  &  & NONE & 180 & 0.3358 & 0.7611 & 0.5916 & 0.3831 \\\arrayrulecolor{gray!50}\cline{3-10}
    &  & Few-shot & Single & 1\_2\_3\_BCNF & 180 & 0.0895 & 0.3944 & 0.5294 & 0.1347 \\\arrayrulecolor{gray!30}\cline{5-10}
    &  &  &  & 2\_3\_BCNF & 180 & 0.3070 & 0.4778 & 0.7017 & 0.4624 \\\arrayrulecolor{gray!30}\cline{5-10}
    &  &  &  & 3\_BCNF & 180 & 0.0980 & 0.5056 & 0.5225 & 0.1271 \\\arrayrulecolor{gray!30}\cline{5-10}
    &  &  &  & BCNF & 180 & 0.1543 & 0.4778 & 0.4889 & 0.3415 \\\arrayrulecolor{gray!30}\cline{5-10}
    &  &  &  & NONE & 180 & 0.2503 & 0.8500 & 0.6332 & 0.1931 \\\arrayrulecolor{gray!40}\cline{4-10}
    &  &  & Complex & 1\_2\_3\_BCNF & 180 & 0.1187 & 0.2167 & 0.5529 & 0.4936 \\\arrayrulecolor{gray!30}\cline{5-10}
    &  &  &  & 2\_3\_BCNF & 180 & 0.0900 & 0.1611 & 0.5910 & 0.4915 \\\arrayrulecolor{gray!30}\cline{5-10}
    &  &  &  & 3\_BCNF & 180 & 0.1373 & 0.2722 & 0.5592 & 0.4509 \\\arrayrulecolor{gray!30}\cline{5-10}
    &  &  &  & BCNF & 180 & 0.1399 & 0.2889 & 0.5122 & 0.5133 \\\arrayrulecolor{gray!30}\cline{5-10}
    &  &  &  & NONE & 180 & 0.3807 & 0.8778 & 0.7202 & 0.3387 \\\arrayrulecolor{gray!40}\cline{4-10}
    &  &  & FD-withheld & 1\_2\_3\_BCNF & 180 & 0.0400 & 0.0778 & 0.5257 & 0.4634 \\\arrayrulecolor{gray!30}\cline{5-10}
    &  &  &  & 2\_3\_BCNF & 180 & 0.0471 & 0.0833 & 0.5504 & 0.4924 \\\arrayrulecolor{gray!30}\cline{5-10}
    &  &  &  & 3\_BCNF & 180 & 0.1378 & 0.2722 & 0.5421 & 0.4341 \\\arrayrulecolor{gray!30}\cline{5-10}
    &  &  &  & BCNF & 180 & 0.0697 & 0.1722 & 0.4583 & 0.4644 \\\arrayrulecolor{gray!30}\cline{5-10}
    &  &  &  & NONE & 180 & 0.2783 & 0.8222 & 0.6832 & 0.2468 \\\arrayrulecolor{gray!60}\cline{2-10}
    & Llama-3.3-70B-Instruct & Zero-shot & Single & 1\_2\_3\_BCNF & 180 & 0.3170 & 0.6889 & 0.6272 & 0.3750 \\\arrayrulecolor{gray!30}\cline{5-10}
    &  &  &  & 2\_3\_BCNF & 180 & 0.5830 & 0.7611 & 0.7974 & 0.6935 \\\arrayrulecolor{gray!30}\cline{5-10}
    &  &  &  & 3\_BCNF & 180 & 0.2031 & 0.7500 & 0.6259 & 0.1293 \\\arrayrulecolor{gray!30}\cline{5-10}
    &  &  &  & BCNF & 180 & 0.1771 & 0.6889 & 0.5434 & 0.1954 \\\arrayrulecolor{gray!30}\cline{5-10}
    &  &  &  & NONE & 180 & 0.6835 & 0.8667 & 0.8752 & 0.7685 \\\arrayrulecolor{gray!40}\cline{4-10}
    &  &  & Complex & 1\_2\_3\_BCNF & 180 & 0.3249 & 0.5778 & 0.6338 & 0.5224 \\\arrayrulecolor{gray!30}\cline{5-10}
    &  &  &  & 2\_3\_BCNF & 180 & 0.4038 & 0.7056 & 0.6735 & 0.5048 \\\arrayrulecolor{gray!30}\cline{5-10}
    &  &  &  & 3\_BCNF & 180 & 0.3822 & 0.7167 & 0.6332 & 0.4422 \\\arrayrulecolor{gray!30}\cline{5-10}
    &  &  &  & BCNF & 180 & 0.2578 & 0.6167 & 0.5388 & 0.3818 \\\arrayrulecolor{gray!30}\cline{5-10}
    &  &  &  & NONE & 180 & 0.6282 & 0.8222 & 0.8413 & 0.7391 \\\arrayrulecolor{gray!40}\cline{4-10}
    &  &  & FD-withheld & 1\_2\_3\_BCNF & 180 & 0.1363 & 0.2611 & 0.5912 & 0.4716 \\\arrayrulecolor{gray!30}\cline{5-10}
    &  &  &  & 2\_3\_BCNF & 180 & 0.1948 & 0.3944 & 0.6461 & 0.3773 \\\arrayrulecolor{gray!30}\cline{5-10}
    &  &  &  & 3\_BCNF & 180 & 0.1254 & 0.3889 & 0.5947 & 0.2182 \\\arrayrulecolor{gray!30}\cline{5-10}
    &  &  &  & BCNF & 180 & 0.0805 & 0.3556 & 0.5118 & 0.1444 \\\arrayrulecolor{gray!30}\cline{5-10}
    &  &  &  & NONE & 180 & 0.5344 & 0.8389 & 0.7293 & 0.5483 \\\arrayrulecolor{gray!50}\cline{3-10}
    &  & Few-shot & Single & 1\_2\_3\_BCNF & 180 & 0.3684 & 0.6056 & 0.6102 & 0.6239 \\\arrayrulecolor{gray!30}\cline{5-10}
    &  &  &  & 2\_3\_BCNF & 180 & 0.6147 & 0.7944 & 0.7332 & 0.7426 \\\arrayrulecolor{gray!30}\cline{5-10}
    &  &  &  & 3\_BCNF & 180 & 0.2993 & 0.6667 & 0.6469 & 0.3504 \\\arrayrulecolor{gray!30}\cline{5-10}
    &  &  &  & BCNF & 180 & 0.1599 & 0.7000 & 0.4969 & 0.2111 \\\arrayrulecolor{gray!30}\cline{5-10}
    &  &  &  & NONE & 180 & 0.7853 & 0.9778 & 0.8947 & 0.7633 \\\arrayrulecolor{gray!40}\cline{4-10}
    &  &  & Complex & 1\_2\_3\_BCNF & 180 & 0.4547 & 0.6000 & 0.7824 & 0.6818 \\\arrayrulecolor{gray!30}\cline{5-10}
    &  &  &  & 2\_3\_BCNF & 180 & 0.5528 & 0.7611 & 0.7992 & 0.6286 \\\arrayrulecolor{gray!30}\cline{5-10}
    &  &  &  & 3\_BCNF & 180 & 0.3916 & 0.6333 & 0.7037 & 0.5431 \\\arrayrulecolor{gray!30}\cline{5-10}
    &  &  &  & BCNF & 180 & 0.2794 & 0.6611 & 0.5499 & 0.3302 \\\arrayrulecolor{gray!30}\cline{5-10}
    &  &  &  & NONE & 180 & 0.7462 & 0.9444 & 0.8753 & 0.7785 \\\arrayrulecolor{gray!40}\cline{4-10}
    &  &  & FD-withheld & 1\_2\_3\_BCNF & 180 & 0.2292 & 0.3611 & 0.6346 & 0.5964 \\\arrayrulecolor{gray!30}\cline{5-10}
    &  &  &  & 2\_3\_BCNF & 180 & 0.2746 & 0.4500 & 0.6572 & 0.5410 \\\arrayrulecolor{gray!30}\cline{5-10}
    &  &  &  & 3\_BCNF & 180 & 0.2666 & 0.4778 & 0.6320 & 0.4601 \\\arrayrulecolor{gray!30}\cline{5-10}
    &  &  &  & BCNF & 180 & 0.1324 & 0.4278 & 0.5003 & 0.2557 \\\arrayrulecolor{gray!30}\cline{5-10}
    &  &  &  & NONE & 180 & 0.3576 & 0.6056 & 0.7929 & 0.6219 \\\arrayrulecolor{black}\hline
  \end{xltabular}
}

\clearpage


\begin{center}
  {\scriptsize
    \setlength{\tabcolsep}{2pt}
    \renewcommand{\arraystretch}{1.15}
    \setlength{\LTleft}{0pt}
    \setlength{\LTright}{0pt}
    \setlength{\LTcapwidth}{\linewidth}
    \captionsetup{hypcap=false,position=below,skip=6pt}
    \begin{xltabular}{\linewidth}{|l|l|l|l|>{\centering\arraybackslash}X|>{\centering\arraybackslash}X|>{\centering\arraybackslash}X|>{\centering\arraybackslash}X|>{\centering\arraybackslash}X|}
      \hline
      \multirow{2}{*}{\textbf{Dataset}} & \multirow{2}{*}{\textbf{Model}} & \multirow{2}{*}{\textbf{Shot}} & \multirow{2}{*}{\textbf{Violation Path}} & \multirow{2}{*}{\textbf{Num.}} & \multirow{2}{*}{\textbf{{\DNS}}} & \multicolumn{3}{c|}{\textbf{Components}} \\
      \cline{7-9}
      & & & & & & \textbf{Semantic} & \textbf{Structural} & \textbf{Logical} \\
      \hline
      \endfirsthead

      \hline
      \multirow{2}{*}{\textbf{Dataset}} & \multirow{2}{*}{\textbf{Model}} & \multirow{2}{*}{\textbf{Shot}} & \multirow{2}{*}{\textbf{Violation Path}} & \multirow{2}{*}{\textbf{Num.}} & \multirow{2}{*}{\textbf{{\DNS}}} & \multicolumn{3}{c|}{\textbf{Components}} \\
      \cline{7-9}
      & & & & & & \textbf{Semantic} & \textbf{Structural} & \textbf{Logical} \\
      \hline
      \endhead

      \endfoot

      \noalign{\vskip 6pt}
      \caption{\textbf{MARS} evaluation matrix: {\DNS} and its three components for every (dataset, model, prompting setting, violation path) cell in the \emph{FD-withheld} setting.}\label{tab:multi-agent_full_matrix}
      \endlastfoot

      BIRD & Gemma-3-27B-it & Zero-shot & 1\_2\_3\_BCNF & 77 & 0.3227 & 0.7013 & 0.5943 & 0.4104 \\\arrayrulecolor{gray!30}\cline{4-9}
      &  &  & 2\_3\_BCNF & 77 & 0.3621 & 0.7532 & 0.6442 & 0.4169 \\\arrayrulecolor{gray!30}\cline{4-9}
      &  &  & 3\_BCNF & 77 & 0.3746 & 0.7922 & 0.6492 & 0.4108 \\\arrayrulecolor{gray!30}\cline{4-9}
      &  &  & BCNF & 77 & 0.2311 & 0.7013 & 0.5561 & 0.2660 \\\arrayrulecolor{gray!30}\cline{4-9}
      &  &  & NONE & 77 & 0.7093 & 0.7532 & 0.9392 & 0.7095 \\\arrayrulecolor{gray!50}\cline{3-9}
      &  & Few-shot & 1\_2\_3\_BCNF & 77 & 0.4027 & 0.8182 & 0.6172 & 0.4079 \\\arrayrulecolor{gray!30}\cline{4-9}
      &  &  & 2\_3\_BCNF & 77 & 0.3580 & 0.7273 & 0.6606 & 0.4071 \\\arrayrulecolor{gray!30}\cline{4-9}
      &  &  & 3\_BCNF & 77 & 0.4666 & 0.8961 & 0.6951 & 0.4268 \\\arrayrulecolor{gray!30}\cline{4-9}
      &  &  & BCNF & 77 & 0.2157 & 0.7403 & 0.5646 & 0.2195 \\\arrayrulecolor{gray!30}\cline{4-9}
      &  &  & NONE & 77 & 0.6728 & 0.7143 & 0.8685 & 0.6719 \\\arrayrulecolor{gray!60}\cline{2-9}
      & Qwen3-30B-A3B-Instruct-2507 & Zero-shot & 1\_2\_3\_BCNF & 77 & 0.3618 & 0.6753 & 0.5519 & 0.5081 \\\arrayrulecolor{gray!30}\cline{4-9}
      &  &  & 2\_3\_BCNF & 77 & 0.4201 & 0.8571 & 0.6698 & 0.4299 \\\arrayrulecolor{gray!30}\cline{4-9}
      &  &  & 3\_BCNF & 77 & 0.3834 & 0.8701 & 0.5938 & 0.3892 \\\arrayrulecolor{gray!30}\cline{4-9}
      &  &  & BCNF & 77 & 0.2560 & 0.9091 & 0.5408 & 0.1978 \\\arrayrulecolor{gray!30}\cline{4-9}
      &  &  & NONE & 77 & 0.9309 & 0.9610 & 0.9793 & 0.9208 \\\arrayrulecolor{gray!50}\cline{3-9}
      &  & Few-shot & 1\_2\_3\_BCNF & 77 & 0.3537 & 0.6623 & 0.5649 & 0.4952 \\\arrayrulecolor{gray!30}\cline{4-9}
      &  &  & 2\_3\_BCNF & 77 & 0.3892 & 0.6883 & 0.6906 & 0.4956 \\\arrayrulecolor{gray!30}\cline{4-9}
      &  &  & 3\_BCNF & 77 & 0.3436 & 0.8182 & 0.5654 & 0.3801 \\\arrayrulecolor{gray!30}\cline{4-9}
      &  &  & BCNF & 77 & 0.2740 & 0.8831 & 0.5517 & 0.2361 \\\arrayrulecolor{gray!30}\cline{4-9}
      &  &  & NONE & 77 & 0.9520 & 1.0000 & 0.9677 & 0.9463 \\\arrayrulecolor{gray!60}\cline{2-9}
      & Mixtral-8x7B-Instruct-v0.1 & Zero-shot & 1\_2\_3\_BCNF & 77 & 0.2435 & 0.8571 & 0.5128 & 0.2101 \\\arrayrulecolor{gray!30}\cline{4-9}
      &  &  & 2\_3\_BCNF & 77 & 0.2320 & 0.8831 & 0.5567 & 0.1841 \\\arrayrulecolor{gray!30}\cline{4-9}
      &  &  & 3\_BCNF & 77 & 0.2726 & 0.8831 & 0.5217 & 0.2095 \\\arrayrulecolor{gray!30}\cline{4-9}
      &  &  & BCNF & 77 & 0.3076 & 0.9740 & 0.5485 & 0.2173 \\\arrayrulecolor{gray!30}\cline{4-9}
      &  &  & NONE & 77 & 0.9841 & 1.0000 & 0.9957 & 0.9749 \\\arrayrulecolor{gray!50}\cline{3-9}
      &  & Few-shot & 1\_2\_3\_BCNF & 77 & 0.1379 & 0.3896 & 0.4742 & 0.3072 \\\arrayrulecolor{gray!30}\cline{4-9}
      &  &  & 2\_3\_BCNF & 77 & 0.1512 & 0.5065 & 0.5280 & 0.2495 \\\arrayrulecolor{gray!30}\cline{4-9}
      &  &  & 3\_BCNF & 77 & 0.2182 & 0.7013 & 0.5030 & 0.2390 \\\arrayrulecolor{gray!30}\cline{4-9}
      &  &  & BCNF & 77 & 0.2862 & 0.8571 & 0.5307 & 0.2372 \\\arrayrulecolor{gray!30}\cline{4-9}
      &  &  & NONE & 77 & 0.9496 & 0.9610 & 0.9683 & 0.9532 \\\arrayrulecolor{gray!60}\cline{2-9}
      & Llama-3.3-70B-Instruct & Zero-shot & 1\_2\_3\_BCNF & 77 & 0.3912 & 0.7273 & 0.6307 & 0.4933 \\\arrayrulecolor{gray!30}\cline{4-9}
      &  &  & 2\_3\_BCNF & 77 & 0.3242 & 0.8442 & 0.6160 & 0.3047 \\\arrayrulecolor{gray!30}\cline{4-9}
      &  &  & 3\_BCNF & 77 & 0.3297 & 0.9221 & 0.5537 & 0.2861 \\\arrayrulecolor{gray!30}\cline{4-9}
      &  &  & BCNF & 77 & 0.3344 & 0.9870 & 0.5590 & 0.2381 \\\arrayrulecolor{gray!30}\cline{4-9}
      &  &  & NONE & 77 & 0.9350 & 0.9610 & 0.9668 & 0.9286 \\\arrayrulecolor{gray!50}\cline{3-9}
      &  & Few-shot & 1\_2\_3\_BCNF & 77 & 0.4036 & 0.7273 & 0.6143 & 0.5115 \\\arrayrulecolor{gray!30}\cline{4-9}
      &  &  & 2\_3\_BCNF & 77 & 0.3404 & 0.8571 & 0.6080 & 0.3246 \\\arrayrulecolor{gray!30}\cline{4-9}
      &  &  & 3\_BCNF & 77 & 0.3344 & 0.9221 & 0.5367 & 0.2935 \\\arrayrulecolor{gray!30}\cline{4-9}
      &  &  & BCNF & 77 & 0.3194 & 0.9870 & 0.5590 & 0.2286 \\\arrayrulecolor{gray!30}\cline{4-9}
      &  &  & NONE & 77 & 0.9107 & 0.9610 & 0.9509 & 0.8996 \\\arrayrulecolor{black}\hline
      Spider & Gemma-3-27B-it & Zero-shot & 1\_2\_3\_BCNF & 180 & 0.4029 & 0.7944 & 0.6180 & 0.4244 \\\arrayrulecolor{gray!30}\cline{4-9}
      &  &  & 2\_3\_BCNF & 180 & 0.4541 & 0.8944 & 0.6592 & 0.4289 \\\arrayrulecolor{gray!30}\cline{4-9}
      &  &  & 3\_BCNF & 180 & 0.4919 & 0.9056 & 0.7214 & 0.4507 \\\arrayrulecolor{gray!30}\cline{4-9}
      &  &  & BCNF & 180 & 0.2498 & 0.8833 & 0.5603 & 0.2098 \\\arrayrulecolor{gray!30}\cline{4-9}
      &  &  & NONE & 180 & 0.7860 & 0.8389 & 0.9569 & 0.7769 \\\arrayrulecolor{gray!50}\cline{3-9}
      &  & Few-shot & 1\_2\_3\_BCNF & 180 & 0.4236 & 0.8333 & 0.6179 & 0.4193 \\\arrayrulecolor{gray!30}\cline{4-9}
      &  &  & 2\_3\_BCNF & 180 & 0.4304 & 0.8333 & 0.6841 & 0.4166 \\\arrayrulecolor{gray!30}\cline{4-9}
      &  &  & 3\_BCNF & 180 & 0.4935 & 0.9111 & 0.7445 & 0.4291 \\\arrayrulecolor{gray!30}\cline{4-9}
      &  &  & BCNF & 180 & 0.2872 & 0.9000 & 0.5646 & 0.2363 \\\arrayrulecolor{gray!30}\cline{4-9}
      &  &  & NONE & 180 & 0.8213 & 0.8556 & 0.9349 & 0.8235 \\\arrayrulecolor{gray!60}\cline{2-9}
      & Qwen3-30B-A3B-Instruct-2507 & Zero-shot & 1\_2\_3\_BCNF & 180 & 0.4285 & 0.7778 & 0.6105 & 0.5133 \\\arrayrulecolor{gray!30}\cline{4-9}
      &  &  & 2\_3\_BCNF & 180 & 0.4446 & 0.8889 & 0.6897 & 0.4138 \\\arrayrulecolor{gray!30}\cline{4-9}
      &  &  & 3\_BCNF & 180 & 0.4297 & 0.9222 & 0.6060 & 0.4067 \\\arrayrulecolor{gray!30}\cline{4-9}
      &  &  & BCNF & 180 & 0.2784 & 0.9556 & 0.5409 & 0.2119 \\\arrayrulecolor{gray!30}\cline{4-9}
      &  &  & NONE & 180 & 0.9717 & 0.9889 & 0.9894 & 0.9648 \\\arrayrulecolor{gray!50}\cline{3-9}
      &  & Few-shot & 1\_2\_3\_BCNF & 180 & 0.4324 & 0.7444 & 0.6053 & 0.5336 \\\arrayrulecolor{gray!30}\cline{4-9}
      &  &  & 2\_3\_BCNF & 180 & 0.4460 & 0.8500 & 0.6796 & 0.4431 \\\arrayrulecolor{gray!30}\cline{4-9}
      &  &  & 3\_BCNF & 180 & 0.4083 & 0.9056 & 0.6048 & 0.4004 \\\arrayrulecolor{gray!30}\cline{4-9}
      &  &  & BCNF & 180 & 0.2939 & 0.9722 & 0.5400 & 0.2191 \\\arrayrulecolor{gray!30}\cline{4-9}
      &  &  & NONE & 180 & 0.9728 & 0.9889 & 0.9869 & 0.9683 \\\arrayrulecolor{gray!60}\cline{2-9}
      & Mixtral-8x7B-Instruct-v0.1 & Zero-shot & 1\_2\_3\_BCNF & 180 & 0.2461 & 0.7667 & 0.5035 & 0.2205 \\\arrayrulecolor{gray!30}\cline{4-9}
      &  &  & 2\_3\_BCNF & 180 & 0.2573 & 0.8611 & 0.5768 & 0.1984 \\\arrayrulecolor{gray!30}\cline{4-9}
      &  &  & 3\_BCNF & 180 & 0.2817 & 0.9333 & 0.5507 & 0.2011 \\\arrayrulecolor{gray!30}\cline{4-9}
      &  &  & BCNF & 180 & 0.2949 & 0.9722 & 0.5451 & 0.2081 \\\arrayrulecolor{gray!30}\cline{4-9}
      &  &  & NONE & 180 & 0.9757 & 0.9944 & 0.9921 & 0.9683 \\\arrayrulecolor{gray!50}\cline{3-9}
      &  & Few-shot & 1\_2\_3\_BCNF & 180 & 0.1245 & 0.3556 & 0.5393 & 0.3317 \\\arrayrulecolor{gray!30}\cline{4-9}
      &  &  & 2\_3\_BCNF & 180 & 0.1125 & 0.3167 & 0.5806 & 0.3100 \\\arrayrulecolor{gray!30}\cline{4-9}
      &  &  & 3\_BCNF & 180 & 0.2429 & 0.6722 & 0.5358 & 0.2695 \\\arrayrulecolor{gray!30}\cline{4-9}
      &  &  & BCNF & 180 & 0.2466 & 0.7667 & 0.5434 & 0.2331 \\\arrayrulecolor{gray!30}\cline{4-9}
      &  &  & NONE & 180 & 0.9451 & 0.9833 & 0.9667 & 0.9493 \\\arrayrulecolor{gray!60}\cline{2-9}
      & Llama-3.3-70B-Instruct & Zero-shot & 1\_2\_3\_BCNF & 180 & 0.4775 & 0.7889 & 0.6787 & 0.5406 \\\arrayrulecolor{gray!30}\cline{4-9}
      &  &  & 2\_3\_BCNF & 180 & 0.3780 & 0.9444 & 0.6313 & 0.2988 \\\arrayrulecolor{gray!30}\cline{4-9}
      &  &  & 3\_BCNF & 180 & 0.3607 & 0.9556 & 0.5754 & 0.2930 \\\arrayrulecolor{gray!30}\cline{4-9}
      &  &  & BCNF & 180 & 0.3440 & 0.9944 & 0.5524 & 0.2419 \\\arrayrulecolor{gray!30}\cline{4-9}
      &  &  & NONE & 180 & 0.9766 & 1.0000 & 0.9898 & 0.9713 \\\arrayrulecolor{gray!50}\cline{3-9}
      &  & Few-shot & 1\_2\_3\_BCNF & 180 & 0.4647 & 0.7556 & 0.6592 & 0.5693 \\\arrayrulecolor{gray!30}\cline{4-9}
      &  &  & 2\_3\_BCNF & 180 & 0.3963 & 0.9333 & 0.6354 & 0.3347 \\\arrayrulecolor{gray!30}\cline{4-9}
      &  &  & 3\_BCNF & 180 & 0.3858 & 0.9389 & 0.5780 & 0.3333 \\\arrayrulecolor{gray!30}\cline{4-9}
      &  &  & BCNF & 180 & 0.3355 & 0.9944 & 0.5503 & 0.2470 \\\arrayrulecolor{gray!30}\cline{4-9}
      &  &  & NONE & 180 & 0.9710 & 0.9944 & 0.9887 & 0.9657 \\\arrayrulecolor{black}\hline
    \end{xltabular}
  }
\end{center}

\begin{center}
  {\scriptsize
    \setlength{\tabcolsep}{2pt}
    \renewcommand{\arraystretch}{1.15}
    \setlength{\LTleft}{0pt}
    \setlength{\LTright}{0pt}
    \setlength{\LTcapwidth}{\linewidth}
    \captionsetup{hypcap=false,position=below,skip=6pt}
    \begin{xltabular}{\linewidth}{|l|l|l|l|>{\centering\arraybackslash}X|>{\centering\arraybackslash}X|>{\centering\arraybackslash}X|>{\centering\arraybackslash}X|>{\centering\arraybackslash}X|}
      \hline
      \multirow{2}{*}{\textbf{Dataset}} & \multirow{2}{*}{\textbf{Model}} & \multirow{2}{*}{\textbf{Shot}} & \multirow{2}{*}{\textbf{Violation Path}} & \multirow{2}{*}{\textbf{Num.}} & \multirow{2}{*}{\textbf{{\DNS}}} & \multicolumn{3}{c|}{\textbf{Components}} \\
      \cline{7-9}
      & & & & & & \textbf{Semantic} & \textbf{Structural} & \textbf{Logical} \\
      \hline
      \endfirsthead

      \hline
      \multirow{2}{*}{\textbf{Dataset}} & \multirow{2}{*}{\textbf{Model}} & \multirow{2}{*}{\textbf{Shot}} & \multirow{2}{*}{\textbf{Violation Path}} & \multirow{2}{*}{\textbf{Num.}} & \multirow{2}{*}{\textbf{{\DNS}}} & \multicolumn{3}{c|}{\textbf{Components}} \\
      \cline{7-9}
      & & & & & & \textbf{Semantic} & \textbf{Structural} & \textbf{Logical} \\
      \hline
      \endhead

      \endfoot

      \noalign{\vskip 6pt}
      \caption{\textbf{Miffie} evaluation matrix: {\DNS} and its three components for every (dataset, model, prompting setting, violation path) cell in the \emph{FD-withheld} setting.}\label{tab:miffie_full_matrix}
      \endlastfoot

      BIRD & Gemma-3-27B-it & Zero-shot & 1\_2\_3\_BCNF & 77 & 0.2436 & 0.5455 & 0.5173 & 0.4949 \\\arrayrulecolor{gray!30}\cline{4-9}
      &  &  & 2\_3\_BCNF & 77 & 0.1865 & 0.3896 & 0.5761 & 0.4174 \\\arrayrulecolor{gray!30}\cline{4-9}
      &  &  & 3\_BCNF & 77 & 0.1679 & 0.4935 & 0.5865 & 0.2515 \\\arrayrulecolor{gray!30}\cline{4-9}
      &  &  & BCNF & 77 & 0.1183 & 0.5584 & 0.5257 & 0.1281 \\\arrayrulecolor{gray!30}\cline{4-9}
      &  &  & NONE & 77 & 0.3387 & 0.8312 & 0.7026 & 0.2740 \\\arrayrulecolor{gray!50}\cline{3-9}
      &  & Few-shot & 1\_2\_3\_BCNF & 77 & 0.1850 & 0.3766 & 0.5580 & 0.4049 \\\arrayrulecolor{gray!30}\cline{4-9}
      &  &  & 2\_3\_BCNF & 77 & 0.2159 & 0.4156 & 0.5937 & 0.4290 \\\arrayrulecolor{gray!30}\cline{4-9}
      &  &  & 3\_BCNF & 77 & 0.1953 & 0.6494 & 0.6038 & 0.2193 \\\arrayrulecolor{gray!30}\cline{4-9}
      &  &  & BCNF & 77 & 0.1146 & 0.5455 & 0.5120 & 0.1541 \\\arrayrulecolor{gray!30}\cline{4-9}
      &  &  & NONE & 77 & 0.4848 & 0.7403 & 0.7554 & 0.5113 \\\arrayrulecolor{gray!60}\cline{2-9}
      & Qwen3-30B-A3B-Instruct-2507 & Zero-shot & 1\_2\_3\_BCNF & 77 & 0.1892 & 0.3766 & 0.5633 & 0.5635 \\\arrayrulecolor{gray!30}\cline{4-9}
      &  &  & 2\_3\_BCNF & 77 & 0.2403 & 0.4935 & 0.6034 & 0.4547 \\\arrayrulecolor{gray!30}\cline{4-9}
      &  &  & 3\_BCNF & 77 & 0.2490 & 0.5195 & 0.5726 & 0.4279 \\\arrayrulecolor{gray!30}\cline{4-9}
      &  &  & BCNF & 77 & 0.1978 & 0.5844 & 0.5059 & 0.2781 \\\arrayrulecolor{gray!30}\cline{4-9}
      &  &  & NONE & 77 & 0.5549 & 0.8961 & 0.7646 & 0.5195 \\\arrayrulecolor{gray!50}\cline{3-9}
      &  & Few-shot & 1\_2\_3\_BCNF & 77 & 0.2244 & 0.4545 & 0.5548 & 0.4816 \\\arrayrulecolor{gray!30}\cline{4-9}
      &  &  & 2\_3\_BCNF & 77 & 0.2209 & 0.4156 & 0.6188 & 0.4395 \\\arrayrulecolor{gray!30}\cline{4-9}
      &  &  & 3\_BCNF & 77 & 0.2710 & 0.5455 & 0.5838 & 0.4310 \\\arrayrulecolor{gray!30}\cline{4-9}
      &  &  & BCNF & 77 & 0.3015 & 0.7403 & 0.5044 & 0.3959 \\\arrayrulecolor{gray!30}\cline{4-9}
      &  &  & NONE & 77 & 0.7943 & 0.8831 & 0.8958 & 0.8195 \\\arrayrulecolor{gray!60}\cline{2-9}
      & Mixtral-8x7B-Instruct-v0.1 & Zero-shot & 1\_2\_3\_BCNF & 77 & 0.0101 & 0.2857 & 0.2826 & 0.1194 \\\arrayrulecolor{gray!30}\cline{4-9}
      &  &  & 2\_3\_BCNF & 77 & 0.0092 & 0.2857 & 0.3195 & 0.1482 \\\arrayrulecolor{gray!30}\cline{4-9}
      &  &  & 3\_BCNF & 77 & 0.0383 & 0.3506 & 0.3284 & 0.1004 \\\arrayrulecolor{gray!30}\cline{4-9}
      &  &  & BCNF & 77 & 0.0331 & 0.4416 & 0.3184 & 0.0823 \\\arrayrulecolor{gray!30}\cline{4-9}
      &  &  & NONE & 77 & 0.3861 & 0.5844 & 0.5780 & 0.5074 \\\arrayrulecolor{gray!50}\cline{3-9}
      &  & Few-shot & 1\_2\_3\_BCNF & 77 & 0.0131 & 0.2857 & 0.3696 & 0.1417 \\\arrayrulecolor{gray!30}\cline{4-9}
      &  &  & 2\_3\_BCNF & 77 & 0.0226 & 0.1818 & 0.3927 & 0.1429 \\\arrayrulecolor{gray!30}\cline{4-9}
      &  &  & 3\_BCNF & 77 & 0.0829 & 0.4416 & 0.4356 & 0.1519 \\\arrayrulecolor{gray!30}\cline{4-9}
      &  &  & BCNF & 77 & 0.1031 & 0.5195 & 0.4129 & 0.1335 \\\arrayrulecolor{gray!30}\cline{4-9}
      &  &  & NONE & 77 & 0.3910 & 0.6234 & 0.6901 & 0.4459 \\\arrayrulecolor{gray!60}\cline{2-9}
      & Llama-3.3-70B-Instruct & Zero-shot & 1\_2\_3\_BCNF & 77 & 0.1259 & 0.2597 & 0.5454 & 0.4077 \\\arrayrulecolor{gray!30}\cline{4-9}
      &  &  & 2\_3\_BCNF & 77 & 0.1367 & 0.3506 & 0.5692 & 0.3257 \\\arrayrulecolor{gray!30}\cline{4-9}
      &  &  & 3\_BCNF & 77 & 0.1260 & 0.4156 & 0.5582 & 0.2457 \\\arrayrulecolor{gray!30}\cline{4-9}
      &  &  & BCNF & 77 & 0.0923 & 0.4026 & 0.4902 & 0.1643 \\\arrayrulecolor{gray!30}\cline{4-9}
      &  &  & NONE & 77 & 0.4102 & 0.6753 & 0.6469 & 0.4732 \\\arrayrulecolor{gray!50}\cline{3-9}
      &  & Few-shot & 1\_2\_3\_BCNF & 77 & 0.2255 & 0.3506 & 0.5783 & 0.5962 \\\arrayrulecolor{gray!30}\cline{4-9}
      &  &  & 2\_3\_BCNF & 77 & 0.3149 & 0.4675 & 0.7239 & 0.5808 \\\arrayrulecolor{gray!30}\cline{4-9}
      &  &  & 3\_BCNF & 77 & 0.2862 & 0.4545 & 0.6484 & 0.5232 \\\arrayrulecolor{gray!30}\cline{4-9}
      &  &  & BCNF & 77 & 0.1559 & 0.3896 & 0.5156 & 0.3316 \\\arrayrulecolor{gray!30}\cline{4-9}
      &  &  & NONE & 77 & 0.6422 & 0.8571 & 0.8044 & 0.6740 \\\arrayrulecolor{black}\hline
      Spider & Gemma-3-27B-it & Zero-shot & 1\_2\_3\_BCNF & 180 & 0.1415 & 0.2611 & 0.5731 & 0.5474 \\\arrayrulecolor{gray!30}\cline{4-9}
      &  &  & 2\_3\_BCNF & 180 & 0.1461 & 0.2833 & 0.6213 & 0.4607 \\\arrayrulecolor{gray!30}\cline{4-9}
      &  &  & 3\_BCNF & 180 & 0.1056 & 0.3333 & 0.5653 & 0.2280 \\\arrayrulecolor{gray!30}\cline{4-9}
      &  &  & BCNF & 180 & 0.0590 & 0.3444 & 0.4989 & 0.1137 \\\arrayrulecolor{gray!30}\cline{4-9}
      &  &  & NONE & 180 & 0.3274 & 0.8444 & 0.6613 & 0.2605 \\\arrayrulecolor{gray!50}\cline{3-9}
      &  & Few-shot & 1\_2\_3\_BCNF & 180 & 0.1884 & 0.4111 & 0.5856 & 0.4217 \\\arrayrulecolor{gray!30}\cline{4-9}
      &  &  & 2\_3\_BCNF & 180 & 0.2571 & 0.5000 & 0.6582 & 0.4374 \\\arrayrulecolor{gray!30}\cline{4-9}
      &  &  & 3\_BCNF & 180 & 0.1976 & 0.5833 & 0.5951 & 0.2251 \\\arrayrulecolor{gray!30}\cline{4-9}
      &  &  & BCNF & 180 & 0.0821 & 0.4778 & 0.4890 & 0.1129 \\\arrayrulecolor{gray!30}\cline{4-9}
      &  &  & NONE & 180 & 0.4842 & 0.8500 & 0.7532 & 0.4819 \\\arrayrulecolor{gray!60}\cline{2-9}
      & Qwen3-30B-A3B-Instruct-2507 & Zero-shot & 1\_2\_3\_BCNF & 180 & 0.1811 & 0.3667 & 0.5653 & 0.5219 \\\arrayrulecolor{gray!30}\cline{4-9}
      &  &  & 2\_3\_BCNF & 180 & 0.1820 & 0.3611 & 0.6215 & 0.4815 \\\arrayrulecolor{gray!30}\cline{4-9}
      &  &  & 3\_BCNF & 180 & 0.2672 & 0.5500 & 0.5854 & 0.4193 \\\arrayrulecolor{gray!30}\cline{4-9}
      &  &  & BCNF & 180 & 0.1801 & 0.5222 & 0.5142 & 0.2706 \\\arrayrulecolor{gray!30}\cline{4-9}
      &  &  & NONE & 180 & 0.6351 & 0.9278 & 0.8051 & 0.5996 \\\arrayrulecolor{gray!50}\cline{3-9}
      &  & Few-shot & 1\_2\_3\_BCNF & 180 & 0.2093 & 0.3889 & 0.5708 & 0.4864 \\\arrayrulecolor{gray!30}\cline{4-9}
      &  &  & 2\_3\_BCNF & 180 & 0.2427 & 0.4833 & 0.6367 & 0.4400 \\\arrayrulecolor{gray!30}\cline{4-9}
      &  &  & 3\_BCNF & 180 & 0.3787 & 0.7278 & 0.5882 & 0.4830 \\\arrayrulecolor{gray!30}\cline{4-9}
      &  &  & BCNF & 180 & 0.2736 & 0.6778 & 0.4911 & 0.3818 \\\arrayrulecolor{gray!30}\cline{4-9}
      &  &  & NONE & 180 & 0.8412 & 0.9389 & 0.9130 & 0.8369 \\\arrayrulecolor{gray!60}\cline{2-9}
      & Mixtral-8x7B-Instruct-v0.1 & Zero-shot & 1\_2\_3\_BCNF & 180 & 0.0316 & 0.2000 & 0.2692 & 0.1726 \\\arrayrulecolor{gray!30}\cline{4-9}
      &  &  & 2\_3\_BCNF & 180 & 0.0157 & 0.1667 & 0.3103 & 0.1644 \\\arrayrulecolor{gray!30}\cline{4-9}
      &  &  & 3\_BCNF & 180 & 0.0654 & 0.3667 & 0.3094 & 0.1352 \\\arrayrulecolor{gray!30}\cline{4-9}
      &  &  & BCNF & 180 & 0.0338 & 0.3333 & 0.3048 & 0.1053 \\\arrayrulecolor{gray!30}\cline{4-9}
      &  &  & NONE & 180 & 0.3994 & 0.5889 & 0.6424 & 0.5272 \\\arrayrulecolor{gray!50}\cline{3-9}
      &  & Few-shot & 1\_2\_3\_BCNF & 180 & 0.0333 & 0.2333 & 0.3614 & 0.1968 \\\arrayrulecolor{gray!30}\cline{4-9}
      &  &  & 2\_3\_BCNF & 180 & 0.0390 & 0.2222 & 0.4463 & 0.1945 \\\arrayrulecolor{gray!30}\cline{4-9}
      &  &  & 3\_BCNF & 180 & 0.0828 & 0.4056 & 0.4510 & 0.1723 \\\arrayrulecolor{gray!30}\cline{4-9}
      &  &  & BCNF & 180 & 0.0661 & 0.4333 & 0.4094 & 0.1299 \\\arrayrulecolor{gray!30}\cline{4-9}
      &  &  & NONE & 180 & 0.5315 & 0.7778 & 0.7859 & 0.5233 \\\arrayrulecolor{gray!60}\cline{2-9}
      & Llama-3.3-70B-Instruct & Zero-shot & 1\_2\_3\_BCNF & 180 & 0.1440 & 0.2611 & 0.5856 & 0.4598 \\\arrayrulecolor{gray!30}\cline{4-9}
      &  &  & 2\_3\_BCNF & 180 & 0.1674 & 0.3444 & 0.6131 & 0.3540 \\\arrayrulecolor{gray!30}\cline{4-9}
      &  &  & 3\_BCNF & 180 & 0.1253 & 0.3556 & 0.6068 & 0.2708 \\\arrayrulecolor{gray!30}\cline{4-9}
      &  &  & BCNF & 180 & 0.0564 & 0.2722 & 0.5102 & 0.1578 \\\arrayrulecolor{gray!30}\cline{4-9}
      &  &  & NONE & 180 & 0.5028 & 0.7778 & 0.7181 & 0.5596 \\\arrayrulecolor{gray!50}\cline{3-9}
      &  & Few-shot & 1\_2\_3\_BCNF & 180 & 0.3135 & 0.4222 & 0.7479 & 0.6317 \\\arrayrulecolor{gray!30}\cline{4-9}
      &  &  & 2\_3\_BCNF & 180 & 0.3402 & 0.4611 & 0.7689 & 0.5974 \\\arrayrulecolor{gray!30}\cline{4-9}
      &  &  & 3\_BCNF & 180 & 0.2986 & 0.5000 & 0.6892 & 0.4968 \\\arrayrulecolor{gray!30}\cline{4-9}
      &  &  & BCNF & 180 & 0.1420 & 0.4056 & 0.5010 & 0.3022 \\\arrayrulecolor{gray!30}\cline{4-9}
      &  &  & NONE & 180 & 0.6105 & 0.8944 & 0.7752 & 0.5985 \\\arrayrulecolor{black}\hline
    \end{xltabular}
  }
\end{center}
\twocolumn
\FloatBarrier
\clearpage
\onecolumn

\section{Prompt Examples}
\label{appendix:prompt_examples}
This appendix lists all prompts used in our experiments. Appendix~\ref{appendix:baseline_prompts} presents the single-prompt Baseline prompts for the three experimental settings: \emph{Single}, \emph{Complex}, and \emph{FD-withheld}. Appendix~\ref{appendix:fewshot_prompt} provides the few-shot examples that are inserted into these prompts for the few-shot configurations. Appendix~\ref{appendix:judge_prompt} gives the LLM-as-a-Judge prompt used for the logical axis. Finally, Appendix~\ref{appendix:mars_prompts} presents the system prompts of the three LLM-based MARS agents: the Evidence Agent, the Diagnosis Agent, and the Schema Generator.

\subsection{Baseline Prompt}\label{appendix:baseline_prompts}
\vspace{2ex}

\lstdefinestyle{promptstyle}{
  basicstyle=\ttfamily\footnotesize,
  breaklines=true,
  breakatwhitespace=true,
  columns=fullflexible,
  frame=single,
  framesep=4pt,
  backgroundcolor=\color{gray!8},
  showstringspaces=false,
  keepspaces=true,
  breakautoindent=true,
  breakindent={2\fontcharwd\font`a},
  captionpos=b,
  extendedchars=true,
  literate={--}{{--}}1 {×}{{$\times$}}1 {→}{{$\rightarrow$}}1 {∪}{{$\cup$}}1 {⊆}{{$\subseteq$}}1,
  aboveskip=0pt,
  belowskip=0pt,
}
\makeatletter
\def\lst@ProcessSpace{%
  \ifnum\lst@newlines>\z@ \ifnum\lst@length=\z@
    \global\advance\lst@lostspace\lst@width
    \global\advance\lst@pos\m@ne
    \lst@whitespacetrue
  \else \lst@KeepSpace \fi
  \else \lst@KeepSpace \fi}
\def\lst@KeepSpace{\lst@PrintToken \lst@whitespacetrue
  \lst@AppendOther\lst@outputspace \lst@PrintToken}
\makeatother

\renewcommand{\lstlistingname}{Prompt}

\begin{lstlisting}[style=promptstyle, caption={\emph{Single} experiment. It resolves only the highest-priority normalization violation, with FDs provided. Filling or leaving empty the \texttt{<Few-shot Examples>} placeholder yields the few-shot and zero-shot configurations, respectively.}, label={prompt:fewshot_single_1}]
You are a database expert.

**Chain Rule Reminder:**
- Work through normal forms bottom-up. Even if higher forms are implicitly violated, `violation_types` must list only the earliest violated stage (or "NONE"), and your explanation plus normalization plan must focus on resolving that stage.
- Allowed `violation_types` choices are strictly {"1NF","2NF","3NF","BCNF","NONE"}. If no violation is found at any level, set it to ["NONE"].
- Whenever a violation exists, cite its cause and describe the normalization steps you apply for that stage.

**Instructions:**
- You are given three inputs: (1) the denormalized table schema, (2) table rows, (3) the full functional-dependency list. Inspect these to detect normalization violations.
- Starting from 1NF upward, find the **earliest** violated normal form and normalize the table to fix that violation only (per the chain rule).
- Use the provided schema, rows, and functional dependency list as evidence; cite whether each claim comes from a row pattern or a specific FD.
- Preserve all columns; do not drop any column. Use exact column names from input.
- If decomposing, define proper FOREIGN KEY constraints.
- Every FOREIGN KEY must reference columns that actually exist in the target table (no phantom columns).
- Every FOREIGN KEY must reference a table you explicitly define **within your own SQL DDL**. Never reference placeholder names such as 'InputTable'; only use table names you create.
- When 1NF is violated, keep every column name and value intact and restore atomicity---this necessarily means decomposing multi-valued attributes into separate tables.

**Output Requirements:**
- Output a single JSON object with keys: `sql_ddl`, `explanation`, `violation_types`. **Do NOT include `functional_dependencies` or any extra keys.**
- Allowed violation_types: ["1NF","2NF","3NF","BCNF","NONE"].
- If `violation_types` is ["NONE"], keep the schema unchanged in `sql_ddl`, do not decompose, and explain briefly why all normal forms hold.
- Strict consistency: violation_types must exactly match explanation sections.

**Normalization Rules:**
- If there are multivalued cells, 1NF is violated.
- If {{A,B}} is a key and there is a partial FD such as {{A}} -> {{X}}, 2NF is violated.
- If there is a transitive FD (CK -> B -> C) where B is not a key, 3NF is violated.
- If the determinant is not a superkey, BCNF is violated.

% if few-shot: <Few-shot Examples>
Now, analyze the following table.

Schema DDL:
<Original Schema>

Rows JSON:
<Table>

Functional Dependencies:
<Violation FDs>
\end{lstlisting}
\clearpage

\begin{lstlisting}[style=promptstyle, caption={\emph{Complex} experiment. It resolves all violations, with FDs provided.}, label={prompt:fewshot_combined_1}]
You are a database expert.

**Chain Rule (Resolve All Violations):**
- Identify every violated normal form from 1NF up to BCNF and fix them all.
- `violation_types` must list every violated stage in order (e.g., ["1NF","2NF","3NF","BCNF"] when all are broken; use ["NONE"] if none are broken).
- For each violated stage, cite the cause and describe the normalization steps you apply.

**Instructions:**
- You are given three inputs: (1) the denormalized table schema, (2) table rows, (3) the full functional-dependency list. Inspect these to detect normalization violations.
- Starting from 1NF upward, identify every violated normal form through BCNF and normalize the table to fix all detected violations.
- Use the provided schema, rows, and functional dependency list as evidence; cite whether each claim comes from a row pattern or a specific FD.
- Preserve all columns; do not drop any column. Use exact column names from input.
- If decomposing, define proper FOREIGN KEY constraints.
- Every FOREIGN KEY must reference columns that actually exist in the target table (no phantom columns).
- Every FOREIGN KEY must reference a table you explicitly define **within your own SQL DDL**. Never reference placeholder names such as 'InputTable'; only use table names you create.
- When 1NF is violated, keep every column name and value intact and restore atomicity---this necessarily means decomposing multi-valued attributes into separate tables.

**Output Requirements:**
- Output a single JSON object with keys: `sql_ddl`, `explanation`, `violation_types`. **Do NOT include `functional_dependencies` or any extra keys.**
- Allowed violation_types: ["1NF","2NF","3NF","BCNF","NONE"].
- If `violation_types` is ["NONE"], keep the schema unchanged in `sql_ddl`, do not decompose, and explain briefly why all normal forms hold.
- Strict consistency: violation_types must exactly match explanation sections.

**Normalization Rules:**
- If there are multivalued cells, 1NF is violated.
- If {{A,B}} is a key and there is a partial FD such as {{A}} -> {{X}}, 2NF is violated.
- If there is a transitive FD (CK -> B -> C) where B is not a key, 3NF is violated.
- If the determinant is not a superkey, BCNF is violated.

% if few-shot: <Few-shot Examples>
Now, analyze the following table.

Schema DDL:
<Original Schema>

Rows JSON:
<Table>

Functional Dependencies:
<Violation FDs>
\end{lstlisting}
\clearpage

\begin{lstlisting}[style=promptstyle, caption={\emph{FD-withheld} experiment. It resolves all violations, without FDs provided.}, label={prompt:fewshot_combined_0}]
You are a database expert.

**Chain Rule (Resolve All Violations):**
- Identify every violated normal form from 1NF up to BCNF and fix them all.
- `violation_types` must list every violated stage in order (e.g., ["1NF","2NF","3NF","BCNF"] when all are broken; use ["NONE"] if none are broken).
- For each violated stage, cite the cause and describe the normalization steps you apply.

**Instructions:**
- You are given the denormalized table schema and table rows. Functional dependencies may be hidden, so infer only dependencies strongly supported by column semantics, keys, uniqueness constraints, and repeated row patterns.
- Starting from 1NF upward, identify every violated normal form through BCNF and normalize the table to fix all detected violations.
- Use the provided schema and rows as evidence; cite whether each claim comes from row patterns, key/unique constraints, or conservative semantic inference.
- Preserve all columns; do not drop any column. Use exact column names from input.
- If decomposing, define proper FOREIGN KEY constraints.
- Every FOREIGN KEY must reference columns that actually exist in the target table (no phantom columns).
- Every FOREIGN KEY must reference a table you explicitly define **within your own SQL DDL**. Never reference placeholder names such as 'InputTable'; only use table names you create.
- When 1NF is violated, keep every column name and value intact and restore atomicity---this necessarily means decomposing multi-valued attributes into separate tables.

**Output Requirements:**
- Output a single JSON object with keys: `sql_ddl`, `explanation`, `violation_types`. **Do NOT include `functional_dependencies` or any extra keys.**
- Allowed violation_types: ["1NF","2NF","3NF","BCNF","NONE"].
- If `violation_types` is ["NONE"], keep the schema unchanged in `sql_ddl`, do not decompose, and explain briefly why all normal forms hold.
- Strict consistency: violation_types must exactly match explanation sections.

**Normalization Rules:**
- If there are multivalued cells, 1NF is violated.
- If {{A,B}} is a key and there is a partial FD such as {{A}} -> {{X}}, 2NF is violated.
- If there is a transitive FD (CK -> B -> C) where B is not a key, 3NF is violated.
- If the determinant is not a superkey, BCNF is violated.

% if few-shot: <Few-shot Examples>
Now, analyze the following table.

Schema DDL:
<Original Schema>

Rows JSON:
<Table>

**FD Inference Guidance:**
- **Rule 1 (Semantics over Statistics):** Do not rely solely on data patterns. Only accept FDs that make logical sense based on the column meanings and common knowledge. Ignore coincidental correlations.
- **Rule 2 (Decomposition via Superkey):** Identify Candidate Keys based on your inferred FDs. If a determinant X (in X->Y) is NOT a superkey, it is a violation (2NF partial or 3NF transitive). Decompose the table to resolve it.
- **Rule 3 (Conservative Design):** Do NOT decompose tables unless a clear violation of Rule 2 is found. If multiple schema designs are possible, prefer the one that preserves the original structure as much as possible.
- **Rule 4 (Verification):** Ensure Lossless Join and Referential Integrity.
\end{lstlisting}
\clearpage

\subsection{Few-shot Prompt}
\label{appendix:fewshot_prompt}
\begin{lstlisting}[style=promptstyle, caption={Few-shot Example.}, label={prompt:fewshot_exemplars}]
Here are examples to guide you.

---
**Example 1a**
*Input:*
Schema DDL:
```sql
CREATE TABLE InputTable (building_id INT, building_short_name VARCHAR(255), building_full_name VARCHAR(255), building_description VARCHAR(255), building_address VARCHAR(255), building_manager VARCHAR(255), building_phone VARCHAR(255), room_count VARCHAR(255), building_full_name_id INT, building_short_name_description VARCHAR(255), building_id_sub_id INT, PRIMARY KEY (building_id, building_id_sub_id), UNIQUE (building_full_name, building_full_name_id));
```
Rows JSON:
```json
{"columns": ["building_id", "building_short_name", "building_full_name", "building_description", "building_address", "building_manager", "building_phone", "room_count", "building_full_name_id", "building_short_name_description", "building_id_sub_id"],
 "rows": [
  [191, "The Eugene", "The Eugene", "Flat", "71537 Gorczany Inlet Wisozkburgh, AL 08256", "Melyssa", "(609)946-0491", "9", 1, "Description for The Eugene", 1],
  [225, "Columbus Square", "Columbus Square", "Studio", "0703 Danika Mountains Apt. 362 Mohrland, AL 56839-5028", "Kyle", "1-724-982-9507x640", "7|8|6|8", 1, "Description for Columbus Square", 1],
  [624, "Stuyvesant Town", "Stuyvesant Town", "Studio", "101 Queenie Mountains Suite 619 New Korbinmouth, KS 88726-1376", "Marie", "(145)411-6406", "5|8", 1, "Description for Stuyvesant Town", 1],
  [673, "Barclay Tower", "Barclay Tower", "Flat", "1579 Runte Forges Apt. 548 Leuschkeland, OK 12009-8683", "Rogers", "1-326-267-3386x613", "3", 1, "Description for Barclay Tower", 1],
  ... (+4 more rows with building_id_sub_id=1, +8 rows with building_id_sub_id=2) ...
 ]}
```
Functional Dependencies:
- {building_id, building_id_sub_id} -> {room_count, building_short_name, building_short_name_description, building_address, building_full_name, building_phone, building_description, building_manager, building_full_name_id}
- {building_full_name, building_full_name_id} -> {room_count, building_short_name, building_short_name_description, building_address, building_phone, building_description, building_manager, building_id, building_id_sub_id}
- {building_id} -> {building_short_name, building_full_name, building_description, building_address, building_manager, building_phone}
- {building_short_name} -> {building_short_name_description}
- {building_short_name} -> {building_full_name}
*Correct Output:*
```json
{
    "sql_ddl":
      CREATE TABLE Apartment_Buildings_key (
        building_short_name CHAR(15), building_full_name VARCHAR(80),
        building_short_name_description TEXT,
        PRIMARY KEY (building_short_name));
      CREATE TABLE Apartment_Buildings_detail (
        building_id INTEGER, building_short_name CHAR(15),
        building_description VARCHAR(255), building_address VARCHAR(255),
        building_manager VARCHAR(50), building_phone VARCHAR(80),
        PRIMARY KEY (building_id),
        FOREIGN KEY (building_short_name) REFERENCES Apartment_Buildings_key (building_short_name));
      CREATE TABLE Apartment_Buildings_sub_detail (
        building_id INTEGER, building_id_sub_id INTEGER,
        building_full_name_id INTEGER,
        PRIMARY KEY (building_id, building_id_sub_id),
        FOREIGN KEY (building_id) REFERENCES Apartment_Buildings_detail (building_id));
      CREATE TABLE Apartment_Buildings_mv_detail (
        building_id INTEGER, building_id_sub_id INTEGER, room_count TEXT,
        PRIMARY KEY (building_id, building_id_sub_id, room_count),
        FOREIGN KEY (building_id, building_id_sub_id) REFERENCES Apartment_Buildings_sub_detail (building_id, building_id_sub_id));
    "explanation": "First violation: 1NF. Cause: Multi-valued attribute present in column 'room_count' (violates atomicity). Fix: normalize by splitting multi-valued attributes into atomic values.
      Remaining violation: 2NF. Cause: Partial dependency exists ({building_id} -> {building_short_name, building_full_name, building_description, building_address, building_manager, building_phone}). Fix: remove partial dependencies so non-keys depend on the full composite key.
      Remaining violation: 3NF. Cause: Transitive dependency exists ({building_short_name} -> {building_short_name_description}). Fix: remove transitive dependencies so non-keys depend only on keys.
      Remaining violation: BCNF. Cause: Non-superkey determinant violates BCNF ({building_short_name} -> {building_full_name}). Fix: decompose non-superkey determinants so every determinant is a candidate key.
      All normalization violations are resolved.",
    "violation_types": ["1NF", "2NF", "3NF", "BCNF"]
}
```

---
**Example 2a**
*Input:*
Schema DDL:
```sql
CREATE TABLE InputTable (building_id INT, building_short_name VARCHAR(255), building_full_name VARCHAR(255), building_description VARCHAR(255), building_address VARCHAR(255), building_manager VARCHAR(255), building_phone VARCHAR(255), building_full_name_id INT, building_short_name_description VARCHAR(255), building_id_sub_id INT, PRIMARY KEY (building_id, building_id_sub_id), UNIQUE (building_full_name, building_full_name_id));
```
Rows JSON: (same 8 building tuples as Example 1a, replicated across building_id_sub_id in {1,2}, with no multi-valued 'room_count' column).
Functional Dependencies:
- {building_id, building_id_sub_id} -> {building_short_name_description, building_phone, building_full_name_id, building_short_name, building_manager, building_address, building_full_name, building_description}
- {building_id} -> {building_short_name, building_full_name, building_description, building_address, building_manager, building_phone}
- {building_short_name} -> {building_full_name}
- {building_full_name, building_full_name_id} -> {building_short_name_description, building_phone, building_short_name, building_id, building_manager, building_address, building_description, building_id_sub_id}
- {building_short_name} -> {building_short_name_description}
*Correct Output:*
```json
{
    "sql_ddl":
      CREATE TABLE Apartment_Buildings_key (
        building_short_name CHAR(15), building_full_name VARCHAR(80),
        building_short_name_description TEXT,
        PRIMARY KEY (building_short_name));
      CREATE TABLE Apartment_Buildings_detail (
        building_id INTEGER, building_short_name CHAR(15),
        building_description VARCHAR(255), building_address VARCHAR(255),
        building_manager VARCHAR(50), building_phone VARCHAR(80),
        PRIMARY KEY (building_id),
        FOREIGN KEY (building_short_name) REFERENCES Apartment_Buildings_key (building_short_name));
      CREATE TABLE Apartment_Buildings_sub_detail (
        building_id INTEGER, building_id_sub_id INTEGER,
        building_full_name_id INTEGER,
        PRIMARY KEY (building_id, building_id_sub_id),
        FOREIGN KEY (building_id) REFERENCES Apartment_Buildings_detail (building_id));
    "explanation": "First violation: 2NF. Cause: Partial dependency exists ({building_id} -> {building_short_name, building_full_name, building_description, building_address, building_manager, building_phone}). Fix: remove partial dependencies so non-keys depend on the full composite key.
      Remaining violation: 3NF. Cause: Transitive dependency exists ({building_short_name} -> {building_short_name_description}). Fix: remove transitive dependencies so non-keys depend only on keys.
      Remaining violation: BCNF. Cause: Non-superkey determinant violates BCNF ({building_short_name} -> {building_full_name}). Fix: decompose non-superkey determinants so every determinant is a candidate key.
      All normalization violations are resolved.",
    "violation_types": ["2NF", "3NF", "BCNF"]
}
```

---
**Example 3a**
*Input:*
Schema DDL:
```sql
CREATE TABLE InputTable (building_id INT, building_short_name VARCHAR(255), building_full_name VARCHAR(255), building_description VARCHAR(255), building_address VARCHAR(255), building_manager VARCHAR(255), building_phone VARCHAR(255), building_full_name_id INT, building_short_name_description VARCHAR(255), PRIMARY KEY (building_id), UNIQUE (building_full_name, building_full_name_id));
```
Rows JSON: (same 8 building tuples as Example 1a, no sub_id column, no multi-valued cells).
Functional Dependencies:
- {building_id} -> {building_short_name_description, building_phone, building_full_name_id, building_short_name, building_manager, building_address, building_full_name, building_description}
- {building_short_name} -> {building_full_name}
- {building_full_name, building_full_name_id} -> {building_short_name_description, building_phone, building_short_name, building_id, building_manager, building_address, building_description}
- {building_short_name} -> {building_short_name_description}
*Correct Output:*
```json
{
    "sql_ddl":
      CREATE TABLE Apartment_Buildings_key (
        building_short_name CHAR(15), building_full_name VARCHAR(80),
        building_short_name_description TEXT,
        PRIMARY KEY (building_short_name));
      CREATE TABLE Apartment_Buildings_detail (
        building_id INTEGER, building_short_name CHAR(15),
        building_description VARCHAR(255), building_address VARCHAR(255),
        building_manager VARCHAR(50), building_phone VARCHAR(80),
        building_full_name_id INTEGER,
        PRIMARY KEY (building_id),
        FOREIGN KEY (building_short_name) REFERENCES Apartment_Buildings_key (building_short_name));
    "explanation": "First violation: 3NF. Cause: Transitive dependency exists ({building_short_name} -> {building_short_name_description}). Fix: remove transitive dependencies so non-keys depend only on keys.
      Remaining violation: BCNF. Cause: Non-superkey determinant violates BCNF ({building_short_name} -> {building_full_name}). Fix: decompose non-superkey determinants so every determinant is a candidate key.
      All normalization violations are resolved.",
    "violation_types": ["3NF", "BCNF"]
}
```

---
**Example 4a**
*Input:*
Schema DDL:
```sql
CREATE TABLE InputTable (building_id INT, building_short_name VARCHAR(255), building_full_name VARCHAR(255), building_description VARCHAR(255), building_address VARCHAR(255), building_manager VARCHAR(255), building_phone VARCHAR(255), building_full_name_id INT, PRIMARY KEY (building_id), UNIQUE (building_full_name, building_full_name_id));
```
Rows JSON: (same 8 building tuples as Example 1a, without sub_id and without short_name_description columns).
Functional Dependencies:
- {building_id} -> {building_phone, building_full_name_id, building_short_name, building_manager, building_address, building_full_name, building_description}
- {building_short_name} -> {building_full_name}
- {building_full_name, building_full_name_id} -> {building_phone, building_short_name, building_id, building_manager, building_address, building_description}
*Correct Output:*
```json
{
    "sql_ddl":
      CREATE TABLE Apartment_Buildings_key (
        building_short_name CHAR(15), building_full_name VARCHAR(80),
        PRIMARY KEY (building_short_name));
      CREATE TABLE Apartment_Buildings_detail (
        building_id INTEGER, building_short_name CHAR(15),
        building_description VARCHAR(255), building_address VARCHAR(255),
        building_manager VARCHAR(50), building_phone VARCHAR(80),
        building_full_name_id INTEGER,
        PRIMARY KEY (building_id),
        FOREIGN KEY (building_short_name) REFERENCES Apartment_Buildings_key (building_short_name));
    "explanation": "First violation: BCNF. Cause: Non-superkey determinant violates BCNF ({building_short_name} -> {building_full_name}). Fix: decompose non-superkey determinants so every determinant is a candidate key.
      All normalization violations are resolved.",
    "violation_types": ["BCNF"]
}
```

---
**Example 5a**
*Input:*
Schema DDL:
```sql
CREATE TABLE InputTable (building_id INT, building_short_name VARCHAR(255), building_full_name VARCHAR(255), building_description VARCHAR(255), building_address VARCHAR(255), building_manager VARCHAR(255), building_phone VARCHAR(255), PRIMARY KEY (building_id));
```
Rows JSON: (same 8 building tuples as Example 1a, without room_count, sub_id, full_name_id, and short_name_description columns).
Functional Dependencies:
- {building_id} -> {building_address, building_full_name, building_phone, building_short_name, building_manager, building_description}
*Correct Output:*
```json
{
    "sql_ddl":
      CREATE TABLE Apartment_Buildings (
        building_id INTEGER, building_short_name VARCHAR(255),
        building_full_name VARCHAR(255), building_description VARCHAR(255),
        building_address VARCHAR(255), building_manager VARCHAR(255),
        building_phone VARCHAR(255),
        PRIMARY KEY (building_id));
    "explanation": "No normalization violation is present. The primary key building_id determines all non-key attributes, all values are atomic, and there are no partial, transitive, or non-superkey dependencies requiring decomposition.",
    "violation_types": ["NONE"]
}
```
\end{lstlisting}
\clearpage

\subsection{LLM-as-a-Judge Prompt}\label{appendix:judge_prompt}
\vspace{2ex}
\begin{lstlisting}[style=promptstyle, caption={LLM-as-a-Judge prompt.}, label={prompt:llm_judge}]
You are an extremely strict and meticulous database professor. Your task is to evaluate a candidate Language Model's response for a database normalization problem.

CRITICAL: Do NOT grade violation-type identification itself. That is assessed externally via exact set match. Your job is to evaluate ONLY the three criteria below. The candidate must use only allowed violation types: ["1NF","2NF","3NF","BCNF","NONE"]. Do NOT invent categories.

Golden Answer (Ground Truth):
- Golden Violation Types (for reference only; do not grade exact matching here; allowed set = ["1NF","2NF","3NF","BCNF","NONE"]):
<golden_violation_types>
- Golden Explanation:
```
<golden_explanation>
```

Candidate LLM's Output (to be evaluated):
- Candidate Violation Types (must be subset of the allowed set; consistency will be checked against the candidate's own explanation):
<llm_violation_types>
- Candidate Explanation:
```
<llm_explanation>
```
- Candidate SQL DDL:
```sql
<llm_sql_ddl>
```

Evaluation Rubric (score each 0-5, with a concise justification):

1. Logical Coherence of Decomposition (0-5)
   [WHAT: Structural Consistency with Expected Decomposition]
   - Evaluate whether the candidate's final DDL provides a plausible decomposition that satisfies the constraints described in the Golden Explanation.
   - Check semantic matching of tables, primary keys, foreign keys, and column sets implied by the narrative.
   - Naming differences are acceptable; structural mismatches are errors.
   - Do NOT evaluate logical reasoning, terminology, or expression quality---only structural correctness.

2. Explanation-Schema Alignment (0-5)
   [CONSISTENCY: The candidate's own internal consistency]
   - Evaluate consistency between the candidate's explanation, candidate's DDL, and candidate's violation_types.
   - If multiple violation types are listed, the explanation MUST be sectioned by type.
   - Each section MUST include required evidence:
     - 1NF: cite exact multi-valued/repeating column(s) and the pattern (e.g., pipe/comma separation).
     - 2NF: cite an explicit partial FD using set notation, e.g., {A, B} is a composite key and {A} -> {X}.
     - 3NF: cite a transitive chain (e.g., CK -> B -> C) and explicitly state B is not a key.
     - BCNF: explicitly state "the determinant is NOT a superkey" and name the determinant attributes.
   - Do NOT compare with Golden---only check the candidate's internal consistency.

3. Explanation Quality (0-5)
   [EXPLANATION: Conceptual accuracy and clarity]
   - Evaluate the overall quality of the candidate's explanation.
   - Check whether normalization and database terms (e.g., 1NF/2NF/3NF/BCNF,
     functional dependency, determinant, superkey, partial dependency,
     transitive dependency) are used with their standard meanings.
   - Assess the clarity, conciseness, and logical organization of the explanation.
   - Penalize misuse of terminology or unnecessary verbosity.
   - Do NOT evaluate structural correctness or explanation-schema alignment.

Detailed Scoring Scale (reference):
- 5: Perfect against Golden for logic/schema; rigorous evidence and internal consistency.
- 4: Minor inconsequential differences (naming); evidence largely sufficient.
- 3: Generally sound but with gaps (minor schema or evidence omissions).
- 2: Significant logical inconsistencies or missing required evidence.
- 1: Mostly incorrect or incoherent.
- 0: Missing/irrelevant.

Required Output Format:
Return one valid JSON object only. Do not include Markdown fences or any extra text.
{
  "logical_coherence_of_decomposition": { "score": <score_0_to_5>, "justification": "<text>" },
  "explanation_schema_alignment":        { "score": <score_0_to_5>, "justification": "<text>" },
  "explanation_quality":                 { "score": <score_0_to_5>, "justification": "<text>" }
}
Please evaluate the candidate's output now based on the three criteria only.
\end{lstlisting}

\clearpage
\subsection{MARS Agent Prompts}
\label{appendix:mars_prompts}

\begin{lstlisting}[style=promptstyle, caption={MARS Evidence Agent system prompt.}, label={prompt:mars_evidence}]
You are a functional dependency evidence agent for database normalization.
Your task is to identify functional dependency (FD) evidence that may matter for normalization and separate it into reliable, hypothesis, and rejected groups.

Input handling:
- Derive FD candidates from the schema and the sample_rows. The schema gives table_name, columns, and schema_ddl.

FD groups:
- reliable_fds: FDs with strong evidence.
- hypothesis_fds: plausible FDs that are not contradicted by rows but are not strong enough to be reliable.
- rejected_fds: FD candidates contradicted by rows or clearly invalid as FD claims.

Evidence principles:
- Row consistency alone does not prove an FD; absence of a row conflict is not proof.
- A row conflict occurs when two rows have identical non-null LHS values but different non-null RHS values. Such a conflict refutes the FD on the observed data.
- Weak or uncertain evidence is not a rejection reason; use hypothesis_fds for plausible but uncertain FDs.
- An FD's reliability comes from how well the sample_rows witness it, not from how plausible it sounds: a determinant is well-witnessed only when the same value recurs in several rows and always maps to the same RHS. If most of its values appear in only one row it is unproven; keep it in hypothesis_fds (or rejected_fds if clearly incidental), never reliable_fds, even when it looks semantically natural. Sample uniqueness of the LHS by itself is neither proof of an FD nor a reason to discard one.
- A single non-key column is a trustworthy determinant mainly as a lookup: its RHS is a description, label, or attribute of that column that recurs consistently. One that merely lines up with another otherwise-independent value in these few rows is not reliable.
- Foreign keys are not FDs by themselves, but may provide schema context for plausible entity or lookup dependencies.

Evidence types:
- declared_key_fd: derived from a declared primary key or unique key.
- repeated_lhs_support: repeated LHS values consistently map to the same RHS.
- semantic_schema_support: schema meaning supports the dependency.
- fk_context_support: FK or lookup context supports the dependency.
- row_conflict: observed rows contradict the dependency.
- weak_row_only: row-consistent but weak evidence.

Classification guidance:
- Put an FD in reliable_fds when it follows from a declared key or unique constraint, has repeated_lhs_support, or has strong semantic_schema_support.
- Put an FD in hypothesis_fds when it is plausible and row-consistent but lacks enough evidence for reliable_fds.
- Put an FD in rejected_fds only when it has a row conflict or is clearly invalid under the observed schema or rows.

Normalization-relevant search:
- Surface plausible partial dependencies, transitive dependencies, lookup/descriptor dependencies, and alternative determinant dependencies (a non-key column that determines another column).
- Do not label these as normal-form violations; only report them as FD evidence.

1NF evidence:
Inspect sample_rows for non-atomic or multivalued cells, such as delimiter-separated values, JSON/list-like values, arrays, or multiple entities packed into one cell.
Report these in multivalued_columns_evidence.
1NF evidence is not an FD; do not place it in reliable_fds, hypothesis_fds, or rejected_fds.

Output rules:
- Return only valid JSON.
- Every FD must have a unique fd_id.
- lhs and rhs must reference only the given columns.

Return JSON with this structure:

{
  "reliable_fds": [
    {"fd_id": "fd_01", "lhs": ["col"], "rhs": ["col"],
     "evidence_type": ["declared_key_fd | repeated_lhs_support | semantic_schema_support | fk_context_support"],
     "reason": "string"}
  ],
  "hypothesis_fds": [
    {"fd_id": "hfd_01", "lhs": ["col"], "rhs": ["col"],
     "evidence_type": ["weak_row_only"], "reason": "string"}
  ],
  "rejected_fds": [
    {"fd_id": "rfd_01", "lhs": ["col"], "rhs": ["col"],
     "evidence_type": ["row_conflict"], "reason": "string"}
  ],
  "multivalued_columns_evidence": [
    {"column": "col", "status": "confirmed | refuted | newly_proposed", "example_value": "raw cell content"}
  ]
}
\end{lstlisting}

\clearpage
\begin{lstlisting}[style=promptstyle, caption={MARS Diagnosis Agent system prompt.}, label={prompt:mars_diagnosis}]
You are a normal-form diagnosis and decomposition-planning agent for database normalization.

You receive:
- schema: table_name, columns
- initial_relation_state
- candidate_fds: reliable FDs from the Evidence Agent or the provided-FD setting
- weak_fd_candidates: plausible but lower-confidence FDs
- multivalued_columns: columns with explicit non-atomic or multivalued evidence

Your task:
Diagnose which normal forms are violated by the input relation.
Classify candidate_fds and weak_fd_candidates as violating or non-violating.
If supported violations exist, produce a stateful decomposition plan.
Return only valid JSON.

Evaluate each FD X -> A from candidate_fds and weak_fd_candidates with these definitions:
- 1NF: a column stores non-atomic/multivalued values (use multivalued_columns).
- 2NF: a non-prime attribute depends on a proper subset of a composite candidate key.
- 3NF: X -> A is non-trivial, X is not a superkey, and A is non-prime.
- BCNF: X -> A is non-trivial and X is not a superkey, whether A is prime or non-prime.
A valid FD does not necessarily violate a normal form.
Evaluate reliable and weak FDs with the same rules; a weak FD that violates is recorded with evidence_strength="weak".

if resolve_all:
    Chain rule:
    - identify every violated normal form from 1NF up to BCNF and record them all.
    - violation_types must list every violated stage in order (e.g. ["1NF","2NF","3NF","BCNF"]; ["NONE"] if none are broken).
    - For each FD X -> A, record all applicable violations cumulatively; do not stop at the first matching normal form.
    - Plan must resolve every violated stage and continue until no active supported violating FD or multivalued-column violation remains.
    
else:
    Chain rule:
    - Work through normal forms bottom-up and record only the earliest violated stage, even if higher forms are implicitly violated.
    - violation_types must list that single stage (or ["NONE"]), and the plan must resolve that stage only.

If violations exist, build a decomposition plan that resolves the detected violations.
The decomposition plan is a state trace, not an unordered list of independent decompositions.
Each step transforms the current schema state.
Maintain current_relations, initialized to [initial_relation_state].

Normal-form evaluation:
- Whether a relation violates a normal form is determined by its functional dependencies, keys, and multivalued columns alone, as defined in the normal-form definitions above.
- Closures (X+), superkeys, and candidate keys are computed from candidate_fds.
- An FD from weak_fd_candidates enters that computation only when you list it in violating_fds.
- For initial_relation_state and each new relation R, derive candidate keys from attribute closures under the full checked FD set, intersected with that relation's columns. Derive prime attributes as the attributes belonging to at least one candidate key, and non-prime attributes as all remaining attributes.
- Classify input FDs into violating_fds or non_violating_fds. Add derived violations only as needed for planning, with unique IDs and derivations in reason. Judge each violation on its source relation under that FD set.
- An FD whose LHS is a superkey of the relation it is judged on (its closure under that FD set contains every column of that relation) violates no normal form; list it in non_violating_fds.

Stateful planning rules:
- Resolve every violation in violating_fds and every multivalued column that the chain rule requires.
- Step 1 operates on initial_relation_state; each later step must operate on a relation produced by a previous step, never re-applied independently to the original input table.
- An FD is active in R only if lhs ∪ rhs ⊆ R.columns, and a violation is active in R only if it comes from an FD active in R or a multivalued column of R; ignore violations that are not active in the current source relation.
- Resolve lower normal-form violations before higher ones (1NF -> 2NF -> 3NF -> BCNF).
- Each FD-based step splits its source relation on a driver X -> Y_group: X is the LHS of an active FD in violating_fds, X is not a superkey of the source relation, and every attribute of Y_group is in the source relation, not in X, and in X+. Do not invent determinants.
- Group FDs into one driver only when they share the exact same LHS X and are active in the same source relation.
- lossless_join_key is the driver's X, and the created relation holding X -> Y_group has X as its primary key.

Planning requirements:
- relation_blueprint is the final relation state and must cover every original column without data loss.
- Every relation (source, created, and blueprint) needs name, columns, and primary_key. Do not create empty relations, duplicate names, or primary keys outside their relation's columns.

Diagnosis status:
Select exactly one status, checking them in this order:
1. valid_decomposition_plan: violating_fds is non-empty or a multivalued column exists, and the steps resolve every violation the chain rule requires in a complete lossless state trace.
2. plan_unavailable: violating_fds is non-empty or a multivalued column exists, but a lossless decomposition that keeps every original column cannot be constructed.
3. insufficient_supported_evidence: no multivalued column exists, and either both FD lists are empty or every FD in them references a column outside initial_relation_state.
4. confirmed_no_violation: candidate_fds and weak_fd_candidates were examined, violating_fds is empty, and no multivalued column exists.

Return JSON with this structure:

{
  "diagnosis_status": "valid_decomposition_plan | plan_unavailable | insufficient_supported_evidence | confirmed_no_violation",
  "violating_fds": [
    {"fd_id": "fd_01", "lhs": ["key_a"], "rhs": ["attr"], "violates": ["2NF", "3NF", "BCNF"], "reason": "key_a is a proper subset of the candidate key {key_a, key_b} of Remaining_1 and attr is non-prime"}
  ],
  "non_violating_fds": [
    {"fd_id": "fd_02", "reason": "the LHS {key_a, key_b} is a superkey of R"}
  ],
  "violation_types": ["1NF", "2NF", "3NF", "BCNF"],
  "selected_fd_ids": ["fd_01"],
  "planning_mode": "llm_stateful_decomposition_plan",
  "decomposition_steps": [
    {
      "step": 1,
      "target_nf": "1NF",
      "driver_ids": [],
      "fd_ids": [],
      "source_relation": {"name": "R", "columns": ["key_a", "key_b", "attr", "mv_col"], "primary_key": ["key_a", "key_b"]},
      "operation": "atomize_multivalued",
      "multivalued_columns": ["mv_col"],
      "lossless_join_key": ["key_a", "key_b"],
      "created_relations": [
        {"name": "MvValues", "columns": ["key_a", "key_b", "mv_col"], "primary_key": ["key_a", "key_b", "mv_col"]},
        {"name": "Remaining_1", "columns": ["key_a", "key_b", "attr"], "primary_key": ["key_a", "key_b"]}
      ]
    },
    {
      "step": 2,
      "target_nf": "2NF",
      "driver_ids": ["fd_01"],
      "fd_ids": ["fd_01"],
      "source_relation": {"name": "Remaining_1", "columns": ["key_a", "key_b", "attr"], "primary_key": ["key_a", "key_b"]},
      "operation": "decompose_fd_group",
      "driver_lhs": ["key_a"],
      "driver_rhs_group": ["attr"],
      "lossless_join_key": ["key_a"],
      "created_relations": [
        {"name": "KeyAGroup", "columns": ["key_a", "attr"], "primary_key": ["key_a"]},
        {"name": "Remaining_2", "columns": ["key_a", "key_b"], "primary_key": ["key_a", "key_b"]}
      ]
    }
  ],
  "relation_blueprint": [
    {"name": "MvValues", "columns": ["key_a", "key_b", "mv_col"], "primary_key": ["key_a", "key_b", "mv_col"]},
    {"name": "KeyAGroup", "columns": ["key_a", "attr"], "primary_key": ["key_a"]},
    {"name": "Remaining_2", "columns": ["key_a", "key_b"], "primary_key": ["key_a", "key_b"]}
  ]
}

The example shows steps 1-2 (1NF then 2NF); continue with the same X -> Y_group form for any remaining 3NF/BCNF violation.
relation_blueprint lists the final relations. Use actual columns and FD IDs, and concrete primary_key arrays in your output.

For confirmed_no_violation:
- violation_types = ["NONE"]
- selected_fd_ids = []
- planning_mode = "no_decomposition"
- decomposition_steps = []
- relation_blueprint = [initial_relation_state]

For insufficient_supported_evidence:
- violation_types = []
- selected_fd_ids = []
- planning_mode = "insufficient_evidence"
- decomposition_steps = []
- relation_blueprint = [initial_relation_state]

For plan_unavailable:
- violation_types = the detected violated stages
- selected_fd_ids = []
- planning_mode = "no_decomposition"
- decomposition_steps = []
- relation_blueprint = [initial_relation_state]
\end{lstlisting}

\clearpage
\begin{lstlisting}[style=promptstyle, caption={MARS Schema Generator system prompt.}, label={prompt:mars_generation}]
You are the Schema Generator in a database-normalization pipeline.
- Your task is to render the Diagnosis Agent's plan into SQL DDL.

if resolve_all:
    **Chain Rule (Resolve All Violations):**
    - Render a schema that resolves every violation listed in generator_plan.target.violation_types.
    - `violation_types` must list every violated stage in order; use ["NONE"] if none are broken.
    - The SQL DDL and explanation must cover every planned decomposition step needed to resolve all listed violations.
    
else:
    **Chain Rule Reminder:**
    - Render a schema that resolves only the earliest violation selected by generator_plan.
    - `violation_types` must list only that earliest violated stage, or ["NONE"] if no violation exists.
    - The SQL DDL and explanation must focus on that selected stage only.

**Inputs:**
- original schema: table name, columns, declared primary key, and rows.
- generator_plan: diagnosis_status, target (including violation_types and normalization_scope), violating_fds, selected_fd_ids, relation_blueprint, decomposition_steps, and multivalued_columns.
- The generator_plan is the authoritative source for the normalization scope, selected violations, decomposition steps, and final relation blueprint.
- Use the original schema and rows only to preserve exact column names, data semantics, and valid SQL constraints.

**Diagnosis Plan Policy:**
- Follow generator_plan.relation_blueprint and generator_plan.decomposition_steps.
- Apply decomposition_steps sequentially. Each step operates on the current relation state produced by previous steps, not independently on the original input table.
- For each step, use its source_relation, driver_lhs, driver_rhs_group, created_relations, and lossless_join_key.

**SQL DDL Requirements:**
- Preserve all columns. Do not drop any input column.
- Use exact input column names.
- Define primary keys according to generator_plan.relation_blueprint whenever available.
- Each decomposition step must declare a FOREIGN KEY on its lossless_join_key referencing that step's detail relation.
- Every FOREIGN KEY must reference a table explicitly defined in your SQL DDL.
- Every FOREIGN KEY must reference columns that actually exist in the target table.
- Never reference placeholder names such as `InputTable` unless `InputTable` is actually emitted as a table in your SQL DDL.
- Declare parent relations before child relations when foreign keys are used.

**Output Requirements:**
- Output a single JSON object with keys: `sql_ddl`, `explanation`, `violation_types`. **Do NOT include `functional_dependencies` or any extra keys.**
- Allowed violation_types: ["1NF","2NF","3NF","BCNF","NONE"].
- `violation_types` describes which normal forms the original input table violates before decomposition.
- If `violation_types` is ["NONE"], keep the schema unchanged in `sql_ddl`, do not decompose, and explain briefly why all normal forms hold.
- Strict consistency: violation_types must exactly match explanation sections.
\end{lstlisting}

\clearpage
\twocolumn

\label{sec:appendix}

\end{document}